\pdfoutput=1
\RequirePackage{silence}
\documentclass[a4paper,svgnames]{amsart}
\usepackage[hmarginratio={1:1},vmarginratio={1:1},lmargin=60.0pt,tmargin=50.0pt]{geometry}

\usepackage{booktabs}
\allowdisplaybreaks

\usepackage[T1]{fontenc}
\usepackage{latexsym,exscale,amsfonts,amssymb,mathtools}
\usepackage{amsmath,amsthm,amsfonts,amssymb,amscd,textcomp,bbm}
\usepackage{mathrsfs,stackrel}
\usepackage{etoolbox}
\usepackage{tikz,tikz-cd}
\usetikzlibrary{matrix,arrows.meta,decorations.markings,shapes.multipart,decorations.pathreplacing,decorations.pathmorphing}
\usepackage{aliascnt}
\usepackage{ytableau}
\usepackage{xparse}
\usepackage{cite}

\usepackage{dynkin-diagrams}

\usepackage{array}
\newcolumntype{C}{>{$}c<{$}}

\usepackage{xcolor,colortbl}

\definecolor{mygray}{gray}{0.6}
\definecolor{mygraydark}{gray}{0.4}
\definecolor{mygraylight}{gray}{0.85}
\definecolor{spinach}{RGB}{46,139,87}
\definecolor{tomato}{RGB}{255,99,71}
\definecolor{orchid}{RGB}{143,40,194}
\definecolor{neon}{RGB}{77,77,255}
\definecolor{pumpkin}{RGB}{224,180,80}
\definecolor{citron}{RGB}{190,180,90}

\definecolor{lava}{RGB}{207,16,32}
\definecolor{cream}{RGB}{255,253,208}
\definecolor{verdigris}{RGB}{67,179,174}
\definecolor{Black}{RGB}{0,0,0}
\definecolor{mydarkblue}{RGB}{10,10,170}
\definecolor{darkspinach}{RGB}{20,70,20}
\definecolor{darktomato}{RGB}{155,40,30}
\definecolor{darkorchid}{RGB}{50,10,100}
\definecolor{darklava}{RGB}{150,8,16}

\usepackage{todonotes}

\usepackage{enumitem}
\setlist[enumerate]{itemsep=0.15cm,label=\emph{\upshape(\alph*)}}
\setlist[enumerate,2]{itemsep=0.15cm,label=\emph{\upshape(\roman*)}}
\setlist[enumerate,3]{itemsep=0.15cm,label=\emph{\upshape(\Alph*)}}

\let\emph\relax
\DeclareTextFontCommand{\emph}{\bfseries\em}

\newcommand{\acts}{\cdot}

\renewcommand{\dots}{\text{...}}
\renewcommand{\vdots}{\rotatebox{90}{\text{...}}}
\renewcommand{\ddots}{\raisebox{0.175cm}{\rotatebox{-45}{\text{...}}}}

\newcommand{\placeholder}{{}_{-}}

\newcommand{\Hom}{\mathrm{Hom}}
\newcommand{\End}{\mathrm{End}}

\newcommand{\ie}{\text{i.e.}}
\newcommand{\eg}{\text{e.g.}}
\newcommand{\cf}{\text{cf.}}
\newcommand{\etc}{\text{etc.}}

\newcommand{\C}{\mathbb{C}}
\newcommand{\R}{\mathbb{R}}
\newcommand{\HH}{\mathbb{H}}
\newcommand{\N}{\mathbb{Z}_{\geq 0}}

\newcommand{\Z}{\mathbb{Z}}

\newcommand{\fun}{\mathsf{Fun}}
\newcommand{\GL}{\mathsf{GL}}
\newcommand{\im}{\mathrm{Im}}

\newcommand{\lhom}[1][G]{\Hom_{G}}
\newcommand{\plhom}[1][G]{\Hom_{G}^{pl}}
\newcommand{\lend}[1][G]{\End_{G}}
\newcommand{\plend}[1][G]{\End_{G}^{pl}}
\newcommand{\lrep}[1][G]{\mathcal{R}\mathrm{ep}_{\R}(#1)}
\newcommand{\plrep}[1][G]{\mathcal{R}\mathrm{ep}_{\R}^{pl}(#1)}

\newcommand{\rep}[1][V]{#1}
\newcommand{\permrep}[1][X]{\fun(#1,\R)}
\newcommand{\permrepg}[1][X]{\fun_{G}(#1,\R)}
\newcommand{\regrep}{R}
\newcommand{\simple}[1][L]{#1}
\newcommand{\bmat}[1][Q]{\mathtt{#1}}
\newcommand{\mat}[1][k]{\mathtt{N}_{#1}}

\newcommand{\lmat}[1][a]{\mathtt{M}_{#1}}

\newcommand{\ggraph}[2]{\Gamma_{#1}^{#2}}
\newcommand{\iggraph}[2]{i\Gamma_{#1}^{#2}}
\newcommand{\graph}[1][\mathsf{ReLU}]{\Gamma_{#1}}
\newcommand{\igraph}[1][\mathsf{ReLU}]{i\Gamma_{#1}}
\newcommand{\com}[2]{\gamma_{#1}(#2)}

\newcommand{\alcove}{\mathcal{A}}
\newcommand{\zroot}[1][k]{\zeta_{#1}}

\newcommand{\ran}{\theta}

\newcommand{\cyclic}[1][n]{C_{#1}}

\newcommand{\plmap}{\mathsf{PL}}
\newcommand{\relu}{\mathsf{ReLU}}
\newcommand{\absf}{\mathsf{Abs}}

\newcommand{\ord}[1][\simple]{\mathrm{ord}(#1)}
\newcommand{\ordp}[1][\simple]{\mathrm{ord}^{\prime}(#1)}

\usepackage{neuralnetwork}
\tikzset{
anchorbase/.style={baseline={([yshift=#1]current bounding box.center)}},
anchorbase/.default={-0.5ex},
tinynodes/.style={font=\tiny,text height=0.25ex,text depth=0.05ex},
smallnodes/.style={font=\scriptsize,text height=0.75ex,text depth=0.15ex},
}
\tikzset{basic/.style={draw,fill=magenta!20,text width=1em,text badly centered}}
\tikzset{input/.style={basic,circle}}
\tikzset{weights/.style={basic,rectangle}}
\tikzset{functions/.style={basic,circle,fill=spinach!20}}

\def\NewTheorem#1{%
\newaliascnt{#1}{equation}%
\newtheorem{#1}[#1]{#1}%
\aliascntresetthe{#1}%
\expandafter\def\csname #1autorefname\endcsname{#1}%
}
\def\equationautorefname~#1\null{(#1)\null}

\numberwithin{equation}{subsection}

\NewTheorem{Proposition}
\NewTheorem{Theorem}
\NewTheorem{Corollary}
\AtEndEnvironment{Corollary}{\null\hfill$\square$}%
\NewTheorem{Lemma}
\theoremstyle{definition}
\NewTheorem{Definition}
\NewTheorem{Notation}
\NewTheorem{Example}
\AtEndEnvironment{Example}{\null\hfill$\Diamond$}%
\NewTheorem{Examples}
\AtEndEnvironment{Examples}{\vskip-10mm\null\hfill$\Diamond$}%

\theoremstyle{remark}
\NewTheorem{Remark}
\NewTheorem{Assumption}

\usepackage[hypertexnames=false]{hyperref}
\usepackage{bookmark}
\hypersetup{
pdftoolbar=true,
pdfmenubar=true,
pdffitwindow=false,
pdfstartview={FitH},
pdftitle={Equivariant neural networks and piecewise linear representation theory},
pdfauthor={Joel Gibson, Daniel Tubbenhauer and Geordie Williamson},
pdfsubject={},
pdfcreator={Joel Gibson, Daniel Tubbenhauer and Geordie Williamson},
pdfproducer={Joel Gibson, Daniel Tubbenhauer and Geordie Williamson},
pdfkeywords={},
pdfnewwindow=true,
colorlinks=true,
linkcolor=mydarkblue,
citecolor=teal,
filecolor=magenta,
urlcolor=orchid,
linkbordercolor=lava,
citebordercolor=teal,
urlbordercolor=orchid,
linktocpage=true
}

\def\makeautorefname#1#2{\csdef{#1autorefname}{#2}}
\makeautorefname{section}{Section}%
\makeautorefname{subsection}{Section}%
\makeautorefname{subsubsection}{Section}%

\begin{document}
\title[Equivariant neural networks and piecewise linear representation theory]{Equivariant neural networks and piecewise linear representation theory}
\author[J. Gibson, D. Tubbenhauer and G. Williamson]{Joel Gibson, Daniel Tubbenhauer and Geordie Williamson}

\address{J.G.: The University of Sydney, School of Mathematics and Statistics F07, Office Carslaw 827, NSW 2006, Australia, \href{https://www.jgibson.id.au}{www.jgibson.id.au}}
\email{joel@jgibson.id.au}

\address{D.T.: The University of Sydney, School of Mathematics and Statistics F07, Office Carslaw 827, NSW 2006, Australia, \href{http://www.dtubbenhauer.com}{www.dtubbenhauer.com}, https://orcid.org/0000-0001-7265-5047}
\email{daniel.tubbenhauer@sydney.edu.au}

\address{G.W.: The University of Sydney, School of Mathematics and Statistics F07, Office L4.41 Quadrangle A14, NSW 2006, Australia, \href{https://www.maths.usyd.edu.au/u/geordie/}{www.maths.usyd.edu.au/u/geordie/}, https://orcid.org/0000-0003-3672-5284}
\email{g.williamson@sydney.edu.au}

\begin{abstract}
Equivariant neural networks are neural networks with
symmetry. Motivated by the theory of group representations, we
decompose the layers of an 
equivariant neural network into simple representations. The
nonlinear activation functions lead to interesting nonlinear equivariant 
maps between simple representations. For example, the rectified linear unit (ReLU)
gives rise to
piecewise linear maps. We show that these considerations lead to a
filtration of equivariant neural networks, generalizing Fourier
series. This observation might provide a useful tool for
interpreting equivariant neural networks.
\end{abstract}

\subjclass[2020]{Primary: 
20C05; Secondary: 05E10, 68T07.}
\keywords{Equivariant neural networks, representation theory, piecewise linear maps.}

\addtocontents{toc}{\protect\setcounter{tocdepth}{1}}

\maketitle

\tableofcontents

\ytableausetup{centertableaux,mathmode,boxsize=0.4cm}
{\global\arrayrulewidth=0.5mm}

%%%%%%%%%%%%%%%%%%%%%%%%%%%%%%%%%%%%%%%%%

\section{Introduction}\label{S:Introduction}

%%%%%%%%%%%%%%%%%%%%%%%%%%%%%%%%%%%%%%%%%

Neural networks provide flexible and powerful means of approximating a
function. In many applications, one wants to learn a function that is
invariant or equivariant with respect to some symmetries. A
prototypical example is image recognition, where problems are often
invariant under translation. \emph{Equivariant neural networks}
provide a flexible framework for learning such invariant or
equivariant functions.

Equivariant neural networks can be studied using the
mathematical theory of representation theory. (The
mathematical concept of representation is different from the
typical meaning of ``representation'' in machine learning. 
In this paper we exclusively use the term in the mathematical sense.)
In representation theory, simple representations provide the
irreducible atoms of the theory. A main strategy in representation
theory is to take a problem, decompose it into simple representations, and study 
the problem on these basic pieces separately. As we will see, this doesn't quite work
for equivariant neural networks: their nonlinear nature allows for
interaction between simple representations, which is
impossible in the linear world.

However, we will argue in this paper that decomposing 
the layers of an equivariant  neural network into
simple representations is still a very interesting 
thing to do. We are led
naturally to the study of piecewise linear maps 
between simple representations and \emph{piecewise linear representation theory}. 
In concrete terms, the decomposition 
into simple representations leads to a new basis of the
layers of a neural network, generalizing the 
Fourier transform. We hope that this
new basis provides a useful tool to understand and interpret 
equivariant neural networks.

\subsection{What we prove}\label{S:WhatWeProve}

Before diving into the main results 
of our paper, we give a simple but nontrivial example 
of our main observations.

Consider the small vanilla neural network (we often omit labels):
\begin{gather*}
\raisebox{-1cm}{$\begin{neuralnetwork}[height=4]
\newcommand{\vv}[2]{\scalebox{0.8}{$\R$}}
\newcommand{\ww}[4]{$w$}
\inputlayer[count=1,bias=false,text=\vv]
\hiddenlayer[count=3,bias=false,text=\vv]
\link[from layer=0, to layer=1, from node=1, to node=1,label=\ww]
\link[from layer=0, to layer=1, from node=1, to node=2]
\link[from layer=0, to layer=1, from node=1, to node=3]
\outputlayer[count=2,text=\vv] 
\linklayers
\end{neuralnetwork}$}
.
\end{gather*}
As usual, each node represents a copy of $\R$, each arrow is labeled by a weight $w$, and the result of each linear map between layers 
is composed with a nonlinear activation function $f$ before proceeding to the next layer.

The key motivation for building equivariant neural networks
is to replace $\R$ and $w$ by more complicated objects with more symmetry. For example, consider the replacement
\begin{align*}
\R &\rightsquigarrow\text{a suitable space $\fun$ of functions on $\R$},
\\
w &\rightsquigarrow\text{a convolution operator $c_{\ast}\colon\fun\to\fun$},
\\
f &\rightsquigarrow\text{an activation function $\fun\to\fun$ with $\gamma\mapsto f\circ\gamma$}.
\end{align*}
We can depict this as:
\begin{gather}\label{Eq:WhatWeProveConvolution}
\begin{gathered}
\raisebox{-1cm}{$\begin{neuralnetwork}[height=4]
\newcommand{\vv}[2]{\scalebox{0.8}{$\fun$}}
\newcommand{\ww}[4]{conv.}
\inputlayer[count=1,bias=false,text=\vv]
\hiddenlayer[count=3,bias=false,text=\vv]
\link[from layer=0, to layer=1, from node=1, to node=1,label=\ww]
\link[from layer=0, to layer=1, from node=1, to node=2]
\link[from layer=0, to layer=1, from node=1, to node=3]
\outputlayer[count=2,text=\vv] 
\linklayers
\end{neuralnetwork}$}
,\\
\text{where each arrow is e.g. }
\raisebox{-1cm}{\includegraphics[height=2.5cm]{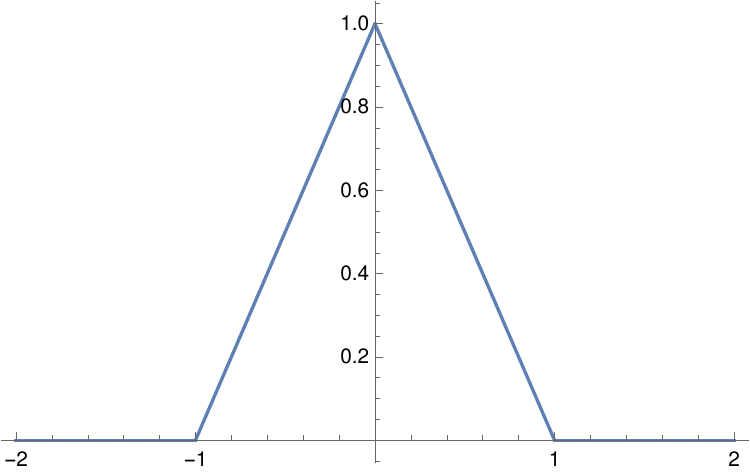}}
\xrightarrow[\text{the unit box}]{\text{convolution with}}
\raisebox{-1cm}{\includegraphics[height=2.5cm]{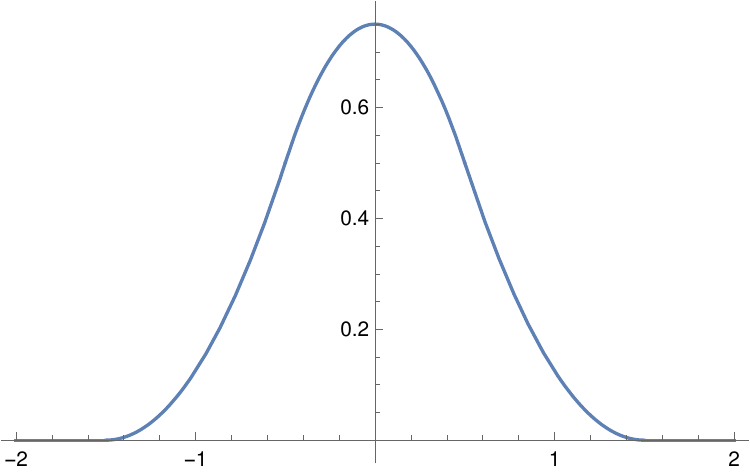}}
.
\end{gathered}
\end{gather}
The actual implementation of this structure on a computer 
would be impossible, but bear with us!

Now assume for a moment that our function is periodic of 
period $2\pi$. It is very natural to ask what happens 
to our neural network \autoref{Eq:WhatWeProveConvolution}
when we expand it in terms of Fourier series. A fundamental result in Fourier theory is that 
convolution operators become diagonal in the Fourier basis.
Hence, in order to understand how signals flow through
the neural network in \autoref{Eq:WhatWeProveConvolution}, it remains to understand how the activation function acts on the 
fundamental frequencies.

A basic, but key, observation is that the Fourier series $\sum_{N\in\Z}c_Ne^{i2\pi(N/P)x}$ of $f(\sin(x))$
\emph{only involves terms of higher resonant
frequency}:
\begin{gather*}
\begin{tikzpicture}[anchorbase]
\node at (0,0) {\includegraphics[height=3.1cm]{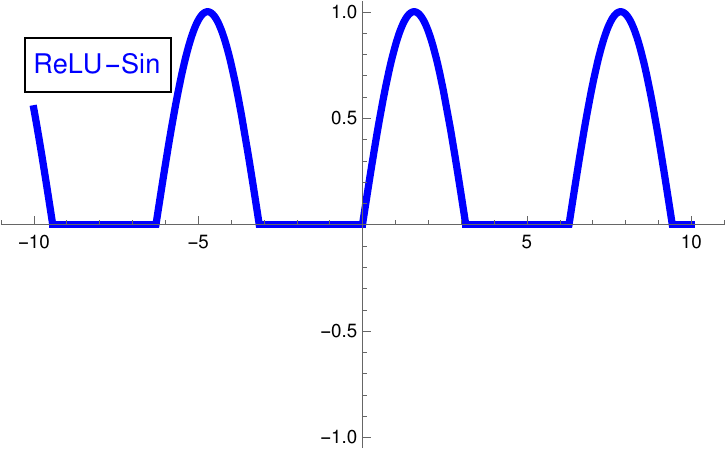}};
\draw[<->] (0,-0.2) to node[below]{$P=2\pi$} (1.45,-0.2);
\end{tikzpicture}
\xrightarrow{\text{Fourier series}}
\begin{aligned}
\raisebox{-1cm}{\includegraphics[height=2.2cm]{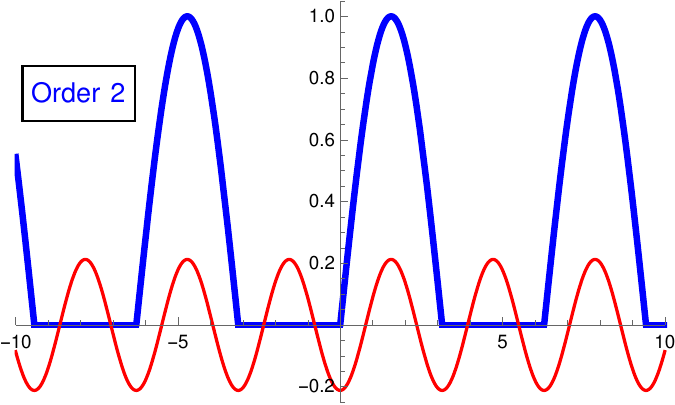}}
&\text{ zoomed in }\colon
\raisebox{-1cm}{\includegraphics[height=2.2cm]{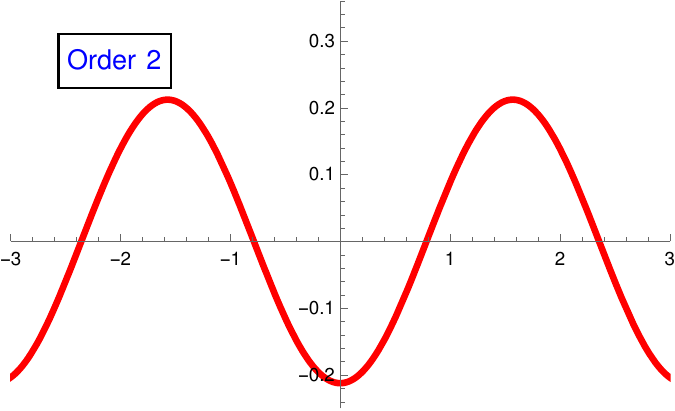}}
\\
\raisebox{-1cm}{\includegraphics[height=2.2cm]{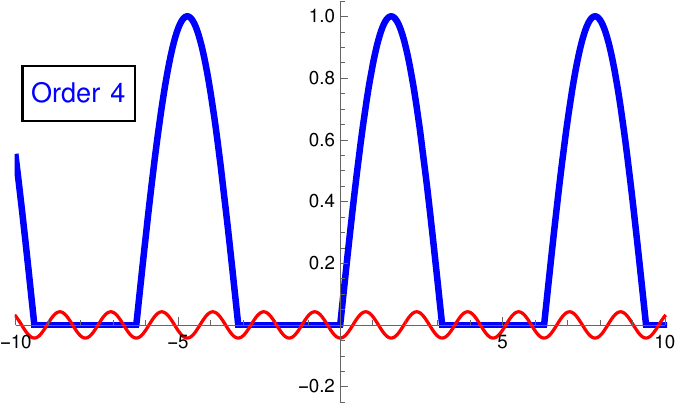}}
&\text{ zoomed in }\colon
\raisebox{-1cm}{\includegraphics[height=2.2cm]{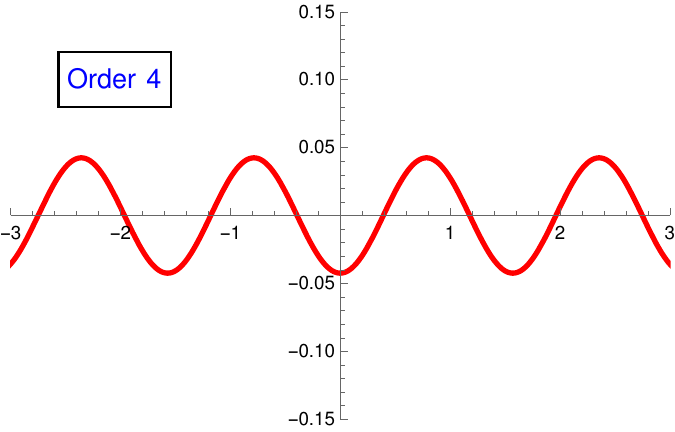}}
\\
\raisebox{-1cm}{\includegraphics[height=2.2cm]{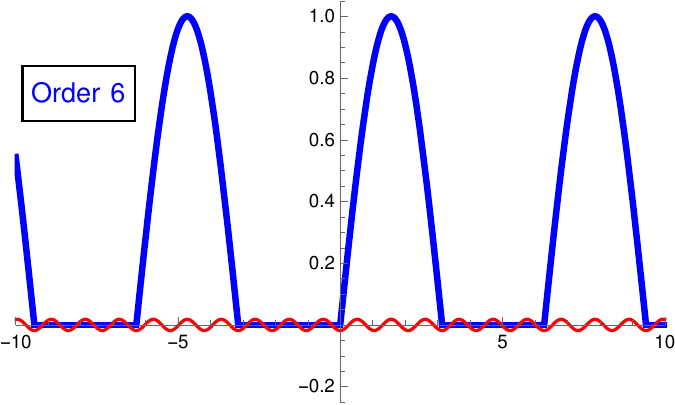}}
&\text{ zoomed in }\colon
\raisebox{-1cm}{\includegraphics[height=2.2cm]{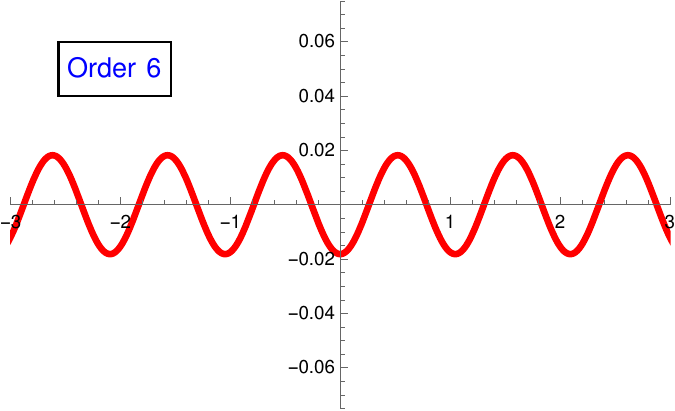}}.
\end{aligned}
\\
\text{Here `Order $N$' means the function $x\mapsto c_{-N}e^{i 2\pi(-N/P)x}+c_{N}e^{i 2\pi(N/P)x}$.}
\end{gather*}
This shows the first few Fourier series terms (small, red) of $f(\sin(x))$ (large, blue) when $f$ is the rectifier linear unit ReLU. What one observes is, while the Fourier series of $\sin(x)$ is trivial, the one of $f(\sin(x))$ has infinitely many nonzero $c_{-N}$, $c_N$ corresponding to summands of higher and higher frequency as $N\to\infty$. This is very similar to what happens when we pluck a string on a guitar: one has a fundamental frequency corresponding to the
note played, as well as higher frequencies (``overtones'', similar to the bottom three pictures above) which
combine to give the distinctive timbre of the guitar.

We show in general that in equivariant neural networks 
one has a \emph{flow of lower frequency to higher resonant frequency, but not conversely}:
\begin{gather*}
\scalebox{0.7}{$\begin{tikzpicture}[anchorbase,->,>=stealth',shorten >=1pt,auto,node distance=3cm,
thick,main node/.style={circle,font=\sffamily\Large\bfseries}]
\node[main node] (1) {high freq.};
\node[main node] (2) [above of=1,yshift=-1.5cm] {mid. freq.};
\node[main node] (5) [above of=2,yshift=-1.5cm] {low freq.};
\node[main node] (3) [right of=1,xshift=1.0cm] {high freq.};
\node[main node] (4) [right of=2,xshift=1.0cm] {mid. freq.};
\node[main node] (6) [right of=5,xshift=1.0cm] {low freq.};
\path[every node/.style={font=\sffamily\small}]
(1) edge [tomato] (3)node[below,xshift=1.5cm]{}
(2) edge [tomato] (3)node[below,xshift=1.5cm]{}
(5) edge [tomato] (3)node[below,xshift=1.5cm]{}
(5) edge [tomato] (4)node[below,xshift=1.5cm]{}
(5) edge [tomato] (6)node[below,xshift=1.5cm]{}
(2) edge [tomato] (4)node[above,xshift=1.5cm]{};
\draw[thick,->] (6,3.0) node[right]{low frequencies} to (6,0) node[right]{high frequencies};
\end{tikzpicture}$}
,\quad
\begin{aligned}
\text{low freq.}&\leftrightsquigarrow\sin(x),
\\
\text{mid. freq.}&\leftrightsquigarrow\sin(5x),
\\
\text{high freq.}&\leftrightsquigarrow\sin(10x).
\end{aligned}
\end{gather*}

This has two concrete consequences for equivariant neural
networks:
\begin{enumerate}[label=(\roman*)]

\item The majority of the complexity of equivariant neural networks
occurs at high frequency.

\item If one would like to learn a low frequency function, then one
can ignore a large part of the neural network corresponding to
high frequency.

\end{enumerate}

For example, a prototypical flow-type diagram, that we will call an \emph{interaction graph}, for equivariant neural networks
built on  $\cyclic[8]\cong\Z/8\Z$, the cyclic group of order 8, is (the colors are for readability only)
\begin{gather*}
%\graph=
\scalebox{0.7}{$\begin{tikzpicture}[anchorbase,->,>=stealth',shorten >=1pt,auto,node distance=3cm,thick,main node/.style={circle,draw,font=\sffamily\Large\bfseries}]
\node[main node] (1) {$+1$};
\node[main node] (2) [above of=1,yshift=-1.2cm] {$-1$};
\node[main node] (3) [above of=2,yshift=-1.2cm] {\scalebox{0.9}{$\pi/2$}};
\node[main node] (4) [right of=3,yshift=1.2cm] {\scalebox{0.9}{$3\pi/4$}};
\node[main node] (5) [left of=3,yshift=1.2cm] {\scalebox{0.9}{$\pi/4$}};
\path[every node/.style={font=\sffamily\small}]
(1) edge [loop below,blue] (1)
(2) edge [loop left,blue] (2)
(3) edge [loop above,blue] (3)
(5) edge [loop left,blue] (5)
(4) edge [loop right,blue] (4)
(2) edge [tomato] (1)
(3) edge [tomato] (2)
edge [tomato,bend left] (1)
(4) edge [tomato] (2)
edge [tomato] (3)
edge [tomato,bend left=10] (1)
edge[<->] (5)
(5) edge [tomato] (2)
edge [tomato] (3)
edge [tomato,bend right=10] (1);
\draw[thick,->] (5,5) node[right]{low frequencies} to (5,0) node[right]{high frequencies};
\end{tikzpicture}$}
,
\end{gather*}
where the nodes are the simple representations of $\cyclic[8]$, and the 
values in the nodes indicates the action of a generator. In this
picture, ``lower frequency'' simple representations are depicted at
the top, and again one has a flow of information from low to high
frequency, but not conversely. Thus, the
same principal as in \autoref{Eq:OverviewComposition} implies that  
high frequency will dominate the picture in large networks. 

\subsection{Main contributions}\label{SS:OverviewResults}

Our main results are:
\begin{enumerate}[label=(\roman*)]

\item We point out that it is meaningful and 
interesting to decompose an equivariant neural network
into simple representations.

\item We argue that equivariant neural networks must be built from permutation representations.
See \autoref{T:PLRepsGEquivariant}.

\item We prove that the existence of equivariant maps 
that are piecewise linear (but not linear) is controlled 
by normal subgroups in fashion similar to Galois theory.
See \autoref{T:PlRepsSchur}.

\item We compute several examples, showing the richness 
of theory even in ``easy'' examples like cyclic groups.
See e.g. \autoref{S:Cyclic}.

\end{enumerate}

\begin{Remark}
All the code that was used for this project can be found following the links on \cite{GiTuWi-pl-reps-code}.
Moreover, some of the examples in \autoref{S:Cyclic} 
can be run in a browser, see the second link in
\cite{GiTuWi-pl-reps-code}. That page also contains additional
examples such as dihedral and symmetric groups and their piecewise
linear representation theory. We encourage the reader to explore these
examples alongside the reading of this paper.
\end{Remark}

\subsection{Related work} 

Various forms of equivariant neural networks
have become basic tools in machine learning. Standard references in the
machine learning literature include \cite{cohen2016group},
\cite[\S 5.2]{bronstein2021geometric} and
\cite{gerken2023geometric}. We will not survey the extensive 
literature here, but instead refer the reader to 
\cite[\S 7]{bronstein2021geometric} and \cite[\S
1.6]{gerken2023geometric} for historical discussion. For excellent
short introductions to equivariant neural networks intended for
mathematicians, see \cite{esteves2020theoretical} and
\cite{lim2022equivariant}.

Many of the results of this paper have already appeared in some form
in the machine learning literature. In particular, our discussion of
the basic structure of equivariant neural networks, and the fact that
all equivariant maps are convolutions (\autoref{L:OverviewEquiNNTwo}) follows the existing
literature closely. Our intention is to give a motivated introduction 
for mathematicians. To the best of our knowledge, our explanation of the
importance of permutation representations (\autoref{T:PLRepsGEquivariant}), our general approach of
decomposing into simple representations and considering piecewise
linear maps, and our classification of such maps in terms of normal
subgroups (\autoref{T:PlRepsSchur}) are new.

%%%%%%%%%%%%%%%%
% Acknowledgments
%%%%%%%%%%%%%%%%

\subsection*{Acknowledgments.} 
We are thankful to Anne Dranowski, Andrew Dudzik, Georg Gottwald, Henry Kvinge, S{\'e}bastien
Racani{\`e}re, and Petar Veli{\v{c}}kovi{\'c} for feedback and
enlightening conversations. We thank the reviewer for a careful reading and valuable feedback. D.T. was 
sponsored by the ARC Future Fellowship FT230100489,
and they
thank rotating pizza slices for inspiration.

%%%%%%%%%%%%%%%%%%%%%%%%%%%%%%%%%%%%%%%%%

\section{Equivariant neural networks and piecewise linear representations}\label{S:Overview}

%%%%%%%%%%%%%%%%%%%%%%%%%%%%%%%%%%%%%%%%%

We now give a (partially informal) account of what we 
understand by equivariant neural networks and 
by piecewise linear representation theory. However, before we get to these topics we need to review a few basics from the
mathematical theory of group representations.

\subsection{Representation theory crash course}\label{SS:OverviewRepTheory}

Representation theory is the 
study of linear symmetries. The emphasis here is 
on symmetries, {e.g.} spaces with a group action.
A thorough introduction can be found in many textbooks such as 
\cite{Be-rep-cohomology}, \cite{FuHa-representation-theory}, \cite{CR} or \cite{Se-rep-theory-finite-groups}.

\begin{Notation}\label{N:OverviewRepTheoryFields}
Unless otherwise stated, we use the real numbers $\R$ as our ground field, all $\R$-vector spaces are finite dimensional, and the groups that we consider are finite.
\end{Notation}

A finite dimensional representation $\phi$ of a finite group $G$ is a
way of writing it down as matrices, {\ie} a homomorphism $\phi\colon
G\to\GL_{n}(\R)$. Equivalently, such a representation is an
$\R$-vector space $\rep$ with a linear action $\acts$ of $G$ such that $g\acts(h\acts v)=gh\acts v$ holds for all $g,h\in G$ and $v\in\rep$.
A subrepresentation is a subspace $\rep[W]\subset\rep[V]$ which is
$G$-stable in the sense that $g\acts w\in\rep[W]$ for all $g\in G$ and $w\in\rep[W]$.

A $G$-equivariant linear 
map $\phi\colon\rep[V]\to\rep[W]$ between two 
$G$-representations is a linear map satisfying $\phi(g\acts v)=g\acts
\phi(v)$ for all $g \in G$ and $v \in V$. Such a map is an isomorphism if it is a bijection, and two $G$-representations are equivalent if these is a $G$-equivariant linear isomorphism between them.
A $G$-representation $V$ is simple if the only $G$-subrepresentations
of $V$ are $0$ and $V$ itself. These are the \emph{elements} or
\emph{primes} of the theory. Indeed, by the Jordan--H{\"o}lder theorem and Maschke's theorem:
every $G$-representation is 
isomorphic to a direct sum of simple
$G$-representations, and the summands that appear are unique up to permutation and isomorphism.
Finally, Schur's lemma says that all $G$-equivariant linear 
maps between two simple real $G$-representations are either invertible or zero. 

Representation theory is a powerful tool to study linear algebra problems with symmetry. For example, representation theorists run the following procedure:
\begin{enumerate}[label=(\roman*)]

\item Assume you have a $G$-equivariant linear map $\phi\colon\rep[V]\to\rep[W]$. This map might be given by some complicated matrix $M_{\phi}$.

\item Now decompose $\rep[V]$ and $\rep[W]$ into simple $G$-representations. This induces a base-change on $M_{\phi}$
and one gets the matrix $N_{\phi}=PM_{\phi}P^{-1}$.

\item Schur's lemma then implies that $N_{\phi}$ is typically much simpler (e.g. it is a diagonal matrix if every
simple representation occurs only once). Now calculations might simplify drastically.

\end{enumerate}
This is illustrated in \autoref{Eq:OverviewComposition}.

The following is our main source of 
$G$-representations (the reason will become clear in 
\autoref{T:PLRepsGEquivariant} below).

\begin{Definition}\label{D:PLRepsPermutationRepresentation}
Let $X$ be a (finite) left $G$-set. The $\R$-vector space $\permrep$ 
of $\R$-valued maps on $X$ 
has a $G$-action: given any function $f$, we obtain a new function $g\acts f$ via
\begin{gather*}
(g\acts f)(x) = f(g^{-1}\acts x).
\end{gather*} 
(The inverse is there to ensure that this is a left action.) 
%The $G$-representation $\permrep$
%has a basis 
%of indicator functions $\{\delta_{x}\mid x\in X\}$ 
%(that we will fix).
%Note that the associated matrices $\rho\colon 
%G\to\GL_{X}(\R)$ merely send $g$ to the permutation matrix associated 
%to the permutation $x\mapsto(g\acts x)$.
We will call representations $\permrep$ 
arising in this way \textit{permutation representations}.
\end{Definition}

Permutation representations give a large class of $G$-representations.

\begin{Example}\label{E:PermutationRepresentation}
Taking $X=G$ in \autoref{D:PLRepsPermutationRepresentation} 
gives the \emph{regular $G$-representation} $\regrep=\permrep[G]$.
\end{Example}

\subsection{Neural networks}\label{SS:OverviewNN}

We assume the reader is somewhat familiar with the classical theory of \emph{neural
networks} as {e.g.} in \cite{GoBeCo-deep-learning} or 
\cite{Ni-deep-learning}. In the simplest incarnation, the picture is:
\begin{gather*}
\raisebox{-2cm}{$\begin{neuralnetwork}[height=4]
\inputlayer[count=3, bias=false, title=Input\\layer]
\hiddenlayer[count=4, bias=false, title=Hidden\\layer 1] \linklayers
\hiddenlayer[count=3, bias=false, title=Hidden\\layer 2] \linklayers
\hiddenlayer[count=3, bias=false, title=Hidden\\layer 3] \linklayers
\outputlayer[count=2, title=Output\\layer] \linklayers
\end{neuralnetwork}$}
.
\end{gather*}
The nodes represent basis vectors $v_{i}$ and the arrows represent linear maps. (For simplicity of exposition we ignore the ``bias'' term as they are typically absent in equivariant neural networks.)
For example, the piece
\begin{gather*}
\raisebox{-2cm}{$\begin{neuralnetwork}[height=4]
\hiddenlayer[count=4, bias=false] 
\hiddenlayer[count=3, bias=false] \linklayers
\end{neuralnetwork}$}
\end{gather*}
represents a linear map $\R^{4}\to\R^{3}$. Labeling each arrow 
with a real number $w_{j}$ called the \emph{weight} expresses the image of a basis vector as a linear combination, using the weights as coefficients, of the basis vectors in the next layer.
This is not the complete story however. To bring 
nonlinearity into 
the neural network, one applies a nonlinear map between each linear 
map, the so-called \emph{activation map} $f$, so that the picture at each layer really is:
\begin{gather*}
\raisebox{-2.0cm}{$\begin{neuralnetwork}[height=4]
\newcommand{\linklabelsA}[4]{$f$}
\inputlayer[count=2, bias=false, title=In]
\hiddenlayer[count=3, bias=false, title=Hidden\\layer]\linklayers
\hiddenlayer[count=3, bias=false, title=Hidden\\layer]
\link[from layer=1, to layer=2, from node=1, to node=1,label=\linklabelsA]
\link[from layer=1, to layer=2, from node=2, to node=2,label=\linklabelsA]
\link[from layer=1, to layer=2, from node=3, to node=3,label=\linklabelsA]
\outputlayer[count=2, bias=false, title=Out] \linklayers
\end{neuralnetwork}$}
.
\end{gather*}
In formulas, the output of such a neural network is
\begin{gather*}
A_{m}f\Big(
A_{m-1}f\big(\dots
f(A_{1}v)
\big)
\Big),
\end{gather*}
where $v$ is the input vector and the $A_{i}$ are the matrices of
weights. As an example, a popular choice for the activation map is 
$f=\relu\colon x\mapsto\max(x,0)$.

One can think of neural networks as providing a
large space of interesting functions, parameterized by the space of
all possible weights. One then trains the neural network to 
approximate a target function via variants of gradient descent over the space of
weights, see for example \cite{GoBeCo-deep-learning,Ni-deep-learning}.

\subsection{Equivariant neural networks: first example}\label{SS:OverviewEquiNN}

The starting point of this paper is that one often is interested in
learning a map that is \emph{equivariant with respect to some
symmetry}. For example:
\begin{enumerate}[label=(\roman*)]

\item Image recognition is often invariant under translation; for example, whether ice cream is present in an image should not depend on the position of the ice cream in the image.

\item Text-to-speech translation is often invariant under 
shift of time; for example, ``ice cream'' should be labeled as such 
regardless of when it appears in an audio recording.

\item Many problems in engineering and applied mathematics involve the analysis of point clouds. Here one is often interested in
qualitative assessments of the collection of points, independent of
ordering. In other words, such problems are invariant under
permuting the points. Thus our learning problem is invariant
under the symmetric group.

\end{enumerate}

In order to explain how to build equivariant neural networks we 
start with a simplistic toy example of a convolutional 
neural networks of \emph{periodic images}. For us such a periodic image 
consists of a choice of a real number for each point on an $n$-by-$n$ 
grid. Interpolating the real numbers as grayscale values we get a picture (for $n=10$):
\begin{gather*}
\begin{tikzpicture}[anchorbase,g/.style={minimum size=0.44cm,fg=#1},
fg/.code={\pgfmathtruncatemacro{\iFill}{100*#1}%
\tikzset{fill=black!\iFill}},ampersand replacement=\&]
\matrix[matrix of nodes,column sep=0pt,row sep=0pt,nodes in empty cells] {
|[g=0.0]| \& |[g=0.6]| \& |[g=0.7]| \& |[g=0.7]| \& |[g=0.0]| \& |[g=0.3]| \& |[g=0.9]| \& |[g=0.8]| \& |[g=0.3]| \& |[g=0.6]|
\\
|[g=0.4]| \& |[g=1.0]| \& |[g=0.9]| \& |[g=0.8]| \& |[g=0.8]| \& |[g=0.3]| \& |[g=0.4]| \& |[g=1.0]| \& |[g=0.9]| \& |[g=0.0]|
\\
|[g=0.9]| \& |[g=0.0]| \& |[g=0.0]| \& |[g=0.8]| \& |[g=0.4]| \& |[g=0.1]| \& |[g=0.6]| \& |[g=0.1]| \& |[g=0.6]| \& |[g=0.0]|
\\
|[g=0.3]| \& |[g=0.6]| \& |[g=0.3]| \& |[g=0.6]| \& |[g=0.0]| \& |[g=0.0]| \& |[g=0.8]| \& |[g=0.1]| \& |[g=0.3]| \& |[g=0.8]|
\\
|[g=0.8]| \& |[g=0.5]| \& |[g=0.1]| \& |[g=0.1]| \& |[g=0.5]| \& |[g=0.7]| \& |[g=0.6]| \& |[g=0.7]| \& |[g=0.2]| \& |[g=0.7]|
\\
|[g=0.2]| \& |[g=0.7]| \& |[g=0.8]| \& |[g=0.8]| \& |[g=0.3]| \& |[g=0.6]| \& |[g=0.6]| \& |[g=0.0]| \& |[g=0.6]| \& |[g=0.3]|
\\
|[g=0.9]| \& |[g=0.4]| \& |[g=0.6]| \& |[g=0.5]| \& |[g=0.0]| \& |[g=0.5]| \& |[g=0.5]| \& |[g=0.8]| \& |[g=0.4]| \& |[g=0.8]|
\\
|[g=0.7]| \& |[g=0.0]| \& |[g=0.6]| \& |[g=0.9]| \& |[g=0.9]| \& |[g=0.1]| \& |[g=0.1]| \& |[g=0.6]| \& |[g=1.0]| \& |[g=0.1]|
\\
|[g=0.9]| \& |[g=0.9]| \& |[g=0.8]| \& |[g=0.9]| \& |[g=0.4]| \& |[g=0.0]| \& |[g=1.0]| \& |[g=0.3]| \& |[g=0.2]| \& |[g=0.5]|
\\
|[g=0.3]| \& |[g=1.0]| \& |[g=0.2]| \& |[g=0.4]| \& |[g=1.0]| \& |[g=0.8]| \& |[g=0.4]| \& |[g=0.0]| \& |[g=0.5]| \& |[g=0.8]|
\\
};
\end{tikzpicture}
.
\end{gather*}
We identify bottom and top, as well as left and right, which makes the picture periodic, so the picture lives on a torus. Let $\cyclic=\Z/n\Z$ 
be the cyclic group of order $n$ and 
$\cyclic^{2}=\cyclic\times\cyclic$. Said 
in a mathematical way, a periodic image is an element of the $\R$-vector space
\begin{gather*}
\rep[V]=\permrep[{\cyclic^{2}}]
\end{gather*}
of maps from the group $\cyclic^{2}$ 
to $\R$. In this model of periodic images,
$\rep[V]$ is a \emph{$\cyclic^{2}$-representation} as in \autoref{SS:OverviewRepTheory}.
Indeed, given $(a,b)\in\cyclic^{2}$ and $f\in\rep[V]$
we obtain a new periodic image by shifting coordinates:
\begin{gather*}
\big((a,b)\acts f\big)(x,y)=f(x+a,y+b).
\end{gather*}
In other words, translation of a periodic images products a new periodic image, e.g.:
\begin{gather*}
\begin{tikzpicture}[anchorbase,g/.style={minimum size=0.44cm,fg=#1},
fg/.code={\pgfmathtruncatemacro{\iFill}{100*#1}%
\tikzset{fill=black!\iFill}},ampersand replacement=\&]
\matrix[matrix of nodes,column sep=0pt,row sep=0pt,nodes in empty cells] {
|[g=0.6]| \& |[g=0.0]| \& |[g=0.6]| \& |[g=0.7]| \& |[g=0.7]| \& |[g=0.0]| \& |[g=0.3]| \& |[g=0.9]| \& |[g=0.8]| \& |[g=0.3]|
\\
|[g=0.0]| \& |[g=0.4]| \& |[g=1.0]| \& |[g=0.9]| \& |[g=0.8]| \& |[g=0.8]| \& |[g=0.3]| \& |[g=0.4]| \& |[g=1.0]| \& |[g=0.9]|
\\
|[g=0.0]| \& |[g=0.9]| \& |[g=0.0]| \& |[g=0.0]| \& |[g=0.8]| \& |[g=0.4]| \& |[g=0.1]| \& |[g=0.6]| \& |[g=0.1]| \& |[g=0.6]|
\\
|[g=0.8]| \& |[g=0.3]| \& |[g=0.6]| \& |[g=0.3]| \& |[g=0.6]| \& |[g=0.0]| \& |[g=0.0]| \& |[g=0.8]| \& |[g=0.1]| \& |[g=0.3]|
\\
|[g=0.7]| \& |[g=0.8]| \& |[g=0.5]| \& |[g=0.1]| \& |[g=0.1]| \& |[g=0.5]| \& |[g=0.7]| \& |[g=0.6]| \& |[g=0.7]| \& |[g=0.2]|
\\
|[g=0.3]| \& |[g=0.2]| \& |[g=0.7]| \& |[g=0.8]| \& |[g=0.8]| \& |[g=0.3]| \& |[g=0.6]| \& |[g=0.6]| \& |[g=0.0]| \& |[g=0.6]|
\\
|[g=0.8]| \& |[g=0.9]| \& |[g=0.4]| \& |[g=0.6]| \& |[g=0.5]| \& |[g=0.0]| \& |[g=0.5]| \& |[g=0.5]| \& |[g=0.8]| \& |[g=0.4]|
\\
|[g=0.1]| \& |[g=0.7]| \& |[g=0.0]| \& |[g=0.6]| \& |[g=0.9]| \& |[g=0.9]| \& |[g=0.1]| \& |[g=0.1]| \& |[g=0.6]| \& |[g=1.0]|
\\
|[g=0.5]| \& |[g=0.9]| \& |[g=0.9]| \& |[g=0.8]| \& |[g=0.9]| \& |[g=0.4]| \& |[g=0.0]| \& |[g=1.0]| \& |[g=0.3]| \& |[g=0.2]|
\\
|[g=0.8]| \& |[g=0.3]| \& |[g=1.0]| \& |[g=0.2]| \& |[g=0.4]| \& |[g=1.0]| \& |[g=0.8]| \& |[g=0.4]| \& |[g=0.0]| \& |[g=0.5]|
\\
};
\end{tikzpicture}
\hspace{-0.2cm}\xleftarrow[\text{by $(1,0)$}]{\text{translation}}\hspace{-0.2cm}
\begin{tikzpicture}[anchorbase,g/.style={minimum size=0.44cm,fg=#1},
fg/.code={\pgfmathtruncatemacro{\iFill}{100*#1}%
\tikzset{fill=black!\iFill}},ampersand replacement=\&]
\matrix[matrix of nodes,column sep=0pt,row sep=0pt,nodes in empty cells] {
|[g=0.0]| \& |[g=0.6]| \& |[g=0.7]| \& |[g=0.7]| \& |[g=0.0]| \& |[g=0.3]| \& |[g=0.9]| \& |[g=0.8]| \& |[g=0.3]| \& |[g=0.6]|
\\
|[g=0.4]| \& |[g=1.0]| \& |[g=0.9]| \& |[g=0.8]| \& |[g=0.8]| \& |[g=0.3]| \& |[g=0.4]| \& |[g=1.0]| \& |[g=0.9]| \& |[g=0.0]|
\\
|[g=0.9]| \& |[g=0.0]| \& |[g=0.0]| \& |[g=0.8]| \& |[g=0.4]| \& |[g=0.1]| \& |[g=0.6]| \& |[g=0.1]| \& |[g=0.6]| \& |[g=0.0]|
\\
|[g=0.3]| \& |[g=0.6]| \& |[g=0.3]| \& |[g=0.6]| \& |[g=0.0]| \& |[g=0.0]| \& |[g=0.8]| \& |[g=0.1]| \& |[g=0.3]| \& |[g=0.8]|
\\
|[g=0.8]| \& |[g=0.5]| \& |[g=0.1]| \& |[g=0.1]| \& |[g=0.5]| \& |[g=0.7]| \& |[g=0.6]| \& |[g=0.7]| \& |[g=0.2]| \& |[g=0.7]|
\\
|[g=0.2]| \& |[g=0.7]| \& |[g=0.8]| \& |[g=0.8]| \& |[g=0.3]| \& |[g=0.6]| \& |[g=0.6]| \& |[g=0.0]| \& |[g=0.6]| \& |[g=0.3]|
\\
|[g=0.9]| \& |[g=0.4]| \& |[g=0.6]| \& |[g=0.5]| \& |[g=0.0]| \& |[g=0.5]| \& |[g=0.5]| \& |[g=0.8]| \& |[g=0.4]| \& |[g=0.8]|
\\
|[g=0.7]| \& |[g=0.0]| \& |[g=0.6]| \& |[g=0.9]| \& |[g=0.9]| \& |[g=0.1]| \& |[g=0.1]| \& |[g=0.6]| \& |[g=1.0]| \& |[g=0.1]|
\\
|[g=0.9]| \& |[g=0.9]| \& |[g=0.8]| \& |[g=0.9]| \& |[g=0.4]| \& |[g=0.0]| \& |[g=1.0]| \& |[g=0.3]| \& |[g=0.2]| \& |[g=0.5]|
\\
|[g=0.3]| \& |[g=1.0]| \& |[g=0.2]| \& |[g=0.4]| \& |[g=1.0]| \& |[g=0.8]| \& |[g=0.4]| \& |[g=0.0]| \& |[g=0.5]| \& |[g=0.8]|
\\
};
\end{tikzpicture}
\hspace{-0.2cm}\xrightarrow[\text{by $(0,1)$}]{\text{translation}}\hspace{-0.2cm}
\begin{tikzpicture}[anchorbase,g/.style={minimum size=0.44cm,fg=#1},
fg/.code={\pgfmathtruncatemacro{\iFill}{100*#1}%
\tikzset{fill=black!\iFill}},ampersand replacement=\&]
\matrix[matrix of nodes,column sep=0pt,row sep=0pt,nodes in empty cells] {
|[g=0.4]| \& |[g=1.0]| \& |[g=0.9]| \& |[g=0.8]| \& |[g=0.8]| \& |[g=0.3]| \& |[g=0.4]| \& |[g=1.0]| \& |[g=0.9]| \& |[g=0.0]|
\\
|[g=0.9]| \& |[g=0.0]| \& |[g=0.0]| \& |[g=0.8]| \& |[g=0.4]| \& |[g=0.1]| \& |[g=0.6]| \& |[g=0.1]| \& |[g=0.6]| \& |[g=0.0]|
\\
|[g=0.3]| \& |[g=0.6]| \& |[g=0.3]| \& |[g=0.6]| \& |[g=0.0]| \& |[g=0.0]| \& |[g=0.8]| \& |[g=0.1]| \& |[g=0.3]| \& |[g=0.8]|
\\
|[g=0.8]| \& |[g=0.5]| \& |[g=0.1]| \& |[g=0.1]| \& |[g=0.5]| \& |[g=0.7]| \& |[g=0.6]| \& |[g=0.7]| \& |[g=0.2]| \& |[g=0.7]|
\\
|[g=0.2]| \& |[g=0.7]| \& |[g=0.8]| \& |[g=0.8]| \& |[g=0.3]| \& |[g=0.6]| \& |[g=0.6]| \& |[g=0.0]| \& |[g=0.6]| \& |[g=0.3]|
\\
|[g=0.9]| \& |[g=0.4]| \& |[g=0.6]| \& |[g=0.5]| \& |[g=0.0]| \& |[g=0.5]| \& |[g=0.5]| \& |[g=0.8]| \& |[g=0.4]| \& |[g=0.8]|
\\
|[g=0.7]| \& |[g=0.0]| \& |[g=0.6]| \& |[g=0.9]| \& |[g=0.9]| \& |[g=0.1]| \& |[g=0.1]| \& |[g=0.6]| \& |[g=1.0]| \& |[g=0.1]|
\\
|[g=0.9]| \& |[g=0.9]| \& |[g=0.8]| \& |[g=0.9]| \& |[g=0.4]| \& |[g=0.0]| \& |[g=1.0]| \& |[g=0.3]| \& |[g=0.2]| \& |[g=0.5]|
\\
|[g=0.3]| \& |[g=1.0]| \& |[g=0.2]| \& |[g=0.4]| \& |[g=1.0]| \& |[g=0.8]| \& |[g=0.4]| \& |[g=0.0]| \& |[g=0.5]| \& |[g=0.8]|
\\
|[g=0.0]| \& |[g=0.6]| \& |[g=0.7]| \& |[g=0.7]| \& |[g=0.0]| \& |[g=0.3]| \& |[g=0.9]| \& |[g=0.8]| \& |[g=0.3]| \& |[g=0.6]|
\\
};
\end{tikzpicture}
.
\end{gather*}
A first crucial observation leading to equivariant neural networks is: whereas the $\R$-vector space of all 
linear maps from $\rep[V]$ to $\rep[V]$ is of
dimension $n^{4}$, the $\R$-vector space of all 
$\cyclic^{2}$-equivariant linear maps is of
dimension $n^{2}$. 

We now consider a typical example of a
$\cyclic^{2}$-equivariant map. For $c=\sum_{g\in\cyclic^{2}}s_{g}g\in\regrep=\permrep[G]$ one gets 
an $\cyclic^{2}$-equivariant map $\rep[V]\to\rep[V]$ by a convolution-type formula:
\begin{gather*}
c\colon\rep[V]\to\rep[V],
f\mapsto\big(
x\mapsto{\textstyle\sum_{g\in\cyclic^{2}}}s_{g}f(x+g)
\big).
\end{gather*}
For example, take $c=\frac{1}{4}\big((1,0)+(0,1)+(-1,0)+(0,-1)\big)$.
Then $c\acts f$ is the periodic image whose value 
at pixel $(a,b)$ is the average of the values 
of its neighbors $(a+1,b)$, $(a,b+1)$, $(a-1,b)$ and $(a,b-1)$. Explicitly (filling the squares with the gray scale values for readability):
\begin{gather*}
\scalebox{0.75}{$\begin{tikzpicture}[anchorbase,g/.style={minimum size=0.6cm,fg=#1},
fg/.code={\pgfmathtruncatemacro{\iFill}{100*#1}%
\tikzset{fill=black!\iFill}},ampersand replacement=\&]
\matrix[matrix of nodes,column sep=0pt,row sep=0pt,nodes in empty cells] {
|[g=0.0]|\scalebox{0.45}{0.0} \& |[g=0.6]|\scalebox{0.45}{0.6} \& |[g=0.7]|\scalebox{0.45}{{\color{white}0.7}} \& |[g=0.7]|\scalebox{0.45}{{\color{white}0.7}} \& |[g=0.0]|\scalebox{0.45}{0.0} \& |[g=0.3]|\scalebox{0.45}{0.3} \& |[g=0.9]|\scalebox{0.45}{{\color{white}0.9}} \& |[g=0.8]|\scalebox{0.45}{{\color{white}0.8}} \& |[g=0.3]|\scalebox{0.45}{0.3} \& |[g=0.6]|\scalebox{0.45}{0.6}
\\
|[g=0.4]|\scalebox{0.45}{0.4} \& |[g=1.0]|\scalebox{0.45}{{\color{white}1.0}} \& |[g=0.9]|\scalebox{0.45}{{\color{white}0.9}} \& |[g=0.8]|\scalebox{0.45}{{\color{white}0.8}} \& |[g=0.8]|\scalebox{0.45}{{\color{white}0.8}} \& |[g=0.3]|\scalebox{0.45}{0.3} \& |[g=0.4]|\scalebox{0.45}{0.4} \& |[g=1.0]|\scalebox{0.45}{{\color{white}1.0}} \& |[g=0.9]|\scalebox{0.45}{{\color{white}0.9}} \& |[g=0.0]|\scalebox{0.45}{0.0}
\\
|[g=0.9]|\scalebox{0.45}{{\color{white}0.9}} \& |[g=0.0]|\scalebox{0.45}{0.0} \& |[g=0.0]|\scalebox{0.45}{0.0} \& |[g=0.8]|\scalebox{0.45}{{\color{white}0.8}} \& |[g=0.4]|\scalebox{0.45}{0.4} \& |[g=0.1]|\scalebox{0.45}{0.1} \& |[g=0.6]|\scalebox{0.45}{0.6} \& |[g=0.1]|\scalebox{0.45}{0.1} \& |[g=0.6]|\scalebox{0.45}{0.6} \& |[g=0.0]|\scalebox{0.45}{0.0}
\\
|[g=0.3]|\scalebox{0.45}{0.3} \& |[g=0.6]|\scalebox{0.45}{0.6} \& |[g=0.3]|\scalebox{0.45}{0.3} \& |[g=0.6]|\scalebox{0.45}{0.6} \& |[g=0.0]|\scalebox{0.45}{0.0} \& |[g=0.0]|\scalebox{0.45}{0.0} \& |[g=0.8]|\scalebox{0.45}{{\color{white}0.8}} \& |[g=0.1]|\scalebox{0.45}{0.1} \& |[g=0.3]|\scalebox{0.45}{0.3} \& |[g=0.8]|\scalebox{0.45}{{\color{white}0.8}}
\\
|[g=0.8]|\scalebox{0.45}{{\color{white}0.8}} \& |[g=0.5]|\scalebox{0.45}{0.5} \& |[g=0.1]|\scalebox{0.45}{0.1} \& |[g=0.1]|\scalebox{0.45}{0.1} \& |[g=0.5]|\scalebox{0.45}{0.5} \& |[g=0.7]|\scalebox{0.45}{{\color{white}0.7}} \& |[g=0.6]|\scalebox{0.45}{0.6} \& |[g=0.7]|\scalebox{0.45}{{\color{white}0.7}} \& |[g=0.2]|\scalebox{0.45}{0.2} \& |[g=0.7]|\scalebox{0.45}{{\color{white}0.7}}
\\
|[g=0.2]|\scalebox{0.45}{0.2} \& |[g=0.7]|\scalebox{0.45}{{\color{white}0.7}} \& |[g=0.8]|\scalebox{0.45}{{\color{white}0.8}} \& |[g=0.8]|\scalebox{0.45}{{\color{white}0.8}} \& |[g=0.3]|\scalebox{0.45}{0.3} \& |[g=0.6]|\scalebox{0.45}{0.6} \& |[g=0.6]|\scalebox{0.45}{0.6} \& |[g=0.0]|\scalebox{0.45}{0.0} \& |[g=0.6]|\scalebox{0.45}{0.6} \& |[g=0.3]|\scalebox{0.45}{0.3}
\\
|[g=0.9]|\scalebox{0.45}{{\color{white}0.9}} \& |[g=0.4]|\scalebox{0.45}{0.4} \& |[g=0.6]|\scalebox{0.45}{0.6} \& |[g=0.5]|\scalebox{0.45}{0.5} \& |[g=0.0]|\scalebox{0.45}{0.0} \& |[g=0.5]|\scalebox{0.45}{0.5} \& |[g=0.5]|\scalebox{0.45}{0.5} \& |[g=0.8]|\scalebox{0.45}{{\color{white}0.8}} \& |[g=0.4]|\scalebox{0.45}{0.4} \& |[g=0.8]|\scalebox{0.45}{{\color{white}0.8}}
\\
|[g=0.7]|\scalebox{0.45}{{\color{white}0.7}} \& |[g=0.0]|\scalebox{0.45}{0.0} \& |[g=0.6]|\scalebox{0.45}{0.6} \& |[g=0.9]|\scalebox{0.45}{{\color{white}0.9}} \& |[g=0.9]|\scalebox{0.45}{{\color{white}0.9}} \& |[g=0.1]|\scalebox{0.45}{0.1} \& |[g=0.1]|\scalebox{0.45}{0.1} \& |[g=0.6]|\scalebox{0.45}{0.6} \& |[g=1.0]|\scalebox{0.45}{{\color{white}1.0}} \& |[g=0.1]|\scalebox{0.45}{0.1}
\\
|[g=0.9]|\scalebox{0.45}{{\color{white}0.9}} \& |[g=0.9]|\scalebox{0.45}{{\color{white}0.9}} \& |[g=0.8]|\scalebox{0.45}{{\color{white}0.8}} \& |[g=0.9]|\scalebox{0.45}{{\color{white}0.9}} \& |[g=0.4]|\scalebox{0.45}{0.4} \& |[g=0.0]|\scalebox{0.45}{0.0} \& |[g=1.0]|\scalebox{0.45}{{\color{white}1.0}} \& |[g=0.3]|\scalebox{0.45}{0.3} \& |[g=0.2]|\scalebox{0.45}{0.2} \& |[g=0.5]|\scalebox{0.45}{0.5}
\\
|[g=0.3]|\scalebox{0.45}{0.3} \& |[g=1.0]|\scalebox{0.45}{{\color{white}1.0}} \& |[g=0.2]|\scalebox{0.45}{0.2} \& |[g=0.4]|\scalebox{0.45}{0.4} \& |[g=1.0]|\scalebox{0.45}{{\color{white}1.0}} \& |[g=0.8]|\scalebox{0.45}{{\color{white}0.8}} \& |[g=0.4]|\scalebox{0.45}{0.4} \& |[g=0.0]|\scalebox{0.45}{0.0} \& |[g=0.5]|\scalebox{0.45}{0.5} \& |[g=0.8]|\scalebox{0.45}{{\color{white}0.8}}
\\
};
\end{tikzpicture}$}
\xrightarrow{\text{action by $c$}}
\scalebox{0.73}{$\begin{tikzpicture}[anchorbase,g/.style={minimum size=0.6cm,fg=#1},
fg/.code={\pgfmathtruncatemacro{\iFill}{100*#1}%
\tikzset{fill=black!\iFill}},ampersand replacement=\&]
\matrix[matrix of nodes,column sep=0pt,row sep=0pt,nodes in empty cells] {
|[g=0.475]|\scalebox{0.45}{0.475} \& |[g=0.675]|\scalebox{0.45}{0.675} \& |[g=0.6]|\scalebox{0.45}{0.6} \& |[g=0.475]|\scalebox{0.45}{0.475} \& |[g=0.7]|\scalebox{0.45}{{\color{white}0.7}} \& |[g=0.5]|\scalebox{0.45}{0.5} \& |[g=0.475]|\scalebox{0.45}{0.475} \& |[g=0.55]|\scalebox{0.45}{0.55} \& |[g=0.7]|\scalebox{0.45}{{\color{white}0.7}} \& |[g=0.275]|\scalebox{0.45}{0.275}
\\
|[g=0.475]|\scalebox{0.45}{0.475} \& |[g=0.475]|\scalebox{0.45}{0.475} \& |[g=0.625]|\scalebox{0.45}{0.625} \& |[g=0.8]|\scalebox{0.45}{{\color{white}0.8}} \& |[g=0.375]|\scalebox{0.45}{0.375} \& |[g=0.4]|\scalebox{0.45}{0.4} \& |[g=0.7]|\scalebox{0.45}{{\color{white}0.7}} \& |[g=0.55]|\scalebox{0.45}{0.55} \& |[g=0.475]|\scalebox{0.45}{0.475} \& |[g=0.475]|\scalebox{0.45}{0.475}
\\
|[g=0.175]|\scalebox{0.45}{0.175} \& |[g=0.625]|\scalebox{0.45}{0.625} \& |[g=0.5]|\scalebox{0.45}{0.5} \& |[g=0.45]|\scalebox{0.45}{0.45} \& |[g=0.425]|\scalebox{0.45}{0.425} \& |[g=0.325]|\scalebox{0.45}{0.325} \& |[g=0.35]|\scalebox{0.45}{0.35} \& |[g=0.575]|\scalebox{0.45}{0.575} \& |[g=0.325]|\scalebox{0.45}{0.325} \& |[g=0.575]|\scalebox{0.45}{0.575}
\\
|[g=0.775]|\scalebox{0.45}{{\color{white}0.775}} \& |[g=0.275]|\scalebox{0.45}{0.275} \& |[g=0.325]|\scalebox{0.45}{0.325} \& |[g=0.3]|\scalebox{0.45}{0.3} \& |[g=0.375]|\scalebox{0.45}{0.375} \& |[g=0.4]|\scalebox{0.45}{0.4} \& |[g=0.325]|\scalebox{0.45}{0.325} \& |[g=0.475]|\scalebox{0.45}{0.475} \& |[g=0.425]|\scalebox{0.45}{0.425} \& |[g=0.325]|\scalebox{0.45}{0.325}
\\
|[g=0.425]|\scalebox{0.45}{0.425} \& |[g=0.55]|\scalebox{0.45}{0.55} \& |[g=0.425]|\scalebox{0.45}{0.425} \& |[g=0.5]|\scalebox{0.45}{0.5} \& |[g=0.275]|\scalebox{0.45}{0.275} \& |[g=0.425]|\scalebox{0.45}{0.425} \& |[g=0.7]|\scalebox{0.45}{{\color{white}0.7}} \& |[g=0.225]|\scalebox{0.45}{0.225} \& |[g=0.575]|\scalebox{0.45}{0.575} \& |[g=0.525]|\scalebox{0.45}{0.525}
\\
|[g=0.675]|\scalebox{0.45}{0.675} \& |[g=0.475]|\scalebox{0.45}{0.475} \& |[g=0.55]|\scalebox{0.45}{0.55} \& |[g=0.425]|\scalebox{0.45}{0.425} \& |[g=0.475]|\scalebox{0.45}{0.475} \& |[g=0.525]|\scalebox{0.45}{0.525} \& |[g=0.425]|\scalebox{0.45}{0.425} \& |[g=0.675]|\scalebox{0.45}{0.675} \& |[g=0.225]|\scalebox{0.45}{0.225} \& |[g=0.575]|\scalebox{0.45}{0.575}
\\
|[g=0.525]|\scalebox{0.45}{0.525} \& |[g=0.55]|\scalebox{0.45}{0.55} \& |[g=0.575]|\scalebox{0.45}{0.575} \& |[g=0.575]|\scalebox{0.45}{0.575} \& |[g=0.55]|\scalebox{0.45}{0.55} \& |[g=0.3]|\scalebox{0.45}{0.3} \& |[g=0.5]|\scalebox{0.45}{0.5} \& |[g=0.375]|\scalebox{0.45}{0.375} \& |[g=0.8]|\scalebox{0.45}{{\color{white}0.8}} \& |[g=0.425]|\scalebox{0.45}{0.425}
\\
|[g=0.475]|\scalebox{0.45}{0.475} \& |[g=0.65]|\scalebox{0.45}{0.65} \& |[g=0.575]|\scalebox{0.45}{0.575} \& |[g=0.725]|\scalebox{0.45}{{\color{white}0.725}} \& |[g=0.35]|\scalebox{0.45}{0.35} \& |[g=0.375]|\scalebox{0.45}{0.375} \& |[g=0.55]|\scalebox{0.45}{0.55} \& |[g=0.55]|\scalebox{0.45}{0.55} \& |[g=0.325]|\scalebox{0.45}{0.325} \& |[g=0.75]|\scalebox{0.45}{{\color{white}0.75}}
\\
|[g=0.6]|\scalebox{0.45}{0.6} \& |[g=0.675]|\scalebox{0.45}{0.675} \& |[g=0.65]|\scalebox{0.45}{0.65} \& |[g=0.625]|\scalebox{0.45}{0.625} \& |[g=0.7]|\scalebox{0.45}{{\color{white}0.7}} \& |[g=0.575]|\scalebox{0.45}{0.575} \& |[g=0.2]|\scalebox{0.45}{0.2} \& |[g=0.45]|\scalebox{0.45}{0.45} \& |[g=0.575]|\scalebox{0.45}{0.575} \& |[g=0.5]|\scalebox{0.45}{0.5}
\\
|[g=0.675]|\scalebox{0.45}{0.675} \& |[g=0.5]|\scalebox{0.45}{0.5} \& |[g=0.725]|\scalebox{0.45}{{\color{white}0.725}} \& |[g=0.7]|\scalebox{0.45}{{\color{white}0.7}} \& |[g=0.4]|\scalebox{0.45}{0.4} \& |[g=0.425]|\scalebox{0.45}{0.425} \& |[g=0.675]|\scalebox{0.45}{0.675} \& |[g=0.5]|\scalebox{0.45}{0.5} \& |[g=0.325]|\scalebox{0.45}{0.325} \& |[g=0.475]|\scalebox{0.45}{0.475}
\\
};
\end{tikzpicture}$}
,\\
\begin{gathered}
\text{zooming}
\\
\text{into}
\\
\text{the node}
\\
\text{at $2$-$2$}
\end{gathered}\colon
\scalebox{1.5}{$\begin{tikzpicture}[anchorbase,g/.style={minimum size=0.6cm,fg=#1},
fg/.code={\pgfmathtruncatemacro{\iFill}{100*#1}%
\tikzset{fill=black!\iFill}},ampersand replacement=\&]
\matrix[matrix of nodes,column sep=0pt,row sep=0pt,nodes in empty cells] {
|[g=0.0]|\scalebox{0.45}{0.0} \& |[g=0.6]|\scalebox{0.45}{0.6} \& |[g=0.7]|\scalebox{0.45}{{\color{white}0.7}}
\\
|[g=0.4]|\scalebox{0.45}{0.4} \& |[g=1.0]|\scalebox{0.45}{{\color{white}1.0}} \& |[g=0.9]|\scalebox{0.45}{{\color{white}0.9}}
\\
|[g=0.9]|\scalebox{0.45}{{\color{white}0.9}} \& |[g=0.0]|\scalebox{0.45}{0.0} \& |[g=0.0]|\scalebox{0.45}{0.0} 
\\
};
\end{tikzpicture}$}
\xrightarrow{\text{action by $c$}}
\scalebox{1.5}{$\begin{tikzpicture}[anchorbase,g/.style={minimum size=0.6cm,fg=#1},
fg/.code={\pgfmathtruncatemacro{\iFill}{100*#1}%
\tikzset{fill=black!\iFill}},ampersand replacement=\&]
\matrix[matrix of nodes,column sep=0pt,row sep=0pt,nodes in empty cells] {
|[g=0.475]|\scalebox{0.45}{0.475} \& |[g=0.675]|\scalebox{0.45}{0.675} \& |[g=0.6]|\scalebox{0.45}{0.6}
\\
|[g=0.475]|\scalebox{0.45}{0.475} \& |[g=0.475]|\scalebox{0.45}{0.475} \& |[g=0.625]|\scalebox{0.45}{0.625}
\\
|[g=0.175]|\scalebox{0.45}{0.175} \& |[g=0.625]|\scalebox{0.45}{0.625} \& |[g=0.5]|\scalebox{0.45}{0.5}
\\
};
\end{tikzpicture}$}
\text{ where $0.475=\frac{1}{4}(0.6+0.9+0.0+0.4)$}
.
\end{gather*}
More generally, convolution for different $c$ provides examples
of maps used widely in image processing, see 
for example \cite[Section I.2]{Ja}.

Now we can define a \emph{$\cyclic^{2}$-equivariant neural network} in this case. Its structure is as follows:
\begin{gather*}
\raisebox{-2cm}{$\begin{neuralnetwork}[height=4]
\newcommand{\vv}[2]{$\rep[V]$}
\newcommand{\ww}[2]{$\rep[W]$}
\newcommand{\xx}[4]{conv.}
\inputlayer[count=1, bias=false, title=Input\\layer,text=\vv]
\hiddenlayer[count=3, bias=false, title=Hidden\\layer 1,text=\vv]
\link[from layer=0, to layer=1, from node=1, to node=1,label=\xx]
\link[from layer=0, to layer=1, from node=1, to node=2]
\link[from layer=0, to layer=1, from node=1, to node=3]
\hiddenlayer[count=2, bias=false, title=Hidden\\layer 2,text=\vv] \linklayers
\outputlayer[count=1, title=Output\\layer,text=\ww] \linklayers
\end{neuralnetwork}$}
.
\end{gather*}
Each arrow is a convolution $c_{ij}^{k}$ with $k$ denoting the layer and $i$ and $j$ being the usual 
matrix indexes. Moreover, $\rep[W]$ is typically $\R$ or
$\rep[V]$. The above picture is a (simplified) picture
of a \emph{convolutional neural network}, which are of great
importance in the machine learning literature, see for example
\cite[Section II.9]{GoBeCo-deep-learning}.

The main conceptual points to note with respect to this construction are:
\begin{enumerate}[label=(\roman*)]

\item The structure of the neural network
forces the resulting map $\rep[V]\to\rep[W]$ to be equivariant.

\item The space of all weights is much smaller 
than for traditional (``fully connected'') neural networks. In practice this 
means that equivariant neural networks can handle much larger examples
than ``vanilla'' neural networks. (This phenomenon is often termed \emph{weight
sharing} in the machine learning literature.)

\end{enumerate}
We stress that the picture above implicitly 
includes the activation map, with our favorite choice being ReLU.
This means that the compositions in neural networks are actually piecewise linear
maps.
Thus, in order to apply our second main observation to equivariant neural networks -- reduction of problems by decomposition into simples -- the study of \emph{piecewise linear representation theory} comes up naturally.
Before going there we will recall piecewise linear maps next.

\subsection{Piecewise linear and affine maps}\label{SS:PLRepsMaps}

Recall that a \emph{piecewise linear map} $f\colon\R\to\R$ is 
a map whose graph is composed of straight-line segments.
The generalization of this to arbitrary $\R$-vector spaces 
needs a little preparation.

\begin{Definition} Let $V$ be an $\R$-vector space.
A subset $A \subset V$ is \emph{convex polyhedral} if it is 
the intersection of finitely many closed half-spaces in $V$, and has
open interior. A \emph{polyhedral covering} $\alcove$ of a subset $P \subset V$ is a finite collection 
of convex polyhedral subsets $(A_{1},\dots,A_{m})$ such that 
$\bigcup_{i=1}^{m}A_{i}=P$.
\end{Definition}

Recall that a map $f\colon A \to\R^{n}$ for $A \subset \R^n$ is \emph{affine linear} if it is
of the form $f(x)=M(x)+b$ for some linear map $M\colon\R^{m}\to\R^{n}$
and bias $b\in\R^{n}$. 

\begin{Definition}\label{D:PLRepsMaps}
A map $f\colon V\to W$ between $\R$-vector spaces $V$ and $W$ is 
called \emph{piecewise linear} if there exists a polyhedral covering 
$\alcove=(A_{1},\dots,A_{m})$ of $V$ such
that the restriction of each $f$ to $A_{i}$ is affine linear.
\end{Definition}

Piecewise linear maps are the main players in this paper.

\begin{Lemma}\label{L:PLRepsMaps}
We have the following.
\begin{enumerate}

\item Piecewise linear maps are continuous.

\item If $f\colon U\to V$, $g\colon V\to W$ and $h\colon X\to Y$ 
are piecewise linear, then so are $g\circ f$ and $f\oplus h$.

\item The inverse of an invertible piecewise linear map is piecewise linear.

\end{enumerate}
\end{Lemma}

\begin{proof}
Easy and omitted.
\end{proof}

It follows that linear maps are piecewise linear, and composition preserves 
the property of being piecewise linear, see \autoref{L:PLRepsMaps}. Here are more examples:

\begin{Example}\label{E:PLRepsMaps}
Key examples for this paper are the
\emph{ReLU (rectified linear unit)}
and the \emph{absolute value} maps given by
\begin{gather*}
\begin{gathered}
\relu\colon\R\to\R,
\\
x\mapsto
\max(x,0)
\end{gathered}\leftrightsquigarrow
\begin{tikzpicture}[anchorbase,scale=1]
\node at (0,0) {\includegraphics[height=2.5cm]{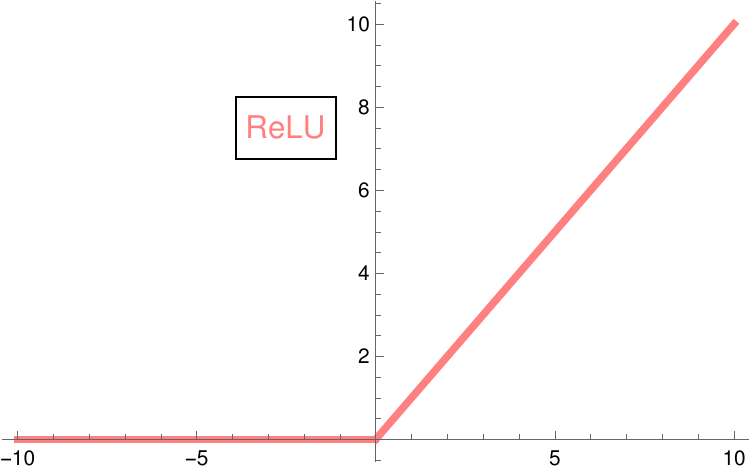}};
\end{tikzpicture}
,
\quad
\begin{gathered}
\absf\colon\R\to\R,
\\
x\mapsto
\max(x,-x)
\end{gathered}
\leftrightsquigarrow
\begin{tikzpicture}[anchorbase,scale=1]
\node at (0,0) {\includegraphics[height=2.5cm]{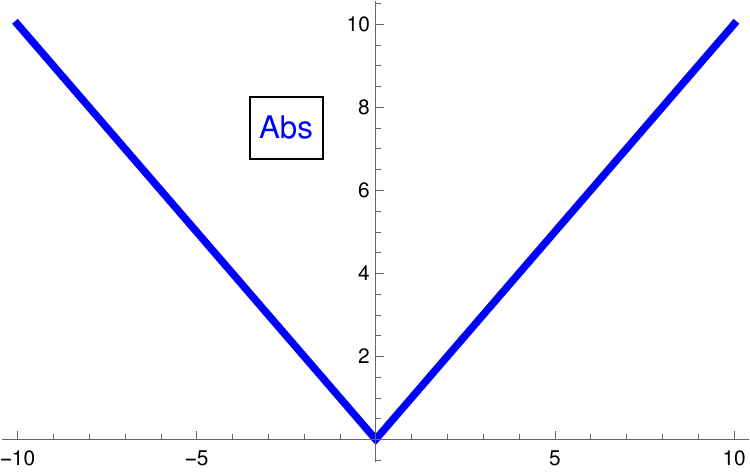}};
\end{tikzpicture}
.
\end{gather*}
We extend these piecewise linear maps componentwise to maps 
$\R^{n}\to\R^{n}$, and 
\autoref{L:PLRepsMaps} shows that these 
extension are still piecewise linear.
\end{Example}

\begin{Example}\label{E:PLRepsMapsTwo}
In order to provide more intuition on piecewise linear maps, we give a two-dimensional example. Let $f\colon\R^{2}\to\R$ be the map $f(x,y)=|\mathrm{sgn}(x)x+\mathrm{sgn}(y)y|$, where $\mathrm{sgn}$ is the sign of a real number. Plotting this map in $\R^{3}$ or as 
a contour (that is, using level sets) in $\R^{2}$ gives
\begin{gather*}
\begin{tikzpicture}[anchorbase,scale=1]
\node at (0,0) {\includegraphics[height=2.5cm]{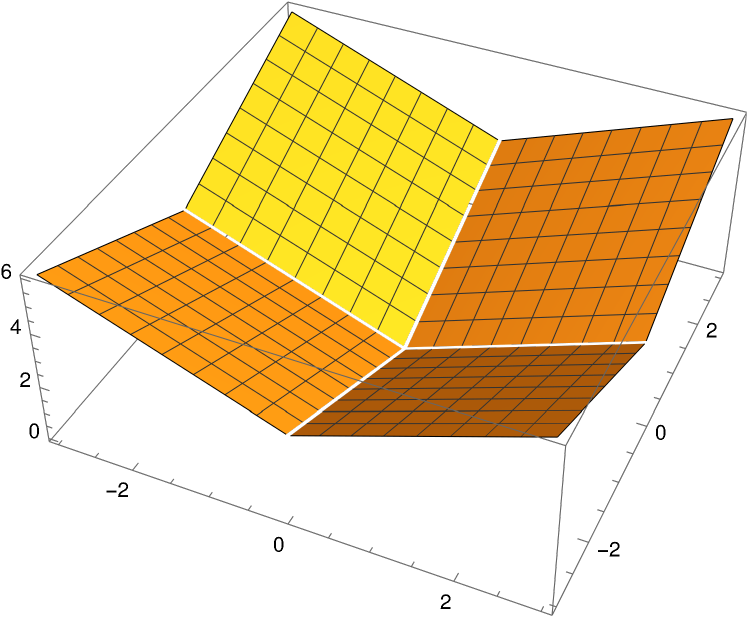}};
\end{tikzpicture}
,\quad
\begin{tikzpicture}[anchorbase,scale=1]
\node at (0,0) {\includegraphics[height=2.5cm]{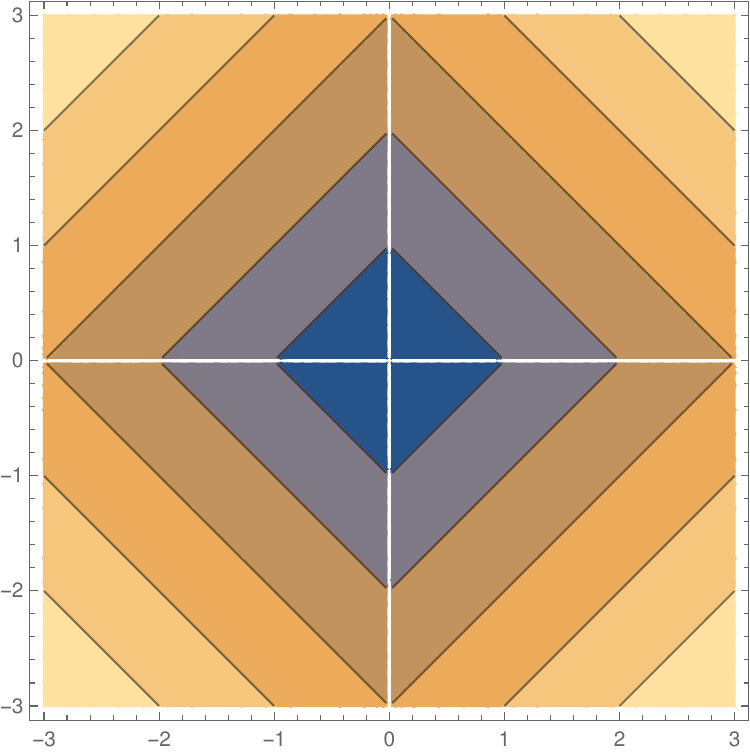}};
\end{tikzpicture}
.
\end{gather*}
Another example is
$g\colon\R^{2}\to\R$ given by
\begin{gather*}
\begin{tikzpicture}[anchorbase,scale=1]
\node at (0,0) {\includegraphics[height=2.5cm]{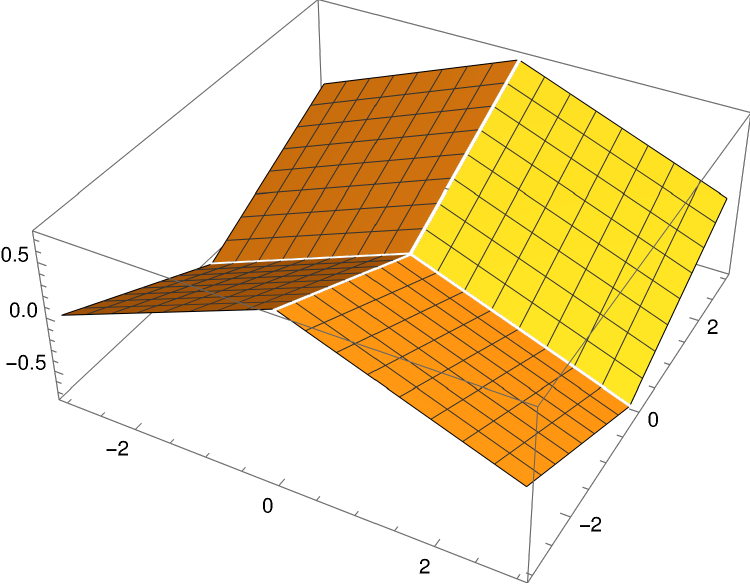}};
\end{tikzpicture}
,\quad
\begin{tikzpicture}[anchorbase,scale=1]
\node at (0,0) {\includegraphics[height=2.5cm]{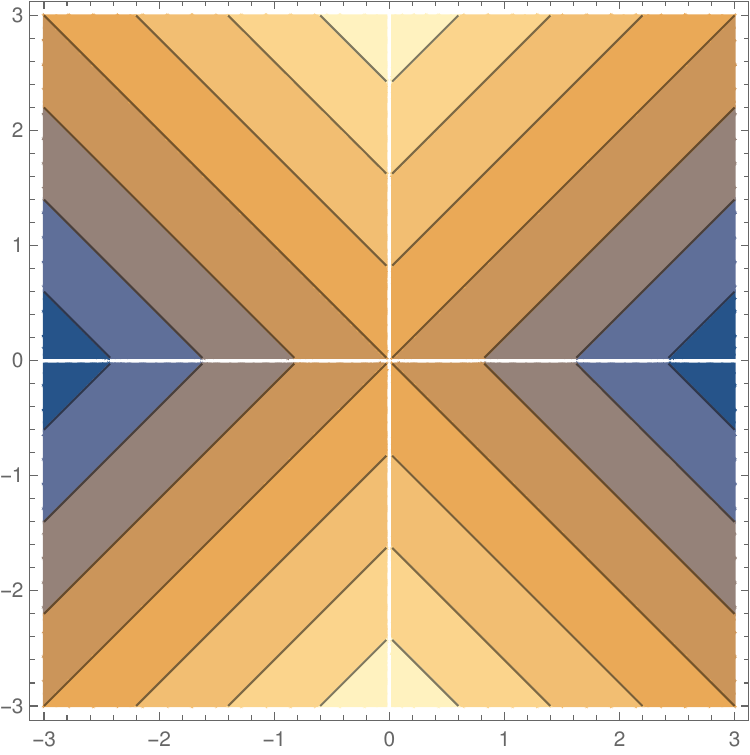}};
\end{tikzpicture}
.
\end{gather*}
(The coordinate description of $g$ is not difficult to obtain.)
We will revisit these maps in \autoref{SS:CyclicHyperplanes}.
\end{Example}

Suppose $V$ is $n$-dimensional. Then any affine linear map on $V$ is
determined by its values on any $n+1$ points which do not lie on a
hyperplane. This implies the following:

\begin{Lemma}\label{L:PLRepsCalculateMaps}
A piecewise linear map $f\colon V\to W$ with polyhedral covering 
$\alcove=(A_{1},\dots,A_{m})$ is determined on each $A_i$ by
evaluating $f$ on any ($n+1$)-vectors in $A_i$ which do not lie in a hyperplane.
\end{Lemma}

\begin{Notation}\label{N:PLReps}
We write $\plmap\colon\R\to\R$ to denote any \emph{nonlinear} piecewise linear map, 
for example, any of the maps in \autoref{E:PLRepsMaps}. In contrast, we 
use $f$ for any piecewise linear map, including the linear ones.
\end{Notation}

We will often consider \emph{based} $G$-representations $(\rep,B)$, that is, $G$-representations $\rep$ together with a choice of basis $B$. We often simply write $\rep$ instead of $(\rep,B)$ if the choice of basis is clear. This is motivated by the following.
Given any vector space $V$ with fixed basis and $\plmap$ as above, we get a piecewise linear endomorphism of $V$ by evaluating $\plmap$ componentwise:

\begin{Definition}\label{D:PLRepsReLUDef}
Fix a basis $B=\{b_{1},\dots,b_{n}\}\subseteq V$ of the $\R$-vector space $V$. Define the piecewise linear map $\plmap_{B}\colon V\to V$ by $\plmap_{B}\big(\sum_{i}\lambda_{i}b_{i}\big):=\sum_{i}\plmap(\lambda_{i})b_{i}$. (We often omit the subscript $B$
if no confusion can arise.)
\end{Definition}

The map $\relu_{B}$ is insensitive to the order of the basis, and the scaling of each basis element by a positive number.
Conversely, given only the map $\relu_{B}\colon V\to V$, we can
determine $B$, up to permutation and scaling by positive numbers, by looking at the convex polyhedron on which $\plmap_{B}$ restricts to
the identity map.

\subsection{Equivariant neural networks}\label{SS:OverviewEquiNNTwo}
In this section we will 
define what we mean by equivariant neural networks. The definition is modeled 
following the example given in
\autoref{SS:OverviewEquiNN}. As discussed in the introduction, equivariant neural
networks are standard in machine learning. Standard references
include \cite{cohen2016group}, \cite[\S 5.2]{bronstein2021geometric} and
\cite{gerken2023geometric}. 

Let $G$ be a finite group. Recall that we consider 
permutation representations $\permrep[X]$ of a finite group $G$ as recalled 
in \autoref{D:PLRepsPermutationRepresentation}. As a based 
$G$-representation we fix the set of indicator functions as a basis.

\begin{Definition} \label{D:OverviewEquiNNTwo}
An \emph{equivariant neural network} is a neural
network in which each layer is a direct sum of permutation
representations, and all linear maps are $G$-equivariant. In
pictures:
\begin{gather*}
\raisebox{-2.0cm}{$\begin{neuralnetwork}[height=4]
\newcommand{\ww}[2]{\scalebox{0.7}{perm}}
\inputlayer[count=2, bias=false,text=\ww]
\hiddenlayer[count=3, bias=false,text=\ww]
\link[from layer=0, to layer=1, from node=1, to node=1]
\link[from layer=0, to layer=1, from node=1, to node=2]
\link[from layer=0, to layer=1, from node=1, to node=3]
\link[from layer=0, to layer=1, from node=2, to node=1]
\link[from layer=0, to layer=1, from node=2, to node=2]
\link[from layer=0, to layer=1, from node=2, to node=3]
\hiddenlayer[count=2, bias=false,text=\ww]
\link[from layer=1, to layer=2, from node=1, to node=1]
\link[from layer=1, to layer=2, from node=1, to node=2]
\link[from layer=1, to layer=2, from node=2, to node=1]
\link[from layer=1, to layer=2, from node=2, to node=2]
\link[from layer=1, to layer=2, from node=3, to node=1]
\link[from layer=1, to layer=2, from node=3, to node=2]
\outputlayer[count=1, bias=false,text=\ww]
\link[from layer=2, to layer=3, from node=1, to node=1]
\link[from layer=2, to layer=3, from node=2, to node=1]
\end{neuralnetwork}$}
\end{gather*}
(Here, the green, blue and red dots indicate the input, hidden and
output layers respectively, and `perm' indicates a permutation
representation, which are not assumed to be equal. As in vanilla
neural networks, we assume a fixed activation function throughout,
which is applied componentwise in each hidden layer.)
\end{Definition}

\begin{Remark}\label{R:OverviewEquiNNTwo}
We stay with finite groups in this paper, 
but \autoref{D:OverviewEquiNNTwo} is also interesting
for other objects as well, e.g. for Lie groups, see
\cite{bronstein2021geometric}.
\end{Remark}

When $G$ is the trivial group, the notion of an equivariant neural
network reduces to that of a vanilla neural network (without
biases). In that setting, the weights emerge as classifying linear
maps
\[
\R \to \R.
\]
In the setting of equivariant neural networks
it is essential to know the space of $G$-equivariant maps
\[
\permrep[X] \to \permrep[Y]
\]
where $X$ and $Y$ are permutation representations. A useful
description of these ``$G$-equivariant weights'' is provided by the
following lemma (which is well-known, both in the representation theory
and machine learning literature, see e.g. \cite[\S 10, Exercise 2]{CR}):

\begin{Lemma}\label{L:OverviewEquiNNTwo}
For finite $G$-sets $X,Y$ we have
\begin{gather*}
\Hom_{G}\big(\permrep[X],\permrep[Y]\big)
\cong\permrepg[{X\times Y}]
\end{gather*}
where $\permrepg[{X\times Y}]$ denotes $G$-invariant functions $X
\times Y \to \R$.
\end{Lemma}

Before we prove the lemma, we want to explain why it is useful. Given
any $G$-orbit on $X \times Y$, its indicator function provides an
element of $\permrepg[{X\times Y}]$ which is manifestly
$G$-invariant. Moreover, one checks easily that such indicator
functions provide a basis of $\permrepg[{X\times Y}]$. Thus,
$G$-equivariant weights are classified by something very concrete:
$G$-orbits on $X \times Y$.

We now give a proof of \autoref{L:OverviewEquiNNTwo}:

\begin{proof} We define maps
\[
\Hom_{G}\big(\permrep[X],\permrep[Y]\big)\ni \alpha \mapsto
\check{\alpha} \in \permrepg[{X\times Y}]
\]
and
\[
\permrepg[{X\times Y}] \ni \beta \mapsto \hat{\beta} \in
\Hom_{G}\big(\permrep[X],\permrep[Y]\big)
\]
and check that they are mutually inverse.

\emph{The map $\beta \mapsto \hat{\beta}$:} Given $\beta \in
\permrepg[{X\times Y}] $ consider the ``convolution operator''
\begin{gather*}
\permrepg[{X\times Y}]\to
\Hom \big(\permrep[X],\permrep[Y]\big)
,
\\
\beta \mapsto(f\mapsto\hat{\beta}(f))\text{ where }
\hat{\beta}(f)(y) = \sum_{x \in X} f(x) \beta(x,y).
\end{gather*}
One checks easily that $\hat{\beta}(g \acts f) = g \acts \hat{\beta}(f)$,
indeed
\[
\hat{\beta}(g \acts f)(y) = \sum_{x \in X} (gf)(x) \beta(x,y) = \sum_{x \in X}
f(g^{-1}x) \beta(x,y) = \sum_{x \in X} f(x) \beta(gx,y) = \sum_{x \in
X} f(x) \beta(x,g^{-1}y) = g \cdot (\hat{\beta}(f)) (y).
\]
Thus the image of $\hat{\beta}$ lies in
$\Hom_{G}\big(\permrep[X],\permrep[Y]\big)$.

\emph{The map $\alpha \mapsto \check{\alpha}$:} For any $x \in X$, let
$\delta_x$ denote the corresponding $\delta$-function. Note that $g
\cdot \delta_x = \delta_{gx}$, as follows 
immediately from \autoref{D:PLRepsPermutationRepresentation}. Now given $\alpha \in
\Hom_{G}\big(\permrep[X],\permrep[Y]\big)$ consider the function
$\check{\alpha}$ on $X \times Y$ defined by 
\[
\check{\alpha}(x,y) = \alpha(\delta_x)(y).
\]
One checks easily that $\check{\alpha}(gx,gy) = \check{\alpha}(x,y)$,
indeed
\[
\check{\alpha}(gx,gy) = \alpha(\delta_{gx})(gy) = \alpha(g \cdot
\delta_x) (gy) = (g \cdot \alpha(\delta_x))(gy) =
\alpha(\delta_x)(g^{-1}gy) = \alpha(\delta_x)(y) = \check{\alpha}(x,y).
\]
Thus $\check{\alpha}$ lies in $\permrepg[{X\times Y}]$.

Finally, we check that $\alpha \mapsto \check{\alpha}$ and $\beta
\mapsto \hat{\beta}$ are mutually inverse bijections. In order to show that $\hat{\check{\alpha}} = \alpha$ it is enough to
that both sides are equal on delta functions, and this is true:
\[
\hat{\check{\alpha}}(\delta_x)(y) = \sum_{z \in X} \delta_x(z)
\check{\alpha}(x,y) = \check{\alpha}(x,y) = \alpha(\delta_x)(y).
\]
Finally, we check that $\beta = \check{\hat{\beta}}$ by checking
equality on every point $(x,y) \in X \times Y$:
\[
\check{\hat{\beta}}(x,y) = \hat{\beta}(\delta_x)(y) = \sum_{z \in X}
\delta_x(z) \beta(z,y) = \beta(x,y). \qedhere
\] 
\end{proof}

\begin{Remark} 
With a little more theory, one
can prove \autoref{L:OverviewEquiNNTwo} slickly, using the
fact that $\permrep[X]$ is self-dual, for any $G$-set $X$:
\begin{gather*}
\Hom_{G}\big(\permrep[X],\permrep[Y]\big)
\cong
\Hom_{G}\big(\permrep[X],\R\otimes\permrep[Y]\big)
\cong
\Hom_{G}\big(\permrep[X]\otimes\permrep[Y]^{\ast},\R\big)
\\
\cong
\Hom_{G}\big(\permrep[X]\otimes\permrep[Y],\R\big)
\cong
\Hom_{G}\big(\permrep[X\times Y],\R\big)
\cong
\permrepg[{X\times Y}].
\end{gather*}
We prefer the more complicated proof that we have given above, as it
gives a concrete form of the bijection.
\end{Remark}

\begin{Example} A blueprint example of an equivariant neural network
is given by point clouds, i.e. collections of $n$ indistinguishable
points in $\R^d$ for some $n, d \in \N$. In this case our group $G$
is $S_n$, the symmetric group on $n$ letters, and our input layer is
given by $(\R^d)^n = (\R^n)^d$ which we can view as $d$ copies of
the permutation module $\permrep[\{1,\dots,n\}]$.  Writing
$\mathtt{n}=\permrep[\{1,\dots,n\}]$, a typical equivariant 
neural network looks like
\begin{gather*}
\raisebox{-2.0cm}{$\begin{neuralnetwork}[height=4]
\newcommand{\linklabelsA}[4]{equi.}
\newcommand{\ww}[2]{n}
\newcommand{\www}[2]{1}
\inputlayer[count=3, bias=false,text=\ww]
\hiddenlayer[count=4, bias=false,text=\ww]
\link[from layer=0, to layer=1, from node=1, to node=1,label=\linklabelsA]
\link[from layer=0, to layer=1, from node=1, to node=2]
\link[from layer=0, to layer=1, from node=1, to node=3]
\link[from layer=0, to layer=1, from node=1, to node=4]
\link[from layer=0, to layer=1, from node=1, to node=2]
\link[from layer=0, to layer=1, from node=2, to node=1]
\link[from layer=0, to layer=1, from node=2, to node=2]
\link[from layer=0, to layer=1, from node=2, to node=3]
\link[from layer=0, to layer=1, from node=2, to node=4]
\link[from layer=0, to layer=1, from node=3, to node=1]
\link[from layer=0, to layer=1, from node=3, to node=2]
\link[from layer=0, to layer=1, from node=3, to node=3]
\link[from layer=0, to layer=1, from node=3, to node=4]
\hiddenlayer[count=2, bias=false,text=\ww]
\link[from layer=1, to layer=2, from node=1, to node=1]
\link[from layer=1, to layer=2, from node=1, to node=2]
\link[from layer=1, to layer=2, from node=2, to node=1]
\link[from layer=1, to layer=2, from node=2, to node=2]
\link[from layer=1, to layer=2, from node=3, to node=1]
\link[from layer=1, to layer=2, from node=3, to node=2]
\link[from layer=1, to layer=2, from node=4, to node=1]
\link[from layer=1, to layer=2, from node=4, to node=2]
\outputlayer[count=1, bias=false,text=\www]
\link[from layer=2, to layer=3, from node=1, to node=1]
\link[from layer=2, to layer=3, from node=2, to node=1]
\end{neuralnetwork}$}
\end{gather*}
(Here $d=3$ and our network has $2$ hidden layers.) As explained above, the linear maps above should be $S_n$-equivariant
and we can use \autoref{L:OverviewEquiNNTwo} to quickly determine
the possible maps. Indeed, by the lemma,
\[
\Hom_{G}(\permrep[\{1,\dots,n\}], \permrep[\{1,\dots,n\}]) =
\permrepg[\{1,\dots,n\} \times \{1,\dots,n\}]
\]
and $G = S_n$ has two orbits on $\{1,\dots,n\} \times \{1,\dots,n\}$
given by the diagonal and its complement. Thus there is a
2-dimensional space of equivariant maps $\mathtt{n} \to \mathtt{n}$,
and this holds independently of $n$.

(Equivariant neural networks for
$S_n$ of this form go by the name \emph{deep sets} in the machine
learning literature \cite{zaheer2017deep}, and are widely used. The reader is referred to
\cite{godfrey2023fast} for a recent interesting application of the
partition algebra to understanding permutation equivariant layers.)
\end{Example}

\begin{Example}\label{E:OverviewEquiNNTwo}
Recall that in \autoref{SS:OverviewEquiNN} 
we had $\cyclic[n]^{2}$ acting on ``gray-scale images'' $\permrep[{\cyclic^{2}}]$ by translation. In this case the number of 
$\cyclic[n]^{2}$-orbits on $\cyclic[n]^{2}\times\cyclic[n]^{2}$ is $n^{2}$. For example, for $n=2$ the $\cyclic[n]^{2}$-orbit that includes 
$\big((0,0),(0,0)\big)$ is 
\begin{gather*}
\Big\{
\big((0,0),(0,0)\big),
\big((1,0),(1,0)\big),
\big((0,1),(0,1)\big),
\big((1,1),(1,1)\big)
\Big\}
\end{gather*}
Similarly for all other elements of 
$\cyclic[n]^{2}\times\cyclic[n]^{2}$.
Thus, $\dim_{\R}\End_{\cyclic[n]^{2}}(\permrep[{\cyclic^{2}}])=n^{2}$ by \autoref{L:OverviewEquiNNTwo}.
\end{Example}

The reader might wonder why we use permutation representations 
in \autoref{D:OverviewEquiNNTwo}. We will now see that there 
is essentially \emph{no other choice}. Recall that a
function $f : \R \to \R$ is \emph{odd} if $f(-x) = -f(x)$ for all $x
\in \R$.

\begin{Theorem}[\textbf{Why permutation representations?}]\label{T:PLRepsGEquivariant}
Let $\rep$ be a $G$-representation and assume that $\plmap$ is not odd. Then
there exists a basis $B=\{b_{1},\dots,b_{n}\}$ such that
$\plmap_{B}\colon\rep\to\rep$ is $G$-equivariant if  and only if
$\rep$ is equivalent to a permutation representation.
\end{Theorem}

It is easy to see that $\plmap_B$ is equivariant if $V$ is
permutation. The content of the theorem is the other implication. We
give two proofs, one when $\plmap$ is assumed to be $\relu$, and one
in the general case.

\begin{proof}[Proof (when our activation function is $\relu$)]
Consider the simplicial cone
\[ C=  \{\sum_{i=1}^{n}s_{i}b_{i} \;  | \; s_{i}\geq
0 \} =\{v\in\rep\mid\relu_{B}(v)=v \}. \]By assumption, 
$\relu_{B}$ is $G$-equivariant and we get $g\acts v\in C$
whenever $v\in C$.

Hence, $G$ acts on $C$ and in particular permutes 
the rays $\{\R_{\geq 0}b_{i}\}$. In other words, all $g\in G$ act in the basis $B$ by a permutation matrix times a diagonal matrix.

It remains to argue that the diagonal matrices can be chosen to be
trivial. By classical representation theory, $V$ admits a positive
definite, symmetric and $G$-invariant form $\langle -, -\rangle$ (see
e.g. \cite[\S 1.3, Remark]{Se-rep-theory-finite-groups}. By
rescaling each $b_i$ if necessary, we can assume that each $b_i$ is of unit length,
i.e. $\langle b_i, b_i \rangle = 1$. Now, $\langle gb_i, gb_i \rangle
= \langle b_i, b_i \rangle = 1$ and hence $G$ permutes the set $\{
b_1, \dots, b_n \}$ which is what we wanted to show.
\end{proof}

\begin{Remark}\label{R:PLRepsGEquivariant}
The assumption in \autoref{T:PLRepsGEquivariant}
that $G$ is finite cannot be dropped. For example, fix $\lambda\in\R$
and let $G = \Z$ act on $\R$ so that $m\in\Z$ acts by $\lambda^{m}$. Then
$\relu$ is $\Z$-equivariant  
if and only if $\lambda>0$.
\end{Remark}

\begin{proof}[Proof (for a general piecewise linear activation
function)] Consider an element $g \in G$, and a fixed but arbitrary element
of our basis $b$. Let us write the action of $g$ on $b$ as
\begin{equation} \label{eq:g}
g \acts b = \sum_i \gamma_{i} b_i.
\end{equation}
Our invariance condition reads $\plmap_{B}(g \acts v) = g \acts \plmap_{B}(v)$ for
any $v \in V$. In particular, for $v = \lambda b$, we
deduce
\[
g \acts \plmap_B(\lambda b) = \plmap_B( g \acts (\lambda b))
\]
which we expand to obtain
\[
\sum_i  \plmap(\lambda) \gamma_{i} b_i = \sum_i \plmap(\lambda
\gamma_i) b_i.
\]
Comparing coefficients yields
\begin{equation}
\label{eq:PLlambda}
\plmap(\lambda \gamma_i) = \plmap(\lambda) \gamma_{i}\quad
\text{for all $\lambda$ and $i$.}
\end{equation}
If $\gamma_i \ne 0$ then this is easily seen to be equivalent to
\begin{equation}
\label{eq:PLlambdainv}
\plmap(\lambda \gamma^{-1}_i) = \plmap(\lambda) \gamma^{-1}_{i}\quad
\text{for all $\lambda$ and $i$, if $\gamma_i \ne 0$.}
\end{equation}

We claim that if $\gamma_i \ne 0, 1, -1$ then \autoref{eq:PLlambda}
and \autoref{eq:PLlambda} force $\plmap$ to be almost linear. We
illustrate this first under the assumption that $\gamma_i > 1$. Because $\plmap$ may be
decomposed into finitely many pieces, all of which are affine
linear, we may find $m$ such that $\plmap$ is given by
\[
\plmap(x) = a_{>}x + b \quad \text{for all $x \in [\gamma_i^m, \gamma_i^{m+1}]$.}
\]
Expanding $\plmap (\gamma_i^{m+1}) = \plmap (\gamma_i^m)\gamma_i$ yields $b =
0$, i.e. that $\plmap$ is linear on $[\gamma_i^m, \gamma_i^{m+1}]$. Now
for any $x >0$ we may find $\ell$ so that $x\gamma_i^\ell \in
[\gamma_i^m, \gamma_i^{m+1}]$ and expanding $ax\gamma_i^l =
\plmap (x\gamma_i^\ell) = \plmap (x) \gamma_i^\ell$ yields that
\[
\plmap(x) = a_{>}x \quad \text{for all $x > 0$.}
\]
Similar considerations show that $\plmap(x) = a_{<} x$ for all $x
< 0$. Thus, by continuity,
\begin{equation} \label{eq:plmapform}
\plmap(x) = \begin{cases} a_{>}x & \text{for $x \ge 0$,} \\
a_{<}x & \text{for $x \le 0$.} \end{cases}
\end{equation}
The same conclusion holds when $0 < \gamma_i < 1$ by using
\autoref{eq:PLlambdainv} in place of \autoref{eq:PLlambda}.

If $\gamma_i < 0$ and $\gamma_i \ne -1$ then, by using
$\plmap(\lambda \gamma_i^2) = \plmap(\lambda) \gamma_i^2$ we can
also conclude that $\plmap$ has the form
\autoref{eq:plmapform}. However now something stronger is true. For
$\lambda > 0$ we have 
\[
a_{<}\lambda \gamma_i = \plmap(\lambda \gamma_i) = \plmap(\lambda) \gamma_i = a_{>}\lambda \gamma_i
\]
and hence $a_{<0} = a_{>0}$. That is, $\plmap$ is linear, which contradicts
our assumptions on $\plmap$.

Finally, when $\gamma_i = -1$ then \autoref{eq:PLlambda} says that
$\plmap$ is odd, which again contradicts our assumptions.

The upshot is that all $\gamma_i \ge 0$ in \autoref{eq:g}. In other
words, $G$ preserves the cone
\[
C = \{ \sum_{i=1}^n s_i b_i \; | \; s_i \ge 0 \}
\]
and we may conclude as in the case of $\relu$.
\end{proof}

\begin{Remark} We were a little surprised to see the appearance of
the assumption that $\plmap$ is not odd pop up in 
\autoref{T:PLRepsGEquivariant}. In hindsight it is not surprising: the
first example of a representation which is not permutation is the sign
representation of $G = \Z/2\Z$ and $\plmap$ is equivariant if and
only if $\plmap$ is odd.
\end{Remark}

\begin{Remark}
The reader may easily adapt the proofs above to the case of
common non-piecewise linear activation functions like the sigmoid function
$1/(1+e^{-x})$ or GELU \cite{hendrycks2016gaussian}. There one
comes to the stronger conclusion that all $\gamma_i$ in the proof
are 1.  (Note that this conclusion does not hold for
$\tanh$, another common activation function, as
it is odd.)
\end{Remark}

\subsection{Piecewise linear representation theory -- first example}\label{SS:OverviewPiecewise}

Representation theory 
is intrinsically about linear maps: actions are 
given by matrices and representations are related by 
linear maps. For us \emph{piecewise linear representation theory} keeps the linear
actions, but allows piecewise linear maps between representations. In
particular, the maps appearing in equivariant neural networks fall
under the umbrella of piecewise linear representation
theory.

In this paper we start a systematical study 
of piecewise linear representation theory. Hereby 
we focus on the \emph{piecewise linear Schur lemma}.
As we will see, the piecewise linear version of Schur's lemma is very different from the linear version.

For starters, consider the group $\cyclic[4]=\Z/4\Z$ acting by 
$90$ degree rotation on $\R^{2}$. In other words, 
$\cyclic[4]$ acts on a pizza by rotating four equally sized 
pizza slices (the reader might want to consult \autoref{SS:CyclicHyperplanes} for a more thorough description):
\begin{gather*}
\begin{tikzpicture}[anchorbase,thick,font=\sffamily\Large]
\draw (0,0) circle (1.55cm);
\draw[very thick,fill=spinach!50] (0,0) -- (0:1.55) arc(0:90:1.55) -- cycle;
\draw[latex-latex] (0:1.0) arc(0:90:1.0) node[midway,right]{\tiny $\pi/2$};
\end{tikzpicture}
\xrightarrow{\text{rotate}}
\begin{tikzpicture}[anchorbase,thick,font=\sffamily\Large]
\draw (0,0) circle (1.55cm);
\draw[very thick,fill=tomato!50] (0,0) -- (90:1.55) arc(90:180:1.55) -- cycle;
\draw[latex-latex] (90:1.0) arc(90:180:1.0) node[midway,left]{\tiny $\pi/2$};
\end{tikzpicture}
\xrightarrow{\text{rotate}}
\begin{tikzpicture}[anchorbase,thick,font=\sffamily\Large]
\draw (0,0) circle (1.55cm);
\draw[very thick,fill=orchid!50] (0,0) -- (180:1.55) arc(180:270:1.55) -- cycle;
\draw[latex-latex] (180:1.0) arc(180:270:1.0) node[midway,left]{\tiny $\pi/2$};
\end{tikzpicture}
\xrightarrow{\text{rotate}}
\begin{tikzpicture}[anchorbase,thick,font=\sffamily\Large]
\draw (0,0) circle (1.55cm);
\draw[very thick,fill=magenta!50] (0,0) -- (270:1.55) arc(270:360:1.55) -- cycle;
\draw[latex-latex] (270:1.0) arc(270:360:1.0) node[midway,right]{\tiny $\pi/2$};
\end{tikzpicture}
\,.
\end{gather*}
Consider a piecewise linear $\R$-valued function on one of the pizza
slices above. By restriction, it induces a linear
function on each edge of the slice. Consider a function which
induces the same value along each edge. Such a function glues
together to give a piecewise linear map $\R^{2}\to\R$ which is
$\cyclic[4]$-equivariant. Here $\R$ denotes the trivial 
$\cyclic[4]$-representation. Thus, Schur's lemma is false in the
piecewise linear world: there exist interesting piecewise linear
maps between \emph{different} (nonisomorphic) simple representations.

This is not an isolated example but actually prototypical. Explicitly, doing the same, but swapping signs with each rotation defines a $\cyclic[4]$-equivariant map with respect to the rotation $\cyclic[4]$-action 
on $\R^{2}$ and the sign $\cyclic[4]$-action on $\R$.

The graphs of explicit piecewise linear maps from the simple
representation $\R^2$ to the trivial and sign representation are given
as follows:
\begin{gather*}
\text{to trivial}\colon
\begin{tikzpicture}[anchorbase,scale=1]
\node at (0,0) {\includegraphics[height=2.5cm]{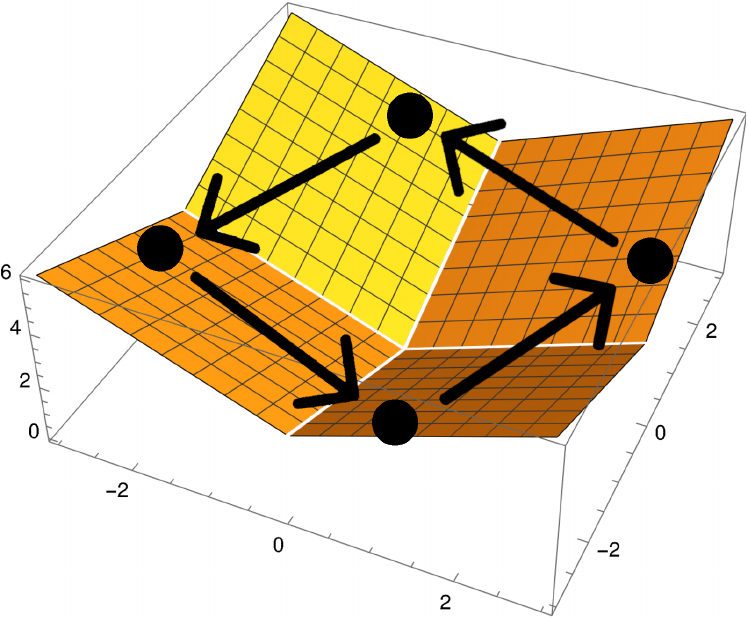}};
\end{tikzpicture}
,\quad
\text{to sign}\colon
\begin{tikzpicture}[anchorbase,scale=1]
\node at (0,0) {\includegraphics[height=2.5cm]{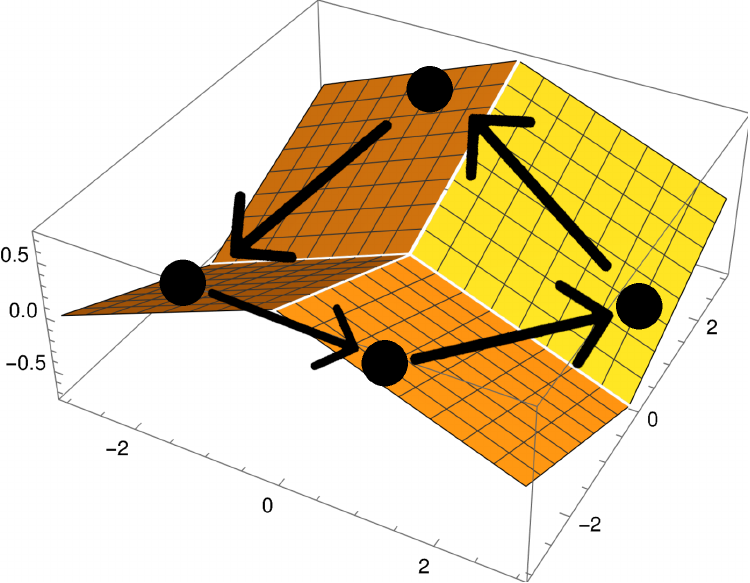}};
\end{tikzpicture}
.
\end{gather*}
We have illustrated the rotation orbit of one point: on the left-hand side the orbit stays at the 
same height while it changes sign with each turn
on the right-hand side.

Note that this construction 
gives uncountable many maps from 
the two dimensional rotation $\cyclic[4]$-representa\-tion 
to the trivial and the sign $\cyclic[4]$-representation. We will see
that there are no nonzero maps back.
The piecewise linear Schur lemma is essentially a hierarchy statement
on $G$-representations.

\subsection{Why is this useful?}\label{SS:OverviewPiecewiseUseful}

In order to explain why this is useful for 
equivariant neural networks let us first assume that 
all maps in such a network are $G$-equivariant and \emph{linear}. In this case \emph{Schur's lemma} implies we can decompose the problem into smaller bits.
To be explicit, say we are interested in computing $\dots\circ f_{2}\circ f_{1}$ for $f_{i}\colon\rep[V]_{i}\to\rep[V]_{i+1}$ being the 
$G$-equivariant linear map from the $i$th to the $(i+1)$th layer. Then Schur's lemma allows us to decompose $\rep[V]_{i}$ into the simple building blocks 
and thus the $f_{i}$ into matrix blocks. Explicitly, 
say $\rep[V]_{i}\cong\simple_{0}\oplus\simple_{1}\oplus\simple_{1}\oplus\simple_{2}$ and $\rep[V]_{i+1}=\rep[V]_{i+2}\cong\simple_{0}\oplus\simple_{1}$ for nonequivalent
simple real $G$-representations $\simple_{i}$.
Then we get the picture
\begin{gather}\label{Eq:OverviewComposition}
f_{1}\leftrightsquigarrow
\scalebox{0.7}{$\begin{tikzpicture}[anchorbase,->,>=stealth',shorten >=1pt,auto,node distance=3cm,
thick,main node/.style={circle,draw,font=\sffamily\Large\bfseries}]
\node[main node] (1) {$\simple_{0}$};
\node[main node] (2) [above of=1,yshift=-1.5cm] {$\simple_{1}$};
\node[main node] (5) [above of=2,yshift=-1.5cm] {$\simple_{1}$};
\node[main node] (6) [above of=5,yshift=-1.5cm] {$\simple_{2}$};
\node[main node] (3) [right of=1] {$\simple_{0}$};
\node[main node] (4) [right of=2] {$\simple_{1}$};
\path[every node/.style={font=\sffamily\small}]
(1) edge [tomato] (3)node[below,xshift=1.5cm]{$f_{0}^{0}$}
(2) edge [tomato] (4)node[above,xshift=1.5cm]{$f_{1}^{1}$}
(5) edge [tomato] (4)node[above,xshift=1.5cm,yshift=-1.5cm]{$f_{1}^{2}$};
\end{tikzpicture}$}
,\quad
f_{2}\leftrightsquigarrow
\scalebox{0.7}{$\begin{tikzpicture}[anchorbase,->,>=stealth',shorten >=1pt,auto,node distance=3cm,
thick,main node/.style={circle,draw,font=\sffamily\Large\bfseries}]
\node[main node] (1) {$\simple_{0}$};
\node[main node] (2) [above of=1,yshift=-1.5cm] {$\simple_{1}$};
\node[main node] (3) [right of=1] {$\simple_{0}$};
\node[main node] (4) [right of=2] {$\simple_{1}$};
\path[every node/.style={font=\sffamily\small}]
(1) edge [tomato] (3)node[below,xshift=1.5cm]{$f_{0}^{0}$}
(2) edge [tomato] (4)node[above,xshift=1.5cm]{$f_{1}^{1}$};
\end{tikzpicture}$}
,\quad
f_{2}\circ f_{1}\leftrightsquigarrow
\scalebox{0.7}{$\begin{tikzpicture}[anchorbase,->,>=stealth',shorten >=1pt,auto,node distance=3cm,
thick,main node/.style={circle,draw,font=\sffamily\Large\bfseries}]
\node[main node] (1) {$\simple_{0}$};
\node[main node] (2) [above of=1,yshift=-1.5cm] {$\simple_{1}$};
\node[main node] (3) [right of=1] {$\simple_{0}$};
\node[main node] (4) [right of=2] {$\simple_{1}$};
\node[main node] (5) [right of=3] {$\simple_{0}$};
\node[main node] (6) [right of=4] {$\simple_{1}$};
\node[main node] (7) [above of=2,yshift=-1.5cm] {$\simple_{1}$};
\node[main node] (8) [above of=7,yshift=-1.5cm] {$\simple_{2}$};
\path[every node/.style={font=\sffamily\small}]
(1) edge [tomato] (3)node[below,xshift=1.5cm]{$f_{0}^{0}$}
(2) edge [tomato] (4)node[above,xshift=1.5cm]{$f_{1}^{1}$}
(3) edge [tomato] (5)node[below,xshift=1.5cm]{$f_{0}^{0}$}
(4) edge [tomato] (6)node[above,xshift=1.5cm]{$f_{1}^{1}$}
(7) edge [tomato] (4)node[above,xshift=1.5cm]{$f_{1}^{2}$};
\end{tikzpicture}$}
.
\end{gather}
Here every $f_{i}^{j}$ is a scalar times the
identity matrix, and the computation of 
$f_{2}\circ f_{1}$ simplifies drastically. For example, 
if one is only interested in the evaluation of the 
composition at $\simple_{0}$, then one only needs to compute a scalar.

Since neural networks do not involve only linear maps,
it is a crucial task to 
study Schur's lemma in the context of piecewise linear maps. For starters, consider the following problem. Let $f\colon\rep[V]\to\rep[W]$ be $G$-equivariant
and piecewise linear. Take two simple real $G$-representations $\simple$ and $\simple[K]$ that appear as direct summands of $\rep[V]$ and $\rep[W]$, respectively. Let $\simple\hookrightarrow\rep$ 
and $\rep[W]\twoheadrightarrow\simple[K]$
be an inclusion and projection. \emph{Schur's lemma for $f$} is then the question whether one can describe 
the $G$-equivariant piecewise linear map
\begin{gather}\label{Eq:IntroductionDecom}
\simple\hookrightarrow\rep
\xrightarrow{f}
\rep[W]\twoheadrightarrow\simple[K].
\end{gather}
As we will see and in contrast to the linear case, the answer crucially depends on $f$. 
A naturally first candidate for $f$ is ReLU and 
we will discuss Schur's lemma for ReLU in this paper.

\subsection{Piecewise linear representation theory}\label{SS:PLRepsPLRep}
The notion of $G$-representations (are always 
on $\R$-vector spaces if not stated otherwise) stays the same as the 
classical one. However, the spaces of homomorphisms we consider have elements being 
$G$-equivariant piecewise linear maps, denoted by
$\plhom(\rep,\rep[W])$. As usual, we denote by $\plend(V)=\plhom(\rep,\rep)$. We therefore
have two different  
categories of $G$-representations: $\lrep$, with the usual 
$G$-equivariant linear maps, and $\plrep$, with 
$G$-equivariant piecewise linear maps.

The following two examples should be compared with
\autoref{T:PLRepsGEquivariant}.

\begin{Example}\label{E:PLRepsGEquivariantZero}
Let $G=\Z/2\Z=\langle a|a^{2}=1\rangle$ acting by $a\mapsto -1$ on
$\R$. Since $1=\relu(1)=\relu(a\acts -1)\neq a\acts\relu(-1)=0$, we
see that $\relu$ is not $G$-equivariant on this $G$-representation. On
$\R^{2}$ we have the $G$-action given by permuting coordinates. Then
$\relu$ is $G$-equivariant with respect to this $G$-representation.
\end{Example}

\begin{Example}\label{E:PLRepsGEquivariant}
As another example, let $G=\Z/3\Z=\langle a|a^{3}=1\rangle$ act on $\R^{2}$ by rotation by $2\pi/3$, so that the action matrix of $a$ is 
$1/2\big((-1,-\sqrt{3}),(\sqrt{3},1)\big)$ in row notation. Let $\relu\colon\R^{2}\to\R^{2}$ be the map $\relu$ defined 
componentwise. This map is not 
$G$-equivariant since $\relu\big(a\acts (1,0)\big)=\relu\big((-1/2,\sqrt{3}/2)\big)=(0,\sqrt{3}/2)$ but $a\acts\relu\big((1,0)\big)=a\acts (1,0)=(-1/2,\sqrt{3}/2)$.

On the other hand, let $G$ act on $\R^{3}=\R\{1,a,a^{2}\}$ 
(with this fixed basis) by left multiplication, {\ie} the action matrix of $a$ in row notation is 
$\big((0,0,1),(1,0,0),(0,1,0)\big)$ where we use $\{1<a<a^{2}\}$. Then $\relu\colon\R^{3}\to\R^{3}$ is $G$-equivariant.
\end{Example}

The following is a crucial difference between linear and piecewise linear 
representation theory:
Let $\simple$ and $\simple[K]$ be two simple real $G$-representations.
The collection $\plhom(\simple,\simple[K])$ of all $G$-equivariant piecewise linear maps $\simple\to\simple[K]$ is an $\R$-vector space for which
Schur's lemma fails in general by \autoref{E:PLRepsGraphs}.

Actually, Schur's lemma in the setting of piecewise linear
representation theory  is ``maximally false'':

\begin{Theorem}[\textbf{Control by normal subgroups}]\label{T:PlRepsSchur}
Let $\simple$ and $\simple[K]$ be two simple real $G$-representations with actions maps $\rho_{\simple}$ and $\rho_{\simple[K]}$.
Then:
\begin{gather*}
\plhom(\simple,\simple[K])\neq 0
\Longleftrightarrow
\ker(\rho_{\simple})\subset\ker(\rho_{\simple[K]}).
\end{gather*}
\end{Theorem}

In particular, a faithful simple real $G$-representation admits
nonzero $G$-equivariant piecewise linear maps to any other
$G$-representation.

\begin{proof}
\textit{$\Rightarrow$.} 
Assume that $f\in\plhom(\simple,\simple[K])$ is nonzero.
Let $H=\ker(\rho_{\simple})\subset G$.
By equivariance, and the fact that $H$ acts trivially on
$\simple$ we have $\im(f)\subset\simple[K]^{H}$.

Clifford theory (see {\eg} \cite[Section 3.13]{Be-rep-cohomology}) shows 
that the restriction of $\simple[K]$ to $H$ is semisimple, and
isomorphic to a direct sum of $G$-conjugates of the same
representation of $H$. Hence, either $\simple[K]^{H}=  0$ or $\simple[K]^{H}=\simple[K]$.
The first case gives a contradiction since then $f$ is zero, the second case gives
$H=\ker(\rho_{\simple})\subset\ker(\rho_{\simple[K]})$, as desired.

\textit{$\Leftarrow$.} By replacing $G$ with $G/\ker(\rho_{\simple})$ we can, and will, assume that 
$\rho_{\simple}$ is faithful. The following well-known 
fact will be useful:

\emph{Claim 1.} There exists $v\in\rep[V]$ 
with trivial stabilizer under the $G$-action.

\emph{Proof of Claim 1.} The set of fixed points 
$\rep[V]^{g}$ is a proper subset of $\rep[V]$ for any $g\in G$ with $g\neq 1$ since $\rep[V]^{g}=\rep[V]$ implies $g\in\ker(\rho_{\simple})$ contradicting faithfulness. Hence, 
$\rep[V]\setminus\bigcup_{g\neq 1}\rep[V]^{g}\neq\emptyset$. Now take $v\in\rep[V]\setminus\bigcup_{g\neq 1}\rep[V]^{g}$ 
and we are done.$\hfill\square$

Consider the set of all $G$-equivariant piecewise linear maps $\rep[V]\to\rep[V]$ denoted as before by $\plend(\rep[V])$. 

Suppose $\plhom(\simple,\simple[K])\neq\{0\}$ 
and take some nonzero $f\in\plhom(\simple,\simple[K])$. After choosing a basis of $\simple[K]$, write it as 
$f=(f_{1},\dots,f_{m})$ where 
$m=\dim_{\R}\simple[K]$. Note that 
$f$ thus gives us a piecewise linear maps on $\simple$ which transforms 
according to $\rho_{\simple[K]}$. In other words, we 
would have found a $G$-equivariant embedding
$\simple[K]\subset\plend(\simple)$. On the other hand, 
if we can find a $G$-equivariant embedding
$\simple[K]\subset\plend(\simple)$, then we can produce a nonzero
map $f\in\plhom(\simple,\simple[K])$.

It thus remains to find such an embedding. Because
every simple $G$-representation embeds into the regular representation, it is enough to find a $G$-equivariant embedding
$\regrep\subset\plend(\simple)$. To this end, let us use $v$ from
Claim 1.

The points in $\{g\acts v\mid g\in G\}$ are all distinct and we can find a piecewise linear function $f$ which is $1$ at $v$ and $0$ at $\{g\acts v\mid g\in G\}\setminus\{v\}$.

\emph{Claim 2.} $g\mapsto g\acts f$ extends to a $G$-equivariant embedding
$\regrep\subset\plend(\simple)$.

\emph{Proof of Claim 2.} The map is $G$-equivariant, by construction. To check that this is an embedding suppose $\sum_{g\in G}s_{g}g\acts f=0$. Evaluating at $g\acts v$ gives $s_{g}=0$, so linear independence follows.$\hfill\square$

The proof completes.
\end{proof}

\begin{Remark}\label{R:PlRepsSchur}
The proof of \autoref{T:PlRepsSchur} is an adaption of 
a familiar fact in representation theory: the so-called 
\emph{Burnside--Brauer--Steinberg theorem}, see 
{\eg} \cite{Br-burnside-brauer-steinberg}, \cite{St-burnside-brauer-steinberg} or \cite{St-burnside-brauer}, which says that every simple $G$-representation appears in some tensor power of a faithful $G$-representation.
\end{Remark}

\begin{Proposition}\label{P:PlRepsSchur}
For $G$-representations $\rep$, $\rep[W]$ 
we have
$\plhom(\rep,\rep[W])=0$ or 
$\dim_{\R}\plhom(\rep,\rep[W])>|\N|$.
\end{Proposition}

\begin{proof}[Proof sketch by example]
Instead of giving a proof here, which is left to the reader, we give an illustrative and prototypical example from which a general proof can be derived.

Consider $\cyclic[6]=\Z/6\Z=\langle a|a^{6}=1\rangle$, i.e. the \emph{cyclic group} of order six. More on 
piecewise linear representation of cyclic groups will follow in \autoref{S:Cyclic}; for now we fix $\rep$ 
to be the two dimensional $\cyclic[6]$-representation 
where $a$ acts by counterclockwise rotation by $2\pi/6$. We think of $\rep$ as being the $x$-$y$-plane. Let $\rep[W]$ be the 
trivial $\cyclic[6]$-representation.

We will explain how to first construct a nonlinear and nonzero map $f\in\plhom(\rep,\rep[W])$, and then we explain how to massage $f$ to construct uncountable many 
such maps.

We first fix a fundamental cone filling $1/6$ of the $x$-$y$-plane. Illustrated as a pizza slice:
\begin{gather*}
\begin{tikzpicture}[anchorbase,thick,font=\sffamily\Large]
\draw (0,0) circle (2cm);
\draw[very thick,fill=spinach!50] (0,0) -- (0:2) arc(0:60:2) -- cycle;
\draw[latex-latex] (0:1.0) arc(0:60:1.0) node[midway,right]{\tiny $2\pi/6$};
\end{tikzpicture}
\xrightarrow{a\acts\placeholder}
\begin{tikzpicture}[anchorbase,thick,font=\sffamily\Large]
\draw (0,0) circle (2cm);
\draw[very thick,fill=tomato!50] (0,0) -- (60:2) arc(60:120:2) -- cycle;
\draw[latex-latex] (60:1.0) arc(60:120:1.0) node[midway,above]{\tiny $2\pi/6$};
\end{tikzpicture}
\xrightarrow{a\acts\placeholder}
\begin{tikzpicture}[anchorbase,thick,font=\sffamily\Large]
\draw (0,0) circle (2cm);
\draw[very thick,fill=orchid!50] (0,0) -- (120:2) arc(120:180:2) -- cycle;
\draw[latex-latex] (120:1.0) arc(120:180:1.0) node[midway,left]{\tiny $2\pi/6$};
\end{tikzpicture}
\xrightarrow{a\acts\placeholder}
\dots
\end{gather*}
We have also illustrated the rotation action of $a$. 
Now, fix a triangular spike on this fundamental cone, that is, a function that is zero at the boundary on the cone and a triangular spike in between:
\begin{gather*}
\text{$\triangle$ spike}\colon
\begin{tikzpicture}[anchorbase,thick,font=\sffamily\Large]
\draw (0,0) to (1,1) to (2,0);
\end{tikzpicture}
.
\end{gather*}
This gives a nonlinear, but piecewise linear, nonzero map on the fundamental cone. Now rotating this map by 
$2\pi k/6$ for $k\in\{1,\dots,5\}$ gives six such maps 
that we can glue together to obtain one map:
\begin{gather*}
f\leftrightsquigarrow
\begin{tikzpicture}[anchorbase,thick,font=\sffamily\Large]
\draw (0,0) circle (2cm);
\draw[very thick,fill=spinach!50] (0,0) -- (0:2) arc(0:60:2) -- cycle;
\draw[thin,spinach,dotted] (0:1.0) arc(0:60:1.0) node[midway,right]{\tiny $\triangle$ spike};
\draw[very thick,fill=tomato!50] (0,0) -- (60:2) arc(60:120:2) -- cycle;
\draw[thin,tomato,dotted] (60:1.0) arc(60:120:1.0) node[midway,above]{\tiny $\triangle$ spike};
\draw[very thick,fill=orchid!50] (0,0) -- (120:2) arc(120:180:2) -- cycle;
\draw[thin,orchid,dotted] (120:1.0) arc(120:180:1.0) node[midway,left]{\tiny $\triangle$ spike};
\draw[very thick,fill=magenta!50] (0,0) -- (180:2) arc(180:240:2) -- cycle;
\draw[thin,magenta,dotted] (180:1.0) arc(180:240:1.0) node[midway,left]{\tiny $\triangle$ spike};
\draw[very thick,fill=blue!50] (0,0) -- (240:2) arc(240:300:2) -- cycle;
\draw[thin,blue,dotted] (240:1.0) arc(240:300:1.0) node[midway,below]{\tiny $\triangle$ spike};
\draw[very thick,fill=pink!50] (0,0) -- (300:2) arc(300:360:2) -- cycle;
\draw[thin,pink,dotted] (300:1.0) arc(300:360:1.0) node[midway,right,red!75]{\tiny $\triangle$ spike};
\end{tikzpicture}
.
\end{gather*}
This map is the desired map $f$. Note moreover that we 
could have use any piecewise linear map instead of the 
triangular spike as long as the boundary hits zero.
Thus, we get uncountable many, and linearly independent, such maps and the example completes.
\end{proof}

\begin{Lemma}\label{L:PLRepsGraphs}
Let $f\colon\rep\to\rep$ be a $G$-equivariant piecewise linear endomorphism 
of a real $G$-representation $\rep$. Fix projections $p_{\simple[K]}\colon\rep\to\simple[K]$ 
and their respective inclusions $i_{\simple}\colon\simple\to\rep$ for all 
simple real $G$-representations $\simple$, $\simple[K]$ that appear in $\rep$.
Then $p_{\simple[K]}\circ f\circ i_{\simple}\colon\simple\to\simple[K]$ 
is $G$-equivariant piecewise linear.
\end{Lemma}

\begin{proof}
The property of being $G$-equivariant piecewise linear
is preserved by composition, {\cf} \autoref{L:PLRepsMaps} for part of 
the statement.
\end{proof}

We will often use 
\autoref{T:PLRepsGEquivariant} and \autoref{L:PLRepsGraphs} silently, {\eg} the examples sections.

\begin{Notation}\label{N:PLRepsComposition}
Write $f_{\simple}^{\simple[K]}=p_{\simple[K]}\circ f\circ i_{\simple}\colon\simple\to\simple[K]$. We also write 
$f_{i}^{j}$ for $p_{\simple_{j}}\circ f\circ i_{\simple_{i}}$
for short.
\end{Notation}

The maps from \autoref{T:PLRepsGEquivariant} combined with 
\autoref{L:PLRepsGraphs} are 
elements in $\plhom(\simple,\simple[K])$, {\eg} 
$\relu_{\simple}^{\simple[K]}\in\plhom(\simple,\simple[K])$. It is 
crucial to determine whether they are zero or not.

\begin{Example}\label{E:PLRepsGraphs}
Let $\regrep\cong\R^{3}$ denote the regular 
($G=\Z/3\Z$)-representation as in \autoref{E:PLRepsGEquivariant}. $\regrep\cong\R^{3}$ decomposes into a one dimensional trivial $G$-representation $\simple_{0}$ and the two dimensional 
simple real $G$-representation $\simple_{1}$ from the first paragraph of \autoref{E:PLRepsGEquivariant}. The change of basis matrix 
that realizes this decomposition is
\begin{gather*}
\bmat=
\begin{psmallmatrix}
1 & -\frac{\sqrt{3}}{2} & -\frac{1}{2} \\
1 & \frac{\sqrt{3}}{2} & -\frac{1}{2} \\
1 & 0 & 1 \\
\end{psmallmatrix}
,\quad
\bmat^{-1}=
\begin{psmallmatrix}
\frac{1}{3} & \frac{1}{3} & \frac{1}{3} \\
-\frac{1}{\sqrt{3}} & \frac{1}{\sqrt{3}} & 0 \\
-\frac{1}{3} & -\frac{1}{3} & \frac{2}{3} \\
\end{psmallmatrix}
\end{gather*}
with the indicated inverse. Then, for example, the 
$G$-equivariant piecewise linear map $\simple_{1}\to\simple_{0}$ obtained 
as $\relu_{1}^{0}$ is determined 
by
\begin{gather*}
(a,b)\mapsto (1,0,0)^{T}(\bmat^{-1}\circ\relu\circ\bmat)(0,a,b)
,\quad
\begin{tikzpicture}[anchorbase,scale=1]
\node at (0,0) {\includegraphics[height=2.5cm]{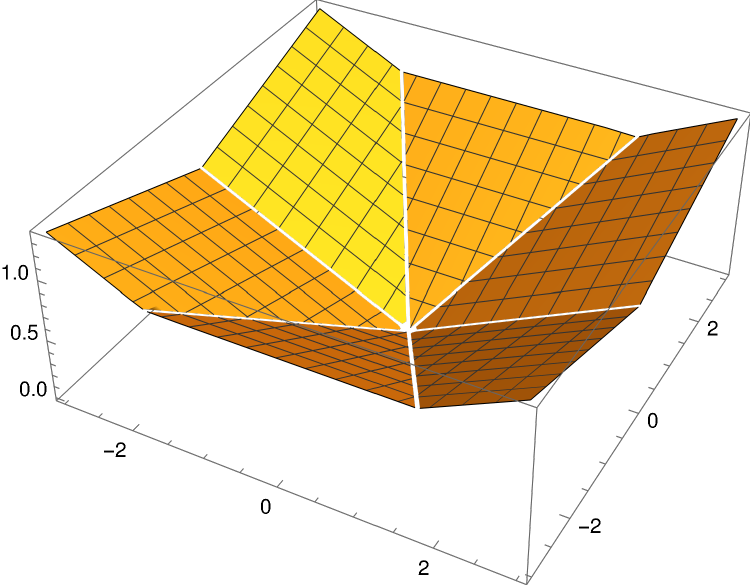}};
\end{tikzpicture}
,\quad
\begin{tikzpicture}[anchorbase,scale=1]
\node at (0,0) {\includegraphics[height=2.5cm]{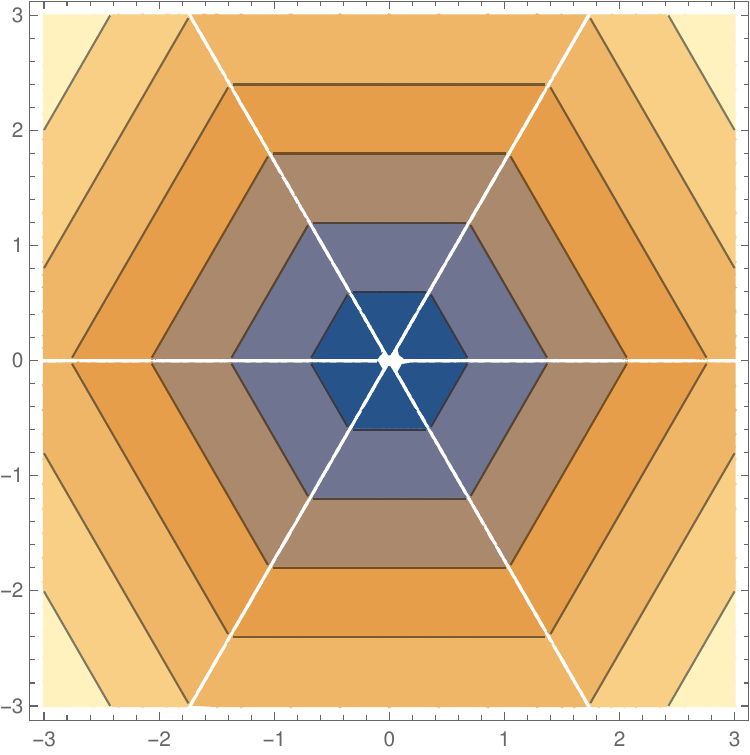}};
\end{tikzpicture}
.
\end{gather*}
As in \autoref{E:PLRepsMapsTwo}, we have plotted the corresponding map.

The other three possibilities can be computed similarly and are as follows.
The $G$-equivariant piecewise linear map $\relu_{0}^{0}$ is 
$\relu$ itself, while $\relu_{0}^{1}$ is zero. The final 
map is $\relu_{1}^{1}$ is a map $\R^{2}\to\R^{2}$
is illustrated on \cite{GiTuWi-pl-reps-code}.
\end{Example}

\subsection{Interaction graphs}\label{SS:PLRepsInteraction}

Suppose we have a $G$-equivariant piecewise linear map 
$f\in\plhom(\rep,\rep[W])$ between real $G$-representations $\rep$ and $\rep[W]$. 
If $f$ is linear, then we can decompose $\rep\cong\simple_{i_{1}}\oplus\dots\oplus\simple_{i_{k}}$ 
and $\rep[W]\cong\simple_{i_{k+1}}\oplus\dots\oplus\simple_{i_{r}}$ and $f$ can be computed blockwise.

For a general $G$-equivariant piecewise linear map this is no longer true, {\cf} \autoref{E:PLRepsGraphs}, and we use the following measurement 
of complexity of how difficult simple real $G$-representations are under $f$.

We will stay with $\permrep=\rep=\rep[W]$ being a permutation $G$-representation for simplicity.

\begin{Definition}\label{D:PLRepsGraphs}
Let $\permrep$ be a permutation $G$-representation.
For $f\in\plend\big(\permrep\big)$ the \emph{interaction graph} $\ggraph{f}{\permrep}$ is the graph with vertices indexed by the simple real constituents $\simple_{i}$ of $V$, counted with multiplicities, and an edge from $\simple_{i}$ 
to $\simple_{j}$ if $f_{i}^{j}\neq 0$.

The \emph{isotypic interaction graph} 
$\iggraph{f}{\permrep}$ is obtained from $\ggraph{f}{\permrep}$ 
by using isotypic components, 
{\ie} by identifying vertices up to isomorphism of real 
$G$-representations (and removing duplicate edges). 
\end{Definition}

The interaction graph depends on the choices of inclusion and projection 
maps for the simple real $G$-representation into $\permrep$, see {\eg} \cite{GiTuWi-pl-reps-code}
%\autoref{SS:SymmetricMinimal} 
for an explicit discussion.

\begin{Lemma}\label{L:PLRepsGraphsUnique}
The isotypic interaction graph is independent of the choice of 
inclusion and projection. The interaction graph is independent of the choice of 
inclusion and projection if and only if the decomposition of 
$\permrep$ into simple real $G$-representations is multiplicity free.
\end{Lemma}

\begin{proof}
In the cases of the lemma the 
respective idempotents in the group algebra are 
unique.
\end{proof}

To simplify notation, we write $\graph[f]=\ggraph{f}{\regrep}$ 
and $\igraph[f]=\iggraph{f}{\regrep}$.

\begin{Remark}
For all examples in this paper we have $\graph[f]=\igraph[f]$, 
and we will strategically ignore the difference.
But the distinction is necessary in general, see \cite{GiTuWi-pl-reps-code}.
\end{Remark}

By \autoref{T:PlRepsSchur}, the interaction graphs need not to be 
symmetric ({\ie} $f_{i}^{j}\neq 0$ does not necessarily imply 
$f_{j}^{i}\neq 0$). Explicitly:

\begin{Example}\label{E:PlRepsGraphs}
For fixed vertex $\simple_{i}$ let $\com{i}{1}$ denote the number of paths 
of length one ending in the fixed vertex.
With respect to \autoref{E:PLRepsGraphs} we get
\begin{gather*}
\graph=
\scalebox{0.7}{$\begin{tikzpicture}[anchorbase,->,>=stealth',shorten >=1pt,auto,node distance=3cm,
thick,main node/.style={circle,draw,font=\sffamily\Large\bfseries}]
\node[main node] (1) {$\simple_{0}$};
\node[main node] (2) [above of=1,yshift=-1.0cm] {$\simple_{1}$};
\path[every node/.style={font=\sffamily\small}]
(1) edge [loop left,blue] (1)
(2) edge [loop left,blue] (2)
(2) edge [tomato] (1);
\end{tikzpicture}$}
,\quad
\begin{gathered}
\com{1}{1}=1,
\\
\com{0}{1}=2.
\end{gathered}
\end{gather*}
The map $\relu\colon\regrep\to\regrep$ will be
\begin{gather*}
\relu\leftrightsquigarrow
\scalebox{0.7}{$\begin{tikzpicture}[anchorbase,->,>=stealth',shorten >=1pt,auto,node distance=3cm,
thick,main node/.style={circle,draw,font=\sffamily\Large\bfseries}]
\node[main node] (1) {$\simple_{0}$};
\node[main node] (2) [above of=1,yshift=-1.0cm] {$\simple_{1}$};
\node[main node] (3) [right of=1] {$\simple_{0}$};
\node[main node] (4) [right of=2] {$\simple_{1}$};
\path[every node/.style={font=\sffamily\small}]
(1) edge [tomato] (3)node[below,xshift=1.5cm]{$f_{0}^{0}$}
(2) edge [tomato] (4)node[above,xshift=1.5cm]{$f_{1}^{1}$}
(2) edge [tomato] (3)node[below,xshift=1.5cm,yshift=-0.35cm]{$f_{1}^{0}$};
\end{tikzpicture}$}
,\quad
\relu\circ\relu\leftrightsquigarrow
\scalebox{0.7}{$\begin{tikzpicture}[anchorbase,->,>=stealth',shorten >=1pt,auto,node distance=3cm,
thick,main node/.style={circle,draw,font=\sffamily\Large\bfseries}]
\node[main node] (1) {$\simple_{0}$};
\node[main node] (2) [above of=1,yshift=-1.0cm] {$\simple_{1}$};
\node[main node] (3) [right of=1] {$\simple_{0}$};
\node[main node] (4) [right of=2] {$\simple_{1}$};
\node[main node] (5) [right of=3] {$\simple_{0}$};
\node[main node] (6) [right of=4] {$\simple_{1}$};
\path[every node/.style={font=\sffamily\small}]
(1) edge [tomato] (3)node[below,xshift=1.5cm]{$f_{0}^{0}$}
(2) edge [tomato] (4)node[above,xshift=1.5cm]{$f_{1}^{1}$}
(2) edge [tomato] (3)node[below,xshift=1.5cm,yshift=-0.35cm]{$f_{1}^{0}$}
(3) edge [tomato] (5)node[below,xshift=1.5cm]{$f_{0}^{0}$}
(4) edge [tomato] (6)node[above,xshift=1.5cm]{$f_{1}^{1}$}
(4) edge [tomato] (5)node[below,xshift=1.5cm,yshift=-0.35cm]{$f_{1}^{0}$};
\end{tikzpicture}$}
,\quad\text{\etc}
\end{gather*}
under the decomposition $\regrep\cong\simple_{0}\oplus\simple_{1}$. Thus, 
$\com{\relu}{i}$ corresponds to the number of calculations needed to evaluate at $\simple_{i}$ after one step.
\end{Example}

More general than the one step complexity 

\begin{Proposition}\label{P:PLRepsEvaluation}
Let $\com{i}{k}$ be the number of path of length $k$ that end in 
the vertex $\simple_{i}$ for $\ggraph{f}{\permrep}$. Then $\com{i}{k}$ equals the number of 
maps needed to calculate $f$ evaluated at $\simple_{i}$.
\end{Proposition}

\begin{proof}
Let $i_{1}\to\dots\to i_{k}\to i$ be a path of length $k$. 
Then $f_{i_{k}}^{i}\circ f_{i_{k-1}}^{i_{k}}\circ\dots\circ f_{1}^{2}$ is needed for the evaluation of $f$ at $\simple_{i}$, and all paths give different maps. By the pictures 
in \autoref{E:PlRepsGraphs}, it is clear that every map 
that we need comes from a path.
\end{proof}

\begin{Remark}\label{R:PLRepsEvaluation}
Motivated by \autoref{P:PLRepsEvaluation} one can ask for which choices 
of inclusions and projections $\ggraph{f}{\permrep}$ is \emph{minimal} 
in the sense that the number of edges is minimal. We do not have a complete
answer but we address this in the special case of the symmetric group on 
$\{1,2,3\}$ in \cite{GiTuWi-pl-reps-code}.
%\autoref{SS:SymmetricMinimal}.
\end{Remark}

\begin{Definition}\label{D:PLRepsSink}
A \emph{sink} in $\ggraph{f}{\permrep}$ is a sink in the graph-theoretical sense 
but ignoring loops. Similarly, a \emph{source} in $\ggraph{f}{\permrep}$ is a source, again ignoring loops.
\end{Definition}

By \autoref{P:PLRepsEvaluation}, sinks in $\ggraph{f}{\permrep}$ are difficult 
to evaluate at, while sources are easy.

\begin{Proposition}\label{P:PLRepsSink}
The trivial $G$-representation $L_{triv}$ is a sink in $\ggraph{f}{\permrep}$, and every other vertex has an edge towards it, {\ie} $\com{triv}{1}$ equals the number of vertices.
\end{Proposition}

\begin{proof}
For any $\permrep$ the projector to the trivial $G$-representation
$\simple_{triv}$ is $f\mapsto\frac{c}{|X|}\sum_{x\in X}f(x)$, where $c\colon X\to\R$ is the constant map with value 
$1$. For any nonzero $G$-subrepresentation $\rep\subseteq\permrep$ (not necessarily simple) there exists $f\in\simple$ and 
$x\in X$ with $f(x)\neq 0$. Now the inclusion 
$\rep\to\permrep$ followed by $\plmap_{X}$ and the 
projector to $\simple_{triv}$ is nonzero. So every $G$-representation 
has an arrow to $\simple_{triv}$. The fact that there 
are no arrows out of the trivial follows from 
\autoref{P:PLMapsFrom1Dim}.
\end{proof}

Piecewise-linear maps can be complicated in general, but we can classify $G$-equivariant piecewise linear maps from a one dimensional representation.

\begin{Proposition}\label{P:PLMapsFrom1Dim}
Let $\chi\colon G\to\{\pm 1\}$ be a 
one dimensional character of $G$, and $\rep[\R]_{\chi}$ the 
associated one dimensional representation.
Suppose that $f\colon\rep[\R]_{\chi}\to V$ is a nonzero piecewise linear $G$-equivariant map.

If $\chi$ is nontrivial, then precisely one of the three cases holds:
\begin{enumerate}

\item $f$ is linear, and $\im f\cong\rep[\R]_{\chi}$.

\item $\im f$ is a ray, $\R\im f$ is a trivial representation, and $f(x)=|x|f(1)$.

\item $\im f$ consists of two distinct rays, and $\R\im f$ is isomorphic to the permutation representation on $\{1,-1\}$ defined by $\chi$.
\end{enumerate}

If $\chi$ is trivial, then the linear span $\R\im f$ is always a trivial $G$-subrepresentation, but beyond that there are no further conditions.
\end{Proposition}

In particular, if $\chi$ is nontrivial and $f\colon\rep[\R]_{\chi}\to V$ is a nonzero $G$-equivariant piecewise-linear map to a simple 
$G$-representation, then only the first two cases can occur.
For interaction graphs, this implies that there may be arrows from $\rep[\R]_{\chi}$ to the trivial representation, but not to any other simple representations.

\begin{proof}
Let $v^{+}=f(1)$ and $v^{-} = f(-1)$, and note 
by equivariance that whenever $\chi(g)=-1$, $g$ swaps $v^{+}$ and $v^{-}$.
If $\chi$ is trivial, then $G$ fixes $v^{+}$ and $v^{-}$ pointwise, therefore $\im f$ is contained in a trivial representation.

Assume from now that $\chi$ is nontrivial, with $g\in G$ 
satisfying $\chi(g)=-1$.
\begin{enumerate}
\item If $\im f$ is a line, then it is a subrepresentation since it is $G$-stable.
Since $v^{+}$ and $v^{-}$ must lie on opposite sides of the origin and are interchanged by $g$, we have $v^{-}=-v^{+}$ and hence $f$ is linear.
It is plain to see that $\im f$ is isomorphic to $\rep[\R]_{\chi}$.

\item If $\im f$ is a ray, then $v^{-}$ is a positive multiple of $v^{+}$, but since they are interchanged by $g$ they must be equal.
Hence each is fixed by $G$, and $\R\im f$ is a trivial representation.
Furthermore, $f(x)=|x|v^{+}$.

\item If $\im f$ is two linearly independent rays, each interchanged by $G$, then $\R\im f$ is a two dimensional subrepresentation, where $G$ permutes the two basis vectors $v^{+}$ and $v^{-}$ according to the character $\chi$, as claimed.
\end{enumerate}
The proof completes.
\end{proof}

We remark that there is another way to interpret \autoref{P:PLMapsFrom1Dim}.
Assume that $\chi$ is nontrivial, and that $\rep[\R]_{\chi}$ and $V$ are equipped with $G$-invariant inner products such that $f: \rep[\R]_{\chi} \to V$ is length-preserving (this can always 
be done, by rescaling $f$ by a positive scalar).
Then $f$ either keeps $\rep[\R]_{\chi}$ as it is (linear), folds it in half on top of itself (a ray), or folds it halfway to 90 degrees (two rays).

\subsection{Product groups and these maps}\label{SS:PLRepsProduct}

We will discuss the case of cyclic groups in \autoref{S:Cyclic}, 
and the discussion below then implies that we understand the case of 
general abelian groups as well.

Let $G\cong H\times I$. For the following lemma recall how 
one can tensor representations of $H$ and $I$ 
to get a $G$-representation, see {\eg} \cite[Section 3.2]{Se-rep-theory-finite-groups}, and the 
\emph{Frobenius--Schur indicator} $\nu$ which takes values 
in $\{0,+1,-1\}$ (these cases are called complex, real and quaternionic, respectively), see {\eg} \cite[Exercise 3.38]{FuHa-representation-theory}. 
We will write 
$\simple_{H}^{\C}$ and $\simple_{I}^{\C}$ denote simple complex $H$- respectively $I$-representations and we drop the subscript for the real versions obtained by restricting scalars (such real versions exist if the complex representation is $\R$-valued on characters), 
and write an overline to indicate composition 
of actions with complex conjugation.

\begin{Lemma}\label{L:PLRepsProduct}
Let $G\cong H\times I$.
\begin{enumerate}

\item We have the following.

\begin{enumerate}

\item If at most one of $\simple_{H}^{\C}$ and $\simple_{I}^{\C}$ is 
$\R$-valued on characters, then 
$(\simple_{H}^{\C}\otimes\simple_{I}^{\C})\oplus(\overline{\simple_{H}^{\C}}\otimes\overline{\simple_{I}^{\C}})$ can be given the structure of a simple real $G$-representation.

\item If both, $\simple_{H}^{\C}$ and $\simple_{I}^{\C}$,
are $\R$-valued on characters (note that thus $\nu$ cannot be zero), then we have two cases:

\begin{enumerate}

\item For $\nu(\simple_{H}^{\C})\nu(\simple_{I}^{\C})=1$ we get that 
$\simple_{H}\otimes\simple_{I}$ can be given the structure of a simple real $G$-representation.

\item For $\nu(\simple_{H}^{\C})\nu(\simple_{I}^{\C})=-1$ we get that 
$(\simple_{H}\otimes\simple_{I})^{\oplus 2}$ can be given the structure of a simple real $G$-representation.

\end{enumerate}

\end{enumerate}

\item Every simple real $G$-representation arises in this way.

\item In the notation of (a), 
let $e(\simple_{H})$ and $e(\simple_{I})$ be the associated 
primitive orthogonal idempotents in the regular 
$H$ and $I$-representation, respectively. Then 
$e(\simple_{H})\otimes e(\simple_{I})+\overline{e(\simple_{H})}\otimes\overline{e(\simple_{I})}$, and 
$e(\simple_{H})\otimes e(\simple_{I})$ or $2\big(e(\simple_{H})\otimes e(\simple_{I})\big)$ is the 
primitive orthogonal idempotent in the regular $G$-representation associated to 
$\simple_{H}\otimes_{\R}\simple_{I}$ (in the order as above).

\end{enumerate}
\end{Lemma}

\begin{proof}
The lemma is well-known, but we still give a (dense) proof here.

For a general group $G$ (not necessarily a product group), 
recall that the endomorphism algebra of a simple real $G$-representation 
$\rep$ is either $\R$, $\C$ or $\HH$. In the first case $V\otimes\C$ is 
simple over $\C$. In the second case, $V\otimes\C$ is the direct sum 
of two nonisomorphic simple complex $G$-representations which must afford mutually conjugate simple characters, so be dual to each other.
In the last case, $V\otimes\C$ is a direct sum of two isomorphic complex simple $G$-representations.

The lemma then follows easily from a case-by-case analysis using the above 
description. For example, for the case in (a).(i) one replaces 
\begin{gather*}
a+bi\leftrightsquigarrow
\begin{psmallmatrix}
a & -b
\\
b & a
\end{psmallmatrix}
\end{gather*}
where $i$ is the usual choice of $\sqrt{-1}$.
\end{proof}

\begin{Example}\label{E:PLRepsProduct}
Let $\zroot[n,k]=\exp(2\pi k/n)$.	
Recall that $\Z/3\Z$ has three simple complex $\Z/3\Z$-representations
$\simple_{i}^{\C}$ for $i\in\{0,1,2\}$, all of dimension one, where 
$\zroot[3,1]^{i}$ is the image of the generator $[1]\in\Z/3\Z$. Similarly, $\Z/5\Z$ has five 
simple complex $\Z/5\Z$-representations
$\simple_{j}^{\C}$ for $j\in\{0,1,2,3,4\}$.
The Frobenius--Schur indicators are zero unless $i=0$ or $j=0$ where they take 
value $1$.
Note that {\eg} $\simple_{j}^{\C}\cong\simple_{5-j}^{\C}$
for $j\in\{1,2,3,4\}$.

Thus, \autoref{L:PLRepsProduct} and getting rid of 
redundancy implies that the simple real 
$(\Z/3\Z\times\Z/5\Z)$-representations are of the form
$\simple_{0}\otimes\simple_{0}$, as well as 
$\simple_{1}^{\C}\oplus\overline{\simple_{1}^{\C}}$ and 
$\simple_{j}^{\C}\oplus\overline{\simple_{j}^{\C}}$ for $j\in\{1,2\}$, and finally $(\simple_{1}^{\C}\otimes\simple_{j}^{\C})\oplus(\overline{\simple_{1}^{\C}}\times\overline{\simple_{j}^{\C}})$, again for $j\in\{1,2\}$. The dimensions of these 
$(\Z/3\Z\times\Z/5\Z)$-representations are $1$, $2$, $2$, $2$, $4$ and $4$.
\end{Example}

\begin{Proposition}\label{P:PLRepsProduct}
The intersection graph of $G\cong H\times I$ is the product of the ones of $H$ and $I$.
\end{Proposition}

\begin{proof}
By \autoref{L:PLRepsProduct} and construction.
\end{proof}

\begin{Remark}\label{R:PLRepsProduct}
Every finite abelian group is a product of cyclic groups. Thus, 
\autoref{L:PLRepsProduct}, \autoref{P:PLRepsProduct} and the discussion in \autoref{S:Cyclic} 
determine the piecewise linear maps on and the
intersection graphs of finite abelian groups.
\end{Remark}

\begin{Example}\label{E:PLRepsProductTwo}
Recall that we have seen the intersection graph for $\Z/3\Z$ in 
\autoref{E:PlRepsGraphs}. The intersection graph for $\Z/5\Z$ 
can be found in \autoref{E:CyclicRelUOne} below. We get:
\begin{gather*}
\graph(\Z/3\Z)=
\scalebox{0.7}{$\begin{tikzpicture}[anchorbase,->,>=stealth',shorten >=1pt,auto,node distance=3cm,
thick,main node/.style={circle,draw,font=\sffamily\Large\bfseries}]
\node[main node] (1) {$\simple_{0}$};
\node[main node] (2) [above of=1,yshift=-1.0cm] {$\simple_{1}$};
\path[every node/.style={font=\sffamily\small}]
(1) edge [loop left,blue] (1)
(2) edge [loop left,blue] (2)
(2) edge [tomato] (1);
\end{tikzpicture}$}
,\quad
\graph(\Z/5\Z)=
\scalebox{0.7}{$\begin{tikzpicture}[anchorbase,->,>=stealth',shorten >=1pt,auto,node distance=3cm,
thick,main node/.style={circle,draw,font=\sffamily\Large\bfseries}]
\node[main node] (1) {$\simple_{0}$};
\node[main node] (2) [above left of=1,yshift=-1.0cm] {$\simple_{1}$};
\node[main node] (3) [above right of=1,yshift=-1.0cm] {$\simple_{2}$};
\path[every node/.style={font=\sffamily\small}]
(1) edge [loop below,blue] (1)
(2) edge [loop left,blue] (2)
(3) edge [loop right,blue] (3)
(2) edge [tomato] (1)
edge [<->] (3)
(3) edge [tomato] (1);
\end{tikzpicture}$}
,\\
\Rightarrow
\graph(\Z/3\Z\times\Z/5\Z)=
\scalebox{0.7}{$\begin{tikzpicture}[anchorbase,->,>=stealth',shorten >=1pt,auto,node distance=3cm,
thick,main node/.style={circle,draw,font=\sffamily\Large\bfseries}]
\node[main node] (1) {$\simple_{00}$};
\node[main node] (2) [above left of=1,yshift=-1.0cm] {$\simple_{01}$};
\node[main node] (3) [above right of=1,yshift=-1.0cm] {$\simple_{02}$};
\node[main node] (4) [above of=1,yshift=0.9cm] {$\simple_{10}$};
\node[main node] (5) [above left of=4,yshift=-0.95cm] {$\simple_{11}$};
\node[main node] (6) [above right of=4,yshift=-0.95cm] {$\simple_{12}$};
\path[every node/.style={font=\sffamily\small}]
(1) edge [loop below,blue] (1)
(2) edge [loop left,blue] (2)
(3) edge [loop right,blue] (3)
(2) edge [tomato] (1)
edge [<->] (3)
(3) edge [tomato] (1)
(4) edge [loop above,blue] (4)
(5) edge [loop left,blue] (5)
(6) edge [loop right,blue] (6)
(5) edge [tomato] (4)
edge [<->] (6)
(6) edge [tomato] (4)
(4) edge [spinach] (1)
(5) edge [spinach] (1)
edge [spinach] (2)
edge [spinach] (3)
(6) edge [spinach] (1)
edge [spinach] (2)
edge [spinach] (3);
\end{tikzpicture}$}
.
\end{gather*}
Here $\simple_{ij}$ is a shorthand notation for what one gets
using \autoref{L:PLRepsProduct} from $\simple_{i}$ 
of $\Z/3\Z$ and $\simple_{j}$ 
of $\Z/5\Z$.
\end{Example}

%%%%%%%%%%%%%%%%%%%%%%%%%%%%%%%%%%%%%%%%%

\section{Example: cyclic groups}\label{S:Cyclic}

%%%%%%%%%%%%%%%%%%%%%%%%%%%%%%%%%%%%%%%%%

Let $\cyclic=\Z/n\Z=\langle a|a^{n}=1\rangle$ be the \emph{cyclic group} of order $n\in\Z_{\geq 1}$. 
This group acts on the $n$ dimensional $\R$-vector space $\regrep=\R\{x_{0},\dots,x_{n-1}\}$ by
cyclic permutation of the coordinates, {\ie}:
\begin{gather}\label{Eq:CyclicMatrixOne}
a\acts x_{i}=x_{i+1},
\end{gather}
reading indexes modulo $n$.
This (based) $\cyclic$-representation $\regrep=(\regrep,B=\{x_{0},\dots,x_{n-1}\})$ is the \emph{regular $\cyclic$-representation}. The action matrix $\lmat$ of $a$ on $\regrep$ 
on the basis $\{x_{0},\dots,x_{n-1}\}$ is
\begin{gather*}
\lmat=
\begin{psmallmatrix}
0&0&\cdots&0&1\\
1&0&\cdots&0&0\\
0&\ddots&\ddots&\vdots&\vdots\\
\vdots&\ddots&\ddots&0&0\\
0&\cdots&0&1&0
\end{psmallmatrix}
.
\end{gather*}

\subsection{Simple representations, projections and inclusions}\label{SS:CyclicPrIn}

We now compute (or rather recall) the simple real $\cyclic$-representations.

\begin{Notation}\label{N:CyclicEvenOdd}
Let $m$ be determined by $n=2m$ or $n=2m+1$, depending on parity. 
%We use this notation in \autoref{S:Dihedral} as well. 
Moreover, the \emph{rotation angle} $\ran=2\pi/n$ will play a crucial role, and we further let
$\zroot=\exp(2\pi i k/n)$ be the respective \emph{complex roots of unity} 
where $i$ is the usual choice of $\sqrt{-1}$. 
\end{Notation}

For $n$ even, the \emph{trivial $\simple_{0}=\R\{v\}$ and sign $\simple_{m}=\R\{v\}$ $\cyclic$-representations} are
\begin{gather*}
\simple_{0}\colon a\acts v=v,
\quad
\simple_{m}\colon a\acts v=-v.
\end{gather*}
Let $\mat[\text{$0$}]$ denote the $1$-by-$1$ matrix $(1)$, and let $\mat[\text{$m$}]$ denote the $1$-by-$1$ matrix $(-1)$.
In terms of matrices we have $a\mapsto\mat[\text{$0$ or $m$}]\in\End_{\R}(\simple_{\text{$0$ or $m$}})$. If $n$ is odd, then we only define $\simple_{0}$.

We let $k\in\{1,\dots,m-1\}$, for $n$ even, and
$k\in\{1,\dots,m\}$, for $n$ odd. For such $k$ endow the two dimensional $\R$-vector space $\simple_{k}=\R\{v_{k},v_{-k}\}$ 
with the $\cyclic$-action given by
\begin{gather*}
a\acts\simple_{k}=\cos(k\ran)\cdot
v_{k}+\sin(k\ran)\cdot v_{-k}
,\quad
a\acts\simple_{-k}=-\sin(k\ran)\cdot
v_{k}+\cos(k\ran)\cdot v_{-k}.
\end{gather*}
Or in terms of matrices:
\begin{gather}\label{Eq:CyclicMatrixTwo}
a\mapsto\mat=
\begin{pmatrix}
\cos(k\ran) & -\sin(k\ran)
\\
\sin(k\ran) & \cos(k\ran)
\end{pmatrix}
\in\End_{\R}(\simple_{k}).
\end{gather}
The representations $\simple_{k}$ are \emph{rotation $\cyclic$-representations}.

\begin{Lemma}\label{L:CyclicRotation}
As real $\cyclic$-representations we have
\begin{gather}\label{Eq:CyclicDecomposition}
\regrep\cong\simple_{0}\oplus \simple_{1}\oplus\dots\oplus\simple_{m-1}\oplus\simple_{m}.
\end{gather}
(Note that $\dim_{\R}\simple_{m}\in\{1,2\}$, depending on parity.)
All of the appearing real $\cyclic$-representations are simple and pairwise 
nonisomorphic. All simple real $\cyclic$-representations appear in this way.
\end{Lemma}

\begin{proof}
We will use the complex case that can, for example, be found in \cite[Section 5.1]{Se-rep-theory-finite-groups}.	

This well-known fact follows from the splitting of 
$\regrep\otimes_{\R}\C$ over $\C$ into one dimensional $\cyclic$-representations 
$\simple_{\zroot}$ given by the eigenvalues $\zroot$.

If $\zroot\in\R$, then we can regard $\simple_{\zroot}$ as a real 
$\cyclic$-representation, which is simple since it is one dimensional. This happens only if $\zroot=\pm 1$, so either $k\in\{0,m\}$ or $k=0$ depending on parity, and we get the trivial and sign $\cyclic$-representations.

If $\zroot\notin\R$, then we pair $\zroot$ and 
its complex conjugate $\zroot[-k]$. It follows that the complex $\cyclic$-representation $\simple_{\zroot}\oplus\simple_{\zroot[-k]}$ can be regarded as a real $\cyclic$-representation via
\begin{gather*}
\begin{pmatrix}
\zroot & 0
\\
0 & \zroot[-k]
\end{pmatrix}
\sim_{\C}
\mat=
\begin{pmatrix}
\cos(k\ran) & -\sin(k\ran)
\\
\sin(k\ran) & \cos(k\ran)
\end{pmatrix}
.
\end{gather*}
Thus, we get the $\cyclic$-representations $\simple_{k}$.

One easily checks that all of the above are nonisomorphic simple real $\cyclic$-representations, and counting dimensions implies that there are no others since $\dim_{\R}\regrep=n=\sum_{i=1}^{m}\dim_{\R}\simple_{i}$.
\end{proof}

We now describe the projections and inclusions $p_{\simple}\colon \regrep\twoheadrightarrow\simple$ and $i_{\simple}\colon\simple\hookrightarrow\regrep$
for $\simple$ a simple real $\cyclic$-representation.

To this end, we observe that it is enough to give the base change 
realizing \autoref{Eq:CyclicDecomposition}. For $n$ even 
this works as follows. 
The matrices for $a$ acting on the left and the right-hand side in 
\autoref{Eq:CyclicDecomposition} are $\bmat[M]$ determined by \autoref{Eq:CyclicMatrixOne} 
and the matrix $\bmat[N]=\mathrm{diag}(1,\mat[1],\dots,\mat[m-1],-1)$ with diagonal blocks given by the trivial one-by-one matrix, the matrices in 
\autoref{Eq:CyclicMatrixTwo} and the negative of the trivial one-by-one matrix.

We now define a change-of-basis matrix $\bmat\in\End_{\R}(\regrep)$ between $\bmat[M]$ and $\bmat[N]$, meaning that $\bmat[N]=\bmat^{-1}\bmat[M]\bmat$. This matrix can be defined as follows. 
For $\simple_{0}$ we use the length $n$ 
vector $w_{0}=(1,\dots,1)$, for $\simple_{m}$ we use 
$w_{m}=(-1,1,-1,\dots,-1,1)$. For $\simple_{k}$ we take
\begin{gather*}
w_{k}=\big(-\sin(jk\ran)\big)_{j\in\{1,\dots,n\}},\quad
w_{-k}=\big(\cos(jk\ran)\big)_{j\in\{1,\dots,n\}}.
\end{gather*}
For $n$ odd the construction of these vectors is similar.

\begin{Lemma}\label{L:CyclicProjection}
Let $n$ be even, and let $\bmat=(w_{0},w_{1},w_{-1},
\dots,w_{m-1},w_{-(m-1)},w_{m})$.
\begin{enumerate}

\item Let $\bmat[D]=\mathrm{diag}(1/n,2/n,\dots,2/n,1/n)$ be a diagonal matrix with the indicated diagonal entries.
The matrix $\bmat$ is invertible with $\bmat^{-1}=\bmat[D]\bmat^{T}$.

\item The matrix $\bmat$ 
is $\cyclic$-equivariant and satisfies $\bmat[N]=\bmat^{-1}\bmat[M]\bmat$.

\end{enumerate} 
Similarly for $\bmat=(w_{0},w_{1},w_{-1},\dots,w_{m},w_{-m})$ 
when $n$ is odd.
\end{Lemma}

\begin{proof}
A direct calculation. As a hint how the matrix $\bmat$ appears, use Jordan 
decomposition for $\bmat[M]$ and $\bmat[N]$, say $\bmat[J]=\bmat[C]^{-1}\bmat[M]\bmat[C]=\bmat[E]^{-1}\bmat[N]\bmat[E]$ for a choice of 
a Jordan matrix $\bmat[J]$, and let $\bmat=\bmat[C]\bmat[E]^{-1}$. Alternatively, the 
columns of $\bmat$ consists of linearly independent columns of the projectors $p_{\simple}$, one column if $\dim_{\R}\simple=1$, and two columns for $\dim_{\R}\simple=2$. These projectors can be
easily computed by general theory.
%computed similarly 
%to the ones in \autoref{S:Dihedral}. Then, using the character table of $\cyclic$, 
%apply the same argument as in the 
%proof of \autoref{L:DihedralProjection} below.
\end{proof}

\begin{Example}\label{E:CyclicProjection}
For $n=5$ we get
\begin{gather*}
\bmat[M]=
\begin{psmallmatrix}
0 & 0 & 0 & 0 & 1
\\
1 & 0 & 0 & 0 & 0
\\
0 & 1 & 0 & 0 & 0
\\
0 & 0 & 1 & 0 & 0
\\
0 & 0 & 0 & 1 & 0
\end{psmallmatrix}
,\quad
\bmat[N]=
\begin{psmallmatrix}
1 & 0 & 0 & 0 & 0
\\
0 & \cos(2\pi/5) & -\sin(2\pi/5) & 0 & 0
\\
0 & \sin(2\pi/5) & \cos(2\pi/5) & 0 & 0
\\
0 & 0 & 0 & \cos(4\pi/5) & -\sin(4\pi/5)
\\
0 & 0 & 0 & \sin(4\pi/5) & \cos(4\pi/5)
\end{psmallmatrix}
,\\
\bmat=
\begin{psmallmatrix}
1 & -\sin(2\pi 1/5) & \cos(2\pi 1/5) & -\sin(4\pi 1/5) & \cos(4\pi 1/5)
\\
1 & -\sin(2\pi 2/5) & \cos(2\pi 2/5) & -\sin(4\pi 1/5) & \cos(4\pi 2/5)
\\
1 & -\sin(2\pi 3/5) & \cos(2\pi 3/5) & -\sin(4\pi 1/5) & \cos(4\pi 3/5)
\\
1 & -\sin(2\pi 4/5) & \cos(2\pi 4/5) & -\sin(4\pi 1/5) & \cos(4\pi 4/5)
\\
1 & -\sin(2\pi 5/5) & \cos(2\pi 5/5) & -\sin(4\pi 1/5) & \cos(4\pi 5/5)
\end{psmallmatrix}
.
\end{gather*}
The inverse $\bmat^{-1}$ is obtained by transposing $\bmat$ and then multiplying the the first row by $1/5$ and the other four by $2/5$.
One easily checks that $\bmat[N]=\bmat^{-1}\bmat[M]\bmat$.
\end{Example}

\subsection{ReLU and cyclic groups}\label{SS:CyclicRelU}

We now want to calculate
\begin{gather}\label{Eq:CyclicInclusionProjection}
\relu_{\simple,\simple[K]}
=
\simple
\xrightarrow{i_{\simple}}
\regrep
\xrightarrow{\relu}
\regrep
\xrightarrow{p_{\simple[K]}}\simple[K]
\in\plhom(\simple,\simple[K])
\end{gather}
for simple real $\cyclic$-representations $\simple$ and $\simple[K]$.
Below we will determine whether $\relu_{\simple,\simple[K]}$ is zero or not, and even better, we will explain what these piecewise linear 
$\cyclic$-equivariant maps are.

To this end, we will actually calculate
\begin{gather*}
\regrep
\xrightarrow{\bmat^{-1}}
\regrep
\xrightarrow{\relu}
\regrep
\xrightarrow{\bmat}
\regrep
\end{gather*}
for the matrix $\bmat$ from \autoref{SS:CyclicPrIn}. This will tell us
the nonlinear interactions between the various simple real $\cyclic$-representations.
We denote the composition $\bmat^{-1}\cdot\relu\cdot\bmat$ as $\relu_{\bmat}$.

\begin{Example}\label{E:CyclicRelUOne}
Let $n=5$ and let $e_{i}$ be the standard vector in $\R^{5}$. We calculate that
\begin{gather*}
\text{$\simple_{0}$ component}\colon\relu_{\bmat}(e_{1})=(1,0,0,0,0)
,\\
\text{$\simple_{1}$ component}\colon \relu_{\bmat}(e_{2})\approx(0.31,0.5,-0.07,0,-0.24)
,\quad
\relu_{\bmat}(e_{3})\approx(0.32,0,0.48,0,0.2)
,\\
\text{$\simple_{2}$ component}\colon \relu_{\bmat}(e_{4})\approx(0.31,0,-0.24,0.5,-0.07)
,\quad
\relu_{\bmat}(e_{5})\approx(0.32,0,0.2,0,0.48)
.
\end{gather*}
Note that $\relu_{\bmat}$ is not linear, for example, 
$\relu_{\bmat}(e_{2}+e_{3})\neq\relu(e_{2})+\relu(e_{3})$. Hence, the above calculation does 
not determine the whole map. However, one can guess that the 
interaction graph for $\relu$ is
\begin{gather*}
\graph=
\scalebox{0.7}{$\begin{tikzpicture}[anchorbase,->,>=stealth',shorten >=1pt,auto,node distance=3cm,
thick,main node/.style={circle,draw,font=\sffamily\Large\bfseries}]
\node[main node] (1) {$\simple_{0}$};
\node[main node] (2) [above left of=1,yshift=-1.0cm] {$\simple_{1}$};
\node[main node] (3) [above right of=1,yshift=-1.0cm] {$\simple_{2}$};
\path[every node/.style={font=\sffamily\small}]
(1) edge [loop below,blue] (1)
(2) edge [loop left,blue] (2)
(3) edge [loop right,blue] (3)
(2) edge [tomato] (1)
edge [<->] (3)
(3) edge [tomato] (1);
\end{tikzpicture}$}
.
\end{gather*}
We will see momentarily that this is indeed the case.
\end{Example}

Note that every simple real $\cyclic$-representation $\simple$ has 
one or two associated complex roots of unity and if there are two 
such roots, then they are conjugates. In particular, we can define:

\begin{Definition}\label{D:CyclicRelU}
The \emph{order} $\ord$ of $\simple$ to be the order of 
any of the associated roots of unity. We use this to define
\begin{gather*}
\ordp=
\begin{cases}
\ord & \text{if $\ord$ is odd},
\\
\ord/2 & \text{if $\ord$ is even}.
\end{cases}
\end{gather*}
\end{Definition}

In other words, $\ord$ is the order of the action matrix of $a$.
For the two dimensional simple real $\cyclic$-representations
$\simple_{i}$ the order is just the order of rotation, and 
$\ordp$ appears due to a parity issue.

\begin{Theorem}\label{T:CyclicRelU}
Consider the interaction graph $\graph$.
Every vertex has a loop. Moreover, for $\simple\not\cong\simple[K]$ there is an edge from $\simple$ to $\simple[K]$ if and only if $\ord[\simple[K]]$ divides $\ordp$.

The isotypic interaction graph $\igraph$ is the same as $\graph$.
\end{Theorem}

\begin{proof}
We give an proof of \autoref{T:CyclicRelU}
later in \autoref{SS:CyclicHyperplanes} 
where we explicitly construct 
$\cyclic$-equivariant piecewise linear maps.

For now we obtain \autoref{T:CyclicRelU} 
directly from \autoref{T:PlRepsSchur}
since the order is just an explicit description of the normal subgroup structure of $\cyclic$. 
\end{proof}

\begin{Example}\label{E:CyclicRelUTwo}
We have the following regarding $\graph$.
\begin{enumerate}

\item If $n$ is an odd prime, then $\graph$ has a sink at $\simple_{0}$ and a fully connected graph on the $\simple_{i}$ for $i\neq 0$. For example, for $n=7$:
\begin{gather*}
\graph=
\scalebox{0.7}{$\begin{tikzpicture}[anchorbase,->,>=stealth',shorten >=1pt,auto,node distance=3cm,thick,main node/.style={circle,draw,font=\sffamily\Large\bfseries}]
\node[main node] (1) {$\simple_{0}$};
\node[main node] (2) [above of=1,yshift=-1.2cm] {$\simple_{2}$};
\node[main node] (3) [left of=2] {$\simple_{1}$};
\node[main node] (4) [right of=2] {$\simple_{3}$};
\path[every node/.style={font=\sffamily\small}]
(1) edge [loop below,blue] (1)
(2) edge [loop above,blue] (2)
(3) edge [loop left,blue] (3)
(4) edge [loop right,blue] (4)
(2) edge [tomato] (1)
edge [<->] (3)
edge [<->] (4)
(3) edge [tomato] (1)
edge [bend left=35,<->] (4)
(4) edge [tomato] (1);
\end{tikzpicture}$}
.
\end{gather*}

\item If $n$ is composite, then the story is more interesting. Let us do $n=8$. In this case we get:
\begin{gather*}
\graph=
\scalebox{0.7}{$\begin{tikzpicture}[anchorbase,->,>=stealth',shorten >=1pt,auto,node distance=3cm,thick,main node/.style={circle,draw,font=\sffamily\Large\bfseries}]
\node[main node] (1) {$\simple_{0}$};
\node[main node] (2) [above of=1,yshift=-1.2cm] {$\simple_{4}$};
\node[main node] (3) [above of=2,yshift=-1.2cm] {$\simple_{2}$};
\node[main node] (4) [right of=3,yshift=1.2cm] {$\simple_{3}$};
\node[main node] (5) [left of=3,yshift=1.2cm] {$\simple_{1}$};
\path[every node/.style={font=\sffamily\small}]
(1) edge [loop below,blue] (1)
(2) edge [loop left,blue] (2)
(3) edge [loop above,blue] (3)
(5) edge [loop left,blue] (5)
(4) edge [loop right,blue] (4)
(2) edge [tomato] (1)
(3) edge [tomato] (2)
edge [tomato,bend left] (1)
(4) edge [tomato] (2)
edge [tomato] (3)
edge [tomato,bend left=10] (1)
edge[<->] (5)
(5) edge [tomato] (2)
edge [tomato] (3)
edge [tomato,bend right=10] (1);
\end{tikzpicture}$}
.
\end{gather*}

\item The composite $n=6$ is also interesting as we will see how 
$\ordp$ enters the game, {\ie} we have:
\begin{gather*}
\graph=
\scalebox{0.7}{$\begin{tikzpicture}[anchorbase,->,>=stealth',shorten >=1pt,auto,node distance=3cm,
thick,main node/.style={circle,draw,font=\sffamily\Large\bfseries}]
\node[main node] (1) {$\simple_{0}$};
\node[main node] (2) [above left of=1,yshift=-0.1cm] {$\simple_{1}$};
\node[main node] (3) [right of=1,xshift=-0.8cm,yshift=0.8cm] {$\simple_{3}$};
\node[main node] (4) [above of=1,yshift=-1.5cm] {$\simple_{2}$};
\path[every node/.style={font=\sffamily\small}]
(1) edge [loop below,blue] (1)
(2) edge [loop left,blue] (2)
(3) edge [loop right,blue] (3)
(4) edge [loop above,blue] (4)
(2) edge [tomato] (1)
edge [tomato] (4)
(3) edge [tomato] (1)
(4) edge [tomato] (1);
\end{tikzpicture}$}
.
\end{gather*}
Note that $\ord[\simple_{1}]=6$ and $\ord[\simple_{3}]=2$, however they 
are not connected by an edge since $\ordp[\simple_{1}]=3$.

\end{enumerate}
Interactive examples can be found 
online following th link in \cite{GiTuWi-pl-reps-code}.
\end{Example}

\subsection{Absolute value and cyclic groups}\label{SS:CyclicAbs}

Similarly as in \autoref{SS:CyclicRelU} we now discuss 
\begin{gather*}
\regrep
\xrightarrow{\bmat^{-1}}
\regrep
\xrightarrow{\absf}
\regrep
\xrightarrow{\bmat}
\regrep
\end{gather*}
for the absolute value map $\absf$.
We denote the composition $\bmat^{-1}\cdot\absf\cdot\bmat$ as $\absf_{\bmat}$.

\begin{Example}\label{E:CyclicAbsOne}
We do a similar calculation as in \autoref{E:CyclicRelUOne} but for $n=6$. We get:
\begin{gather*}
\text{$\simple_{0}$ component}\colon
\absf_{\bmat}(e_{1})=(1,0,0,0,0,0)
,\\
\text{$\simple_{1}$ component}\colon 
\absf_{\bmat}(e_{2}+e_{3})\approx(0.91,0,0,0.58,0.089,0)
,\\
\text{$\simple_{2}$ component}\colon 
\absf_{\bmat}(e_{4}+e_{5})\approx(0.91,0,0,-0.58,0.089,0)
,
\\
\text{$\simple_{3}$ component}\colon\absf_{\bmat}(e_{6})=(1,0,0,0,0,0)
.
\end{gather*}
We get the following graph:
\begin{gather*}
\graph[\absf]=
\scalebox{0.7}{$\begin{tikzpicture}[anchorbase,->,>=stealth',shorten >=1pt,auto,node distance=3cm,
thick,main node/.style={circle,draw,font=\sffamily\Large\bfseries}]
\node[main node] (1) {$\simple_{0}$};
\node[main node] (2) [above left of=1,yshift=-0.1cm] {$\simple_{1}$};
\node[main node] (3) [right of=1,xshift=-0.8cm,yshift=0.8cm] {$\simple_{3}$};
\node[main node] (4) [above of=1,yshift=-1.5cm] {$\simple_{2}$};
\path[every node/.style={font=\sffamily\small}]
(1) edge [loop below,blue] (1)
(4) edge [loop above,blue] (4)
(2) edge [tomato] (1)
edge [tomato] (4)
(3) edge [tomato] (1)
(4) edge [tomato] (1);
\end{tikzpicture}$}
.
\end{gather*}
Note that we have the vertex for $\simple_{1}$ has no loop, so the graph is different from the one for $\relu$.
\end{Example}

Recall $\ordp$ as in \autoref{SS:CyclicRelU}. The proof of the 
next theorem will also be given in tandem 
with the proof of \autoref{T:CyclicRelU}, or follows directly from \autoref{T:PlRepsSchur}.

\begin{Theorem}\label{T:CyclicAbs}
There is an edge from $\simple$ 
to $\simple[K]$ in $\graph[\absf]$ if and only if 
$\ord[\simple[K]]$ divides $\ordp$.

The isotypic interaction graph $\igraph[\absf]$ is the same as $\graph[\absf]$.
\end{Theorem}

In contrast to \autoref{T:CyclicRelU} we do not have a 
separate loop condition in \autoref{T:CyclicAbs}.

\begin{Example}\label{E:CyclicAbsTwo}
If $n$ is an odd prime, then $\graph$ and $\graph[\absf]$ are the same. In contrast, if $n=2^{j}$ for some $j\in\Z_{\geq 1}$, then $\graph$ and $\graph[\absf]$ are quite different and $\graph[\absf]$ is only a subgraph of $\graph$. For example, for $n=8$ 
we get 
\begin{gather*}
\graph[\absf]=
\scalebox{0.7}{$\begin{tikzpicture}[anchorbase,->,>=stealth',shorten >=1pt,auto,node distance=3cm,thick,main node/.style={circle,draw,font=\sffamily\Large\bfseries}]
\node[main node] (1) {$\simple_{0}$};
\node[main node] (2) [above of=1,yshift=-1.2cm] {$\simple_{4}$};
\node[main node] (3) [above of=2,yshift=-1.2cm] {$\simple_{2}$};
\node[main node] (4) [right of=3,yshift=1.2cm] {$\simple_{3}$};
\node[main node] (5) [left of=3,yshift=1.2cm] {$\simple_{1}$};
\path[every node/.style={font=\sffamily\small}]
(1) edge [loop below,blue] (1)
(2) edge [tomato] (1)
(3) edge [tomato] (2)
edge [tomato,bend left] (1)
(4) edge [tomato] (2)
edge [tomato] (3)
edge [tomato,bend left=10] (1)
(5) edge [tomato] (2)
edge [tomato] (3)
edge [tomato,bend right=10] (1);
\end{tikzpicture}$}
.
\end{gather*}
For comparison, the graph 
$\graph$ is given in \autoref{E:CyclicRelUTwo}.
\end{Example}

\subsection{The piecewise linear maps for the cyclic group}\label{SS:CyclicHyperplanes}

For fixed $n\in\Z_{\geq 1}$, we consider all the $n$th roots of unity 
$\zroot$ in the $\R$-vector space $\C\cong\R^{2}$.
Here $k\in\{0,\dots,n-1\}$. 

\begin{Definition}\label{D:CyclicHyperplanes}
For $k\in\{0,\dots,n-1\}$ let $H_{\zroot}^{pa}$ and 
$H_{\zroot}^{pe}$ be the hyperplane in $\C$ that is parallel and perpendicular to the line spanned by $\zroot$.
The \emph{hyperplane arrangment} associated to $\simple_{i}$ are as follows.
For $\dim_{\R}\simple_{i}=2$ we call $H_{i}=\bigcup_{j\in\{1,\zroot[i],\zroot[i]^{2},\dots,\}}H_{j}^{pa}$,
and for $\dim_{\R}\simple_{i}=1$ we use $H_{i}=\bigcup_{j\in\{1,\zroot[i],\zroot[i]^{2},\dots,\}}H_{j}^{pe}$.
\end{Definition}

\begin{Example}\label{E:CyclicHyperplanes}
Let $n=6$ 
and recall from \autoref{L:CyclicRotation} 
that $\regrep\cong\simple_{0}\oplus
\simple_{1}\oplus\simple_{2}\oplus\simple_{3}$ 
of dimensions $1$, $2$, $2$ and $1$. The associated hyperplane arrangments are
\begin{gather*}
H_{0}=H_{3}=
\begin{tikzpicture}[scale=0.5,anchorbase,tinynodes]
\draw[help lines,color=gray!30,dashed] (-4.9,-4.9) grid (4.9,4.9);
\draw[thick,orchid] (0,5) to (0,-5)node[left]{$H_{\zroot[0]}^{pe}{=}H_{\zroot[3]}^{pe}$};
\draw[thick,densely dotted,->,spinach] (0,0) to (5,0)node[above,yshift=0.05cm]{$\zroot[0]$};
\draw[thick,densely dotted,->,spinach] (0,0) to (-5,0)node[above,yshift=0.05cm]{$\zroot[3]$};
\end{tikzpicture}
,\quad
H_{1}=H_{2}=
\begin{tikzpicture}[scale=0.5,anchorbase,tinynodes]
\draw[help lines,color=gray!30,dashed] (-4.9,-4.9) grid (4.9,4.9);
\draw[thick,orchid] (5,0) to (-5,0)node[above,yshift=0.05cm]{$H_{\zroot[0]}^{pa}{=}H_{\zroot[3]}^{pa}$};
\draw[thick,orchid] (2.5,-4.33013) to (-2.5,4.33013)node[above]{$H_{\zroot[2]}^{pa}{=}H_{\zroot[5]}^{pa}$};
\draw[thick,orchid] (-2.5,-4.33013) to (2.5,4.33013)node[above]{$H_{\zroot[1]}^{pa}{=}H_{\zroot[4]}^{pa}$};
\draw[thick,densely dotted,spinach] (0,0) to (0,5)node[above,yshift=0.05cm]{$i\zroot[0]$};
\draw[thick,densely dotted,spinach] (0,0) to (4.33013,2.5)node[above]{$i\zroot[5]$};
\draw[thick,densely dotted,spinach] (0,0) to (4.33013,-2.5)node[below]{$i\zroot[4]$};
\draw[thick,densely dotted,spinach] (0,0) to (0,-5)node[below,yshift=-0.05cm]{$i\zroot[3]$};
\draw[thick,densely dotted,spinach] (0,0) to (-4.33013,-2.5)node[below]{$i\zroot[2]$};
\draw[thick,densely dotted,spinach] (0,0) to (-4.33013,2.5)node[above]{$i\zroot[1]$};
\end{tikzpicture}	
\end{gather*}
We additionally indicated the 
roots of unity $\zroot[0]$ and $\zroot[3]$ in the left illustration.
\end{Example}

\begin{Example}\label{E:CyclicHyperplanesFirst}
Starting at $n=3$, the illustrations
\begin{gather*}
\begin{tikzpicture}[anchorbase,scale=1]
\node at (0,0) {\includegraphics[height=2.5cm]{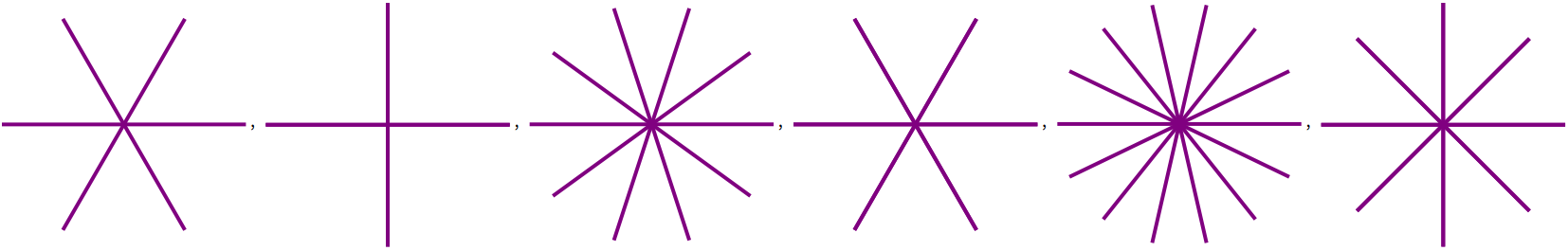}};
\node at (-6.6,-1.4) {$n=3$};
\node at (-3.96,-1.4) {$n=4$};
\node at (-1.32,-1.4) {$n=5$};
\node at (1.32,-1.4) {$n=6$};
\node at (3.96,-1.4) {$n=7$};
\node at (6.6,-1.4) {$n=8$};
\end{tikzpicture}
,
\end{gather*}
show the first few hyperplane arrangements associated to $\simple_{1}$.
\end{Example}

\begin{Remark}\label{R:CyclicHyperplanes}
As in \autoref{E:CyclicHyperplanes}, note that the one dimensional 
simple real $\cyclic$-representations always have the same associated 
hyperplane arrangment. The two dimensional simple real $\cyclic$-representations 
can have the same hyperplane arrangment, as in \autoref{E:CyclicHyperplanes}, 
but need not to. The smallest example where the 
hyperplane arrangements differ is $n=8$. In this case the hyperplane arrangements for $\simple_{1}$ and $\simple_{3}$ are as in 
\autoref{E:CyclicHyperplanesFirst} for $n=8$, while $\simple_{2}$ has the picture for $n=4$ associated to it.
\end{Remark}

Recall the inclusion $i_{\simple}$ from \autoref{Eq:CyclicInclusionProjection}. The pullback $i_{\simple}^{\ast}\colon\simple\to\regrep$ defines a set of hyperplanes $H(\simple)$ in $\simple$.

\begin{Lemma}\label{L:CyclicHyperplanes}
We have $H(\simple_{i})=H_{i}$, and $\relu\circ i_{\simple}$ and $\absf\circ i_{\simple}$ are linear on the complement $\alcove_{i}=\simple\setminus H_{i}$.
Hence, $\alcove=\{\alcove_{1},\dots,\alcove_{k}\}$ is a polyhedral covering for $\relu\circ i_{\simple}$ and $\absf\circ i_{\simple}$. 	
\end{Lemma}

\begin{proof}
Assume $\dim_{\R}\simple_{i}=2$ and $n$ even, and let us focus on 
$\relu$. The lemma follows 
from the explicit form of $\bmat^{-1}$ which was computed in \autoref{L:CyclicProjection}. That is, the respective square submatrix of
$\bmat$ consist of the coordinates of the $n$th roots of unity that appear 
in the definition of the $\cyclic$-action on 
$\simple_{i}$, see \autoref{Eq:CyclicMatrixTwo}. The map $\relu$ is linear unless one hits the input zero. Taking both observations together, we see that $\relu\circ i_{\simple}$ is linear on the alcove between 
$H_{\zroot[0]}^{pa}$ and $H_{\zroot[1]}^{pa}$. Using that $\cyclic$-equivariance,
see \autoref{L:PLRepsGraphs}, the claim follows.

The remaining cases are similar 
but easier and omitted.
\end{proof}

\begin{Lemma}\label{L:CyclicHyperplanesTwo}
\leavevmode	

\begin{enumerate}

\item Let $\dim_{\R}\simple=1$. Then $\alcove$ is of size 
two and has one $\cyclic$-orbit.

\item Let $\dim_{\R}\simple=2$. If $n$ is even, then $\alcove$ is of size $n$ and has one $\cyclic$-orbit.
If $n$ is even, then $\alcove$ is of size $2n$ and has two $\cyclic$-orbits.

\end{enumerate}
\end{Lemma}

\begin{proof}
This follows from \autoref{L:CyclicHyperplanes}.
In more details, let us just do $\simple=\simple_{1}$ 
for $n=5$ and $n=6$. Fixing $A=\alcove_{1}$ for $n=5$ and $n=6$, and additionally $B=\alcove_{2}$ for $n=5$, we get:
\begin{gather*}
n=5\colon
\begin{tikzpicture}[scale=0.5,anchorbase,tinynodes]
\draw[help lines,color=gray!30,dashed] (-4.9,-4.9) grid (4.9,4.9);
\draw[thick,orchid] (5,0) to (-5,0);
\draw[thick,orchid] (1.54508,4.75528) to (-1.54508,-4.75528);
\draw[thick,orchid] (-4.04508,2.93893) to (4.04508,-2.93893);
\draw[thick,orchid] (-4.04508,-2.93893) to (4.04508,2.93893);
\draw[thick,orchid] (1.54508,-4.75528) to (-1.54508,4.75528);
\node[tomato] at (2.37764,0.772542) {$A$};
\node[tomato] at (0,2.5) {$a\acts A$};
\node[tomato] at (-2.37764,0.772542) {$a^{2}\acts A$};
\node[tomato] at (-1.46946,-2.02254) {$a^{3}\acts A$};
\node[tomato] at (1.46946,-2.02254) {$a^{4}\acts A$};
\node[blue] at (-2.37764,-0.772542) {$B$};
\node[blue] at (0,-2.5) {$a\acts B$};
\node[blue] at (2.37764,-0.772542) {$a^{2}\acts B$};
\node[blue] at (1.46946,2.02254) {$a^{3}\acts B$};
\node[blue] at (-1.46946,2.02254) {$a^{4}\acts B$};
\end{tikzpicture}
,\quad
n=6\colon
\begin{tikzpicture}[scale=0.5,anchorbase,tinynodes]
\draw[help lines,color=gray!30,dashed] (-4.9,-4.9) grid (4.9,4.9);
\draw[thick,orchid] (5,0) to (-5,0);
\draw[thick,orchid] (2.5,-4.33013) to (-2.5,4.33013);
\draw[thick,orchid] (-2.5,-4.33013) to (2.5,4.33013);
\node[tomato] at (2.17,1.25) {$A$};
\node[tomato] at (0,2.5) {$a\acts A$};
\node[tomato] at (-2.17,1.25) {$a^{2}\acts A$};
\node[tomato] at (-2.17,-1.25) {$a^{3}\acts A$};
\node[tomato] at (0,-2.5) {$a^{4}\acts A$};
\node[tomato] at (2.17,-1.25) {$a^{5}\acts A$};
\end{tikzpicture}
.
\end{gather*}
The picture generalizes without problems to general $n$.
\end{proof}

The following is useful for calculations:

\begin{Lemma}\label{L:PLRepsCalculateMapsTwo}
If $f$ as in \autoref{L:PLRepsCalculateMaps} is $G$-equivariant, then
it is determined on the polyhedral subsets by evaluate it on one set 
of vectors per $G$-orbit.
\end{Lemma}

\begin{proof}
This follows directly from \autoref{L:PLRepsCalculateMaps}.
\end{proof}

\begin{Lemma}\label{L:CyclicHyperplanesThree}
\leavevmode	

\begin{enumerate}

\item Let $\dim_{\R}\simple=1$. Then $\relu\circ i_{\simple}$ and $\absf\circ i_{\simple}$ are determined by evaluating them on $(1)$.

\item Let $\dim_{\R}\simple=2$. If $n$ is even, then $\relu\circ i_{\simple}$ and $\absf\circ i_{\simple}$ are determined by evaluating them on $(0,1)$.
If $n$ is odd, then $\relu\circ i_{\simple}$ and $\absf\circ i_{\simple}$ 
are determined by evaluating them on $(0,1)$ and $(0,-1)$.

\end{enumerate}
\end{Lemma}

\begin{proof}
This follows from \autoref{L:PLRepsCalculateMapsTwo}, 
\autoref{L:CyclicHyperplanes} and \autoref{L:CyclicHyperplanesTwo}.
\end{proof}

\begin{proof}[Proof of \autoref{T:CyclicRelU} and \autoref{T:CyclicAbs} (via constructing maps)]
To prove the two named theorems, by \autoref{L:CyclicHyperplanesThree}, we 
first need to 
evaluate the matrix $\bmat$ from \autoref{SS:CyclicPrIn} on the images 
of the vectors from \autoref{L:CyclicHyperplanesThree} in $\R^{n}$ upon 
the evident embedding. We then need to compose the result with 
$\relu$ and $\absf$, respectively, and then apply $\bmat^{-1}$. 
We do this for $\relu$ since $\absf$ can be analyzed completely analogously.

Since $\bmat^{-1}$ 
is $\bmat^{T}$ up to a diagonal matrix, see \autoref{L:CyclicProjection}, the 
calculation boils down to computing $(\bmat^{T}\cdot\relu\cdot\bmat)e_{k}$ where $e_{k}$ is the $k$th unit vector. To match the conventions we use, we will 
write $(f_{0},f_{1},f_{-1},\dots)=(e_{1},e_{2},e_{3},\dots)$ 
and let $x_{\pm k}=(\bmat^{T}\cdot\relu\cdot\bmat)f_{\pm k}$.

We get $x_{0}=n\cdot f_{0}$ for the component of the trivial $\cyclic$-representation, and $x_{m}=n/2\cdot f_{0}+n/2\cdot f_{m}$ for the component of the sign $\cyclic$-representation (in case $n$ is even).
For the two dimensional simple real $\cyclic$-representations we 
get $\bmat f_{k}=\big(-\sin(k1\ran),\dots,-\sin(kn\ran)\big)$ 
or $\bmat f_{-k}=\big(\cos(k1\ran),\dots,\cos(kn\ran)\big)$. 
The entries of these vectors that are negative are completely determined by the order $\ord$, and an easy analysis then gives the claimed result. 
For example, consider the regular $n$gon that contains the point $(1,0)$ 
and is centered at $(0,0)$. Then $\bmat f_{-k}$ consists of all the $x$-values 
of this $n$gon, while $\bmat f_{k}$ has its entries being the negatives of the 
$y$-values. Thus, we immediately get the which values of $(\relu\circ\bmat)f_{\pm k}$ are nonzero. Except the 
rows for the trivial and sign $\cyclic$-representation, 
the rows of $\bmat^{T}$ are of the form 
$\big(-\sin(k1\ran),\dots,-\sin(kn\ran)\big)$ and 
$\big(\cos(k1\ran),\dots,\cos(kn\ran)\big)$ so applying $\bmat^{T}$
to $(\relu\circ\bmat)f_{\pm k}$ implies the result. The case for general $k$ is similar and omitted.

Finally, we clearly have $\graph=\igraph$ and the proof completes.
\end{proof}

One can now further analyze the precise form of the various 
piecewise linear maps.
An example of $\relu_{1}^{1}\colon\simple_{1}\to\simple_{1}$ is displayed in \autoref{E:PLRepsGraphs}, 
and the general picture is similar. The map $\relu_{0}^{0}$ is just 
$\relu$ itself, while, for $n$ is even, the map $\relu_{m}^{0}$ is the absolute value.

Let us now analyze the piecewise linear maps in the case of
$f=\relu_{1}^{0}\colon\simple_{1}\to\simple_{0}$ 
and $g=\relu_{1}^{m}\colon\simple_{1}\to\simple_{m}$ for $n$ even or
$n\equiv 0\bmod 4$, respectively,  
which we can display as a map $\R^{2}\to\R$ using level sets.

By the above, $f(0,1)\in\R_{>0}$, more precisely, $f(0,1)$ is equal to $1/n$ times the sum over all positive numbers in $\big(\cos(i\ran)\big)_{i=0}^{n-1}$.
For $n$ even this information is enough to determine the whole map 
since there is only one $\cyclic$-orbit. In other words, in this case the level sets are regular $n$-gons.

For example, for $n=4$ one gets (the level sets are the squares)
\begin{gather*}
n=4\colon
\begin{tikzpicture}[scale=0.5,anchorbase,tinynodes]
\draw[help lines,color=gray!30,dashed] (-4.9,-4.9) grid (4.9,4.9);
\draw[thick,orchid] (0,5) to (0,-5);
\draw[thick,orchid] (5,0) to (-5,0);
\foreach \x in {1,...,4} {
\draw[spinach] (\x,0) to (0,\x) to (-\x,0) to (0,-\x) to (\x,0);}
\end{tikzpicture}
\end{gather*}
and the map $f$ is the one in \autoref{E:PLRepsMapsTwo}.

For $g$ something similar happens, but now with a sign swap after 
each rotation. For $n=4$ one gets the map in
\autoref{E:PLRepsMapsTwo}.

\begin{Remark}
We have also worked out 
\autoref{T:CyclicRelU} and \autoref{T:CyclicAbs} for
the dihedral group and partially for the
symmetric group. An additional document 
containing details can be found following the links in \cite{GiTuWi-pl-reps-code}.
\end{Remark}

\newcommand{\etalchar}[1]{$^{#1}$}

\end{document}